\documentclass{article}

\usepackage{microtype}
\usepackage{graphicx}
\usepackage{booktabs} 
\usepackage{subfig}
\usepackage{hyperref}
\usepackage{xurl}
\usepackage{xcolor}
\usepackage{comment}

\usepackage[accepted]{icml2026}

\usepackage{amsmath}
\usepackage{amssymb}
\usepackage{mathtools}
\usepackage{amsthm}
\usepackage{caption}
\usepackage{subcaption}
\usepackage{multirow}
\usepackage{makecell}
\usepackage{stackengine}
\usepackage{graphicx} 
\usepackage{booktabs} 
\usepackage{caption} 
\usepackage{soul}
\usepackage{bm}

\usepackage[capitalize,noabbrev]{cleveref}

\theoremstyle{plain}

\theoremstyle{definition}

\usepackage{amsthm}

\definecolor{themeGreen}{RGB}{84, 122, 65}   
\definecolor{themePurple}{RGB}{128, 126, 222} 
\definecolor{themeYellow}{RGB}{246, 185, 59}  

\usepackage[textsize=tiny]{todonotes}

\icmltitlerunning{Baguan-TS: A Sequence-Native In-Context Learning Model for Time Series Forecasting with Covariates}

\begin{document}

\twocolumn[
\icmltitle{Baguan-TS: A Sequence-Native In-Context Learning Model\\for Time Series Forecasting with Covariates}




\begin{icmlauthorlist}
\icmlauthor{Linxiao Yang}{ali}
\icmlauthor{Xue Jiang}{ali}
\icmlauthor{Gezheng Xu}{ali}
\icmlauthor{Tian Zhou}{ali}
\icmlauthor{Min Yang}{ali}
\icmlauthor{Zhaoyang Zhu}{ali}
\icmlauthor{Linyuan Geng}{ali}
\icmlauthor{Zhipeng Zeng}{ali}
\icmlauthor{Qiming Chen}{ali}
\icmlauthor{Xinyue Gu}{ali}
\icmlauthor{Rong Jin}{ali}
\icmlauthor{Liang Sun}{ali}
\end{icmlauthorlist}
\icmlaffiliation{ali}{DAMO Academy, Alibaba Group, Hangzhou, China}





\vskip 0.3in
]

\printAffiliationsAndNotice{}  

\begin{abstract}
Transformers enable in-context learning (ICL) for rapid, gradient-free adaptation in time series forecasting, yet most ICL-style approaches rely on tabularized, hand-crafted features, while end-to-end sequence models lack inference-time adaptation. We bridge this gap with a unified framework, Baguan-TS, which integrates the raw-sequence representation learning with ICL, instantiated by a 3D Transformer that attends jointly over temporal, variable, and context axes. 
To make this high-capacity model practical, we tackle two key hurdles: (i) calibration and training stability, improved with a feature-agnostic, target-space retrieval-based local calibration; and (ii) output oversmoothing, mitigated via context-overfitting strategy. On public benchmark with covariates, Baguan-TS consistently outperforms established baselines, achieving the highest win rate and significant reductions in both point and probabilistic forecasting metrics. Further evaluations across diverse real-world energy datasets demonstrate its robustness, yielding substantial improvements.

\end{abstract}

\section{Introduction}
\begin{figure}[!ht]
    \centering
    \includegraphics[width=1\linewidth]{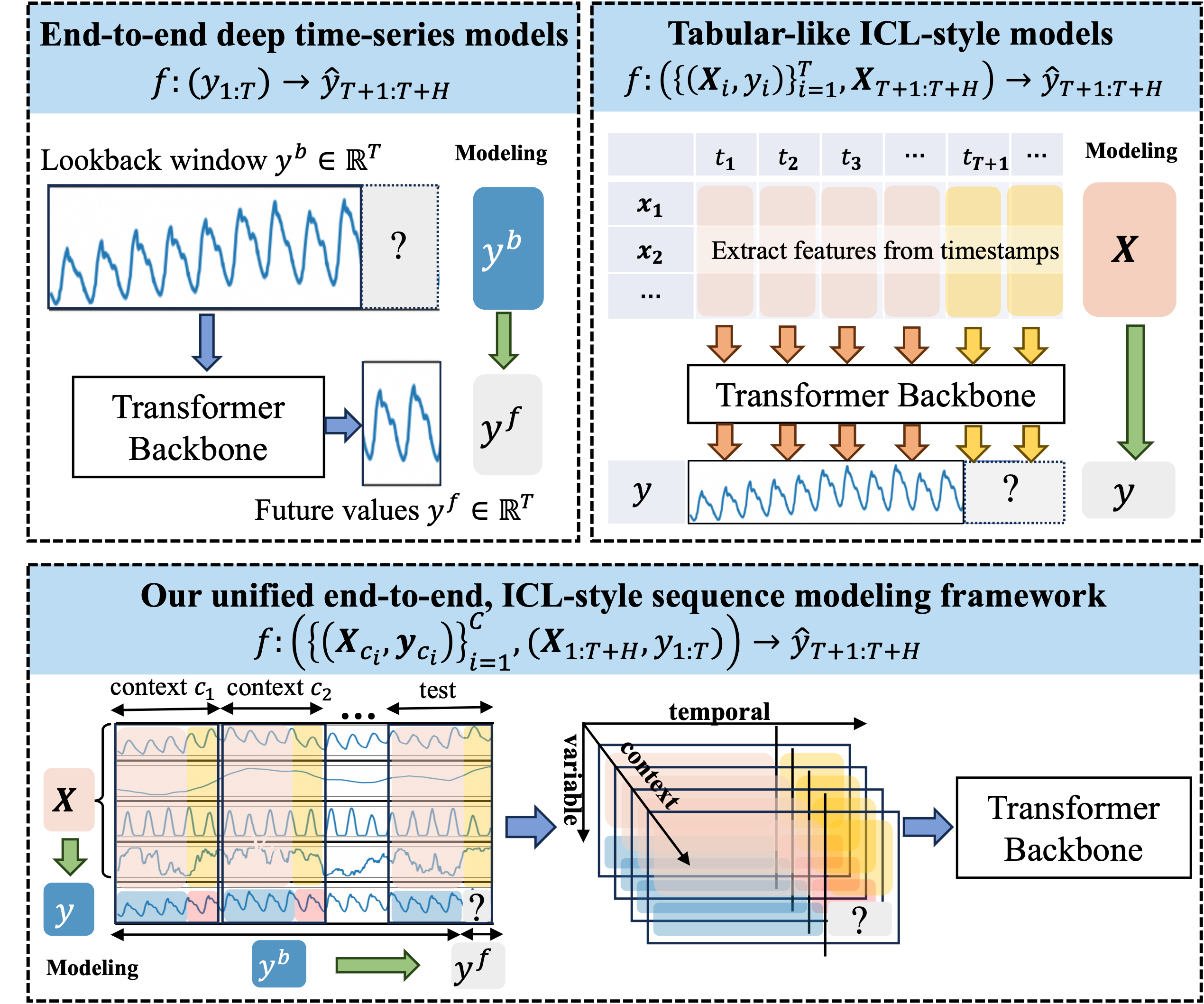}
    \caption{Three paradigms for time series forecasting: (a) End-to-end sequence models learn from raw histories but lack in-context adaptation at inference. (b) Tabular ICL approaches (e.g., TabPFN) perform ICL over feature-engineered representations. (c) Our unified approach (Baguan-TS) enables sequence-native ICL on raw multivariate inputs, attending over temporal, variables, and context for gradient-free adaptation.}
    \label{fig:fig1}
    \vspace{-0.8cm}
\end{figure}

Time series forecasting increasingly demands models that adapt swiftly to new tasks, remain robust under distribution shift, and operate efficiently in data-limited regimes. While Transformers have revealed the promise of in-context learning (ICL)---conditioning predictions on a small support set at inference time without gradient updates---most ICL-style approaches in forecasting rely on tabularization and hand-crafted features~\cite{hoo2025tabular}, limiting their ability to exploit the structure of raw sequences. On the other hand, end-to-end sequence models excel at learning representations directly from raw time series but typically lack ICL-style, gradient-free adaptation~\cite{ansari2024chronos,dasdecoder}. This disconnect motivates a framework that unifies end-to-end representation learning with ICL for time series data, so that a single model can both extract features from raw sequences and adapt on the fly.


We propose Baguan-TS, a general-purpose architecture that enables sequence-native ICL on raw multivariate time series without hand-crafted features.
It builds on a unified, episodic framework that learns to forecast from raw sequences given a support set, thereby removing dependence on feature engineering while bringing ICL's rapid adaptation into a sequence-native architecture. 
An overview of this framework is shown in Fig.~\ref{fig:fig1}. To instantiate the framework, we introduce a 3D Transformer that treats \textbf{temporal, variable}, and \textbf{context} dimensions as first-class axes of attention. Analogous to video-style modeling, the model builds multiscale representations over the temporal $\times$ variable plane and aligns support and query along the context axis via cross-attention. This design lifts the performance ceiling by discovering task-relevant structure directly from raw time series inputs and specializing predictions per episode through the provided context. In addition, the model admits a 2D inference mode by setting the context length to one, which functions as a strong complementary component in ensembles. In this way, it offers a robust fallback and diversity boost when integrating multiple predictors.

However, adopting a high-capacity 3D architecture introduces two main challenges we target:

\begin{description}
\item[Challenge 1: Locality Calibration.] Large-capacity models are prone to miscalibration and optimization instability, especially when dealing with a large 3D Transformer model. A lightweight, \emph{feature-agnostic} mechanism is needed to provide \emph{local}, episode-specific correction without manual features.
\item[Challenge 2: Output Oversmoothing in ICL.] In the pre\-sence of noisy, heterogeneous support examples, the model can oversmooth the spurious signals rather than extracting stable periodic rules. Effective ICL under these conditions requires a careful balance between \emph{denoising} (resisting noise and shift) and \emph{selection} (focusing on a compact, highly relevant subset of samples).
\end{description}
To improve calibration and stability, we propose a target-space retrieval-based forecast (Y-space RBfcst) local calibration module. By referencing nearest support targets within each episode, this feature-agnostic procedure provides a local, distribution-aware adjustment that complements the learned predictor. It improves calibration and stabilizes training and inference under limited per-task data and distribution shift, and integrates naturally with episodic ICL---offering a simple, retrieval-based bias/variance correction that scales with model capacity without relying on hand-crafted features.

To counter output oversmoothing, we introduce a mitigation strategy that integrates reliability-aware weighting of support examples with deliberately overfitting to some exact context sample in the support set. Counterintuitively, this concentrates attention on the few examples that matter for each query. This balance between denoising and selection reduces spurious correlations and improves few-shot robustness when training a large 3D architecture.

Together, these components form a unified framework that marries representation learning with ICL on raw sequences; a 3D Transformer instantiation that raises the performance  (and offers a practical 2D inference mode for ensemble complementarity); and two complementary mechanisms---context-overfitting mitigation and Y-space RBfcst calibration---that make such a high-capacity, ICL-enabled model robust and trainable in practice. The result is a practical route to bringing ICL's strengths to raw time series forecasting under realistic data and shift conditions.
We summarize our contributions as follows: 
\begin{itemize}
\item We propose Baguan-TS, a unified end-to-end framework that performs ICL directly on raw multivariate time series, without relying on feature engineering. It is instantiated as a 3D Transformer attending jointly over temporal, variable, and context axes,
and also supports a 2D inference mode.
On \texttt{fev-bench-cov} (30 covariate-aware tasks), Baguan-TS achieves the best average scaled quantile loss (SQL) and MASE with the highest win rate, reducing SQL versus TabPFN-TS by 4.8\% (Fig.~\ref{fig:cov_winRateSQL_SQL_fev}).
\begin{figure}[!t]
    \centering
    \includegraphics[width=1.0\linewidth]{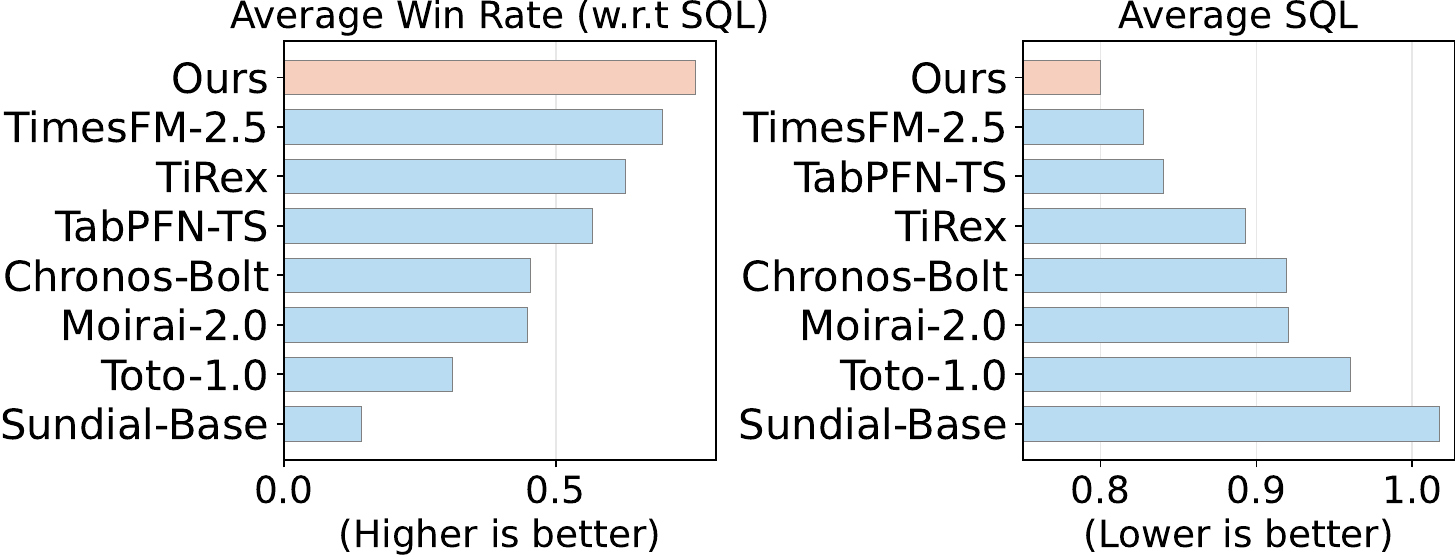}
    \caption{Probabilistic forecasting results on \texttt{fev-bench-cov} (30 covariate-aware tasks): Baguan-TS leads all baselines, achieving the highest average win rate and lowest average SQL.}
    \label{fig:cov_winRateSQL_SQL_fev}
\end{figure}

\item We develop a Y-space RBfcst local calibration module---feature-agnostic and episode-specific---that improves calibration, data efficiency, and scalability when training larger 3D Transformers. In our experiments, this module consistently improves overall forecasting accuracy and robustness under injected noise compared to training without retrieval.



\item We introduce a context-overfitting strategy that explicitly balances sample denoising and sample selection, stabilizing in-context learning in high-capacity models. The strategy consistently lowers training loss and restores periodic spike reconstruction, mitigating oversmoothing without harming trend accuracy.


\end{itemize}

\label{submission}

\section{Related Work}
We situate Baguan-TS within three lines of prior work, including end-to-end time series forecasting, large pretrained time series models, and in-context modeling. We summarize the key capability differences of Baguan-TS and these approaches in Table~\ref{tab:capability_comparison}.
\begin{table}[ht]
\centering \setlength{\tabcolsep}{0.5mm}
\caption{Capability comparison of representative time series forecasting approaches.}
\resizebox{.5\textwidth}{!}{%
\begin{tabular}{cccccc}
\toprule
\textbf{Models} &
\makecell{\textbf{Raw}\\\textbf{time}\\\textbf{sequence}} &
\makecell{\textbf{ICL}\\\textbf{(gradient-}\\\textbf{free)}} &
\makecell{\textbf{No hand-}\\\textbf{crafted}\\\textbf{features}} &
\makecell{\textbf{Cross-}\\\textbf{variable}\\\textbf{interaction}} &
\makecell{\textbf{Local}\\\textbf{retrieval-based}\\\textbf{calibration}} \\
\midrule
\makecell{End-to-end models\\(PatchTST, FedFormer)} & $\checkmark$ & $\times$ & $\checkmark$ &$\times$ & $\times$ \\
\addlinespace
\makecell{Large pretrained models\\(Chronos, TimesFM, Tirex)} & $\checkmark$ & $\times$ & $\checkmark$ & $\times$ & $\times$ \\
\addlinespace

\makecell{Tabular ICL-style models\\(TabPFN-TS)} & $\times$ & $\checkmark$ & $\times$ & $\checkmark$ & $\times$ \\
\addlinespace

\textbf{Ours (Baguan-TS)} & $\checkmark$ & $\checkmark$ & $\checkmark$ & $\checkmark$ & $\checkmark$ \\
\bottomrule
\end{tabular}
}
\label{tab:capability_comparison}
\end{table}

\noindent\textbf{Time Series Forecasting.} Deep neural networks dominate modern time series forecasting, with univariate models focusing on single sequences~\citep{rangapuram2018deep,salinas2020deepar,oreshkinn} and multivariate models jointly modeling many correlated series using Transformers~\citep{wu2021autoformer,zhou2022fedformer,patchtstnietime,liuitransformer,wang2024card,chen2024multi,zhou2023one} and other architectures~\citep{sen2019think,zhou2022film,jin2022multivariate,wangtimemixer,hu2024attractor,qi2024pdetime}. While highly effective on their training domains, these models typically require task-specific retraining, limiting their adaptability and motivating more general-purpose approaches.

\noindent\textbf{Large Pretrained Time Series Models.} Recent work pretrains large sequence models for time series~\citep{woo2024unified,goswamimoment,ansari2024chronos,dasdecoder,rasul2023lagllama,liutimer,shi2024time}, often with Transformer-based architectures and diverse time series data. However, their zero- and few-shot performance, especially for multivariate forecasting with complex cross-channel dependencies, often still trails specialized models.

\noindent\textbf{In-Context Modeling.} A complementary line of work treats forecasting as conditional generation and relies on in-context (few-shot) adaptation~\citep{hoo2025tabular,NEURIPS2024_97dc07f1,zhu2023xtab,NEURIPS2023_0731f0e6,pmlr-v206-hegselmann23a}. While competitive with strong tabular learners, these methods typically rely on hand-crafted features~\citep{chen2016xgboost,ke2017lightgbm}, limiting their practical usages in diverse applications. In contrast, the end-to-end sequence models are preferred by directly capturing temporal structures and cross-channel interactions.


\section{Baguan-TS}
\subsection{Problem Formulation}
In this paper, we consider univariate time series forecasting with known covariates. Let $\{\mathrm{y}_t\}_{t=1}^T$ be the observed series and $\{\mathbf{x}_t\}_{t=1}^{T+H}$, where $\mathbf{x}_t\in\mathbb{R}^M$, the associated $M$-dimensional covariates, available for both the history and the forecast horizon $H$ (e.g., calendar effects or known exogenous inputs such as weather forecasts). The task is to predict the future values $\{\mathrm{y}_t\}_{t=T+1}^{T+H}$.

For a test instance, let $\mathbf{y}^b = (\mathrm{y}_{1:T})\in\mathbb{R}^T$ denote the lookback window, $\mathbf{y}^f = (\mathrm{y}_{T+1:T+H})\in\mathbb{R}^H$ the future values, and $\mathbf{X}\in\mathbb{R}^{(T+H)\times M}$ the covariates over $t=1,\dots,T+H$. We assume an underlying mapping $\mathbf{y}^f = f^*(\mathbf{y}^b,\mathbf{X})$. 

In the in-context learning setting, we are given a context set
$\mathcal{D}_c = \{ (\mathbf{X}_j,\mathbf{y}_j^b,\mathbf{y}_j^f) \}_{j=1}^{C}$
of $C$ input–output examples. 
At test time, we adapt to a target instance by leveraging an inference function $g$ to approximate 
$f^*$, where the approximation is conditioned on 
$\mathcal{D}_c$ to minimize the prediction error
$l(\mathbf{y}^f, g(\mathbf{X},\mathbf{y}^b,\mathcal{D}_c))$.
Our objective is to learn a universal function $g$
that can adaptively fit a wide range of different underlying mappings 
$f^*$ across diverse tasks.

For convenience, we concatenate covariates and the full series into
$\mathbf{Y}_i = [\mathbf{X}_i,\mathbf{y}_i]\in\mathbb{R}^{(T+H)\times (M+1)}$,
where $\mathbf{y}_i\in\mathbb{R}^{T+H}$ stacks $(\mathbf{y}_i^b,\mathbf{y}_i^f)$.
Stacking all context and target samples yields a tensor
$\mathcal{Y}\in\mathbb{R}^{(C+1)\times (T+H)\times (M+1)}$, whose first $C$
slices correspond to $\mathcal{D}_c$ and the last slice to the target instance.
For the target slice, $\mathbf{y}^f$ is unknown and stored as a mask, while it is observed for the context slices.

\subsection{Architecture}
\begin{figure}[!b]
    \centering
    \includegraphics[width=1\linewidth]{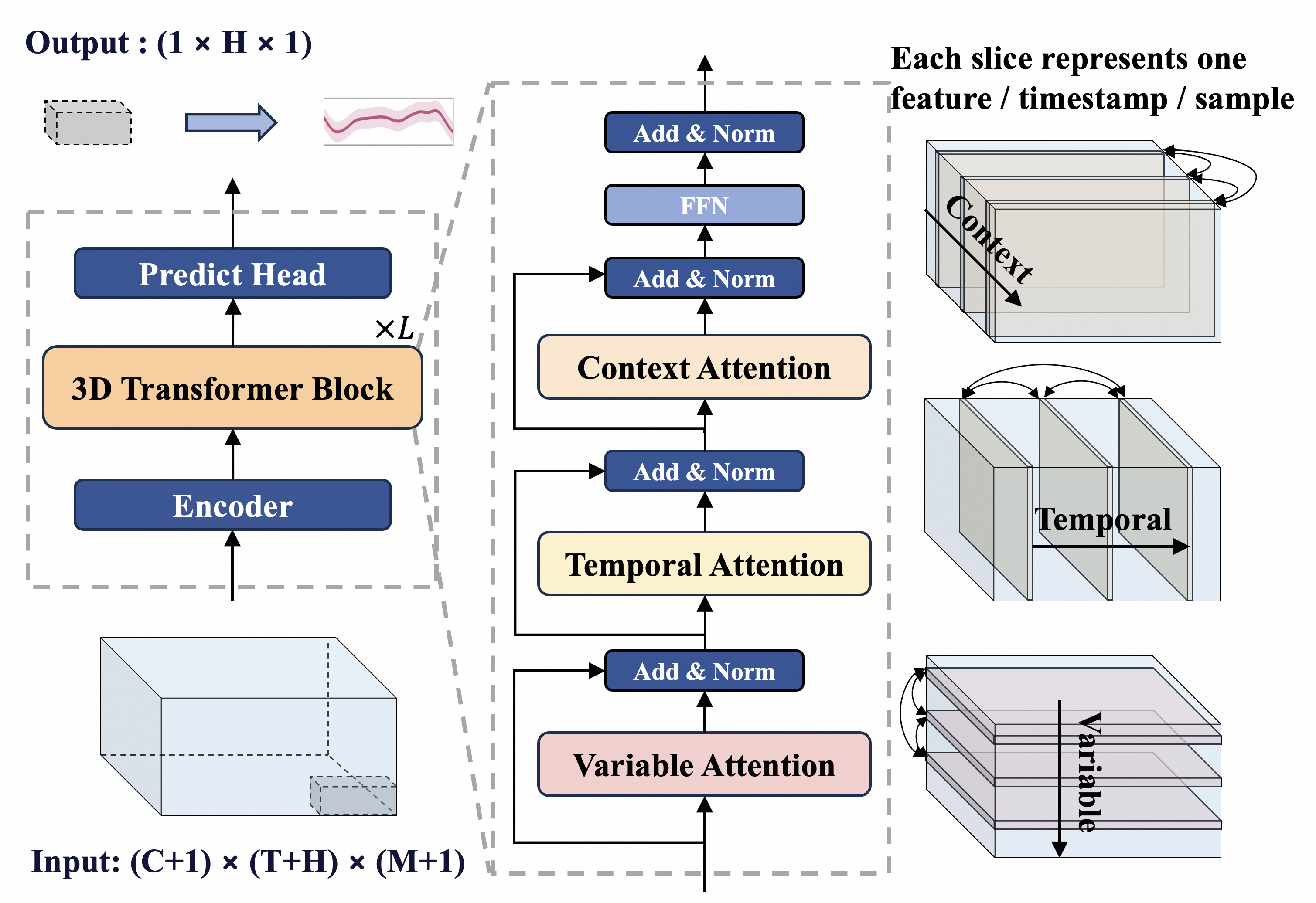}
    \caption{Overall architecture of Baguan-TS. The input tensor $\mathcal{Y}\in\mathbb{R}^{(C+1)\times (T+H)\times (M+1)}$ is encoded into patch tokens, then iteratively processed by stacked 3D Transformer blocks performing variable, temporal, and context attention, and finally mapped by a prediction head to produce the forecasting outputs $\mathbf{y}^f\in\mathbb{R}^H$.}
    \label{fig:overall_arch}
\end{figure}

Fig.~\ref{fig:overall_arch} shows the overall architecture of Baguan-TS, with its components detailed in the following subsections.

\subsubsection{Patching and Tokenization}
Given the tensor $\mathcal{Y}\in\mathbb{R}^{(C+1)\times (T+H)\times (M+1)}$,
we apply an encoder based on temporal patching and random Fourier features~\cite{rahimi2007random}.
Each variable in each slice is split along the temporal axis into $S$ non-overlapping
patches of length $P$ (zero-padding the tail if needed), reducing sequence length while preserving local structure.

For each covariate patch $\mathbf{q}\in\mathbb{R}^{P}$, we map it to
$\mathbf{e}=[\cos(\boldsymbol{\phi});\sin(\boldsymbol{\phi})]\in\mathbb{R}^{D}$,
with $\boldsymbol{\phi}{=}\mathbf{Wq}{+}\mathbf{b}$, where
$\mathbf{W}$ and $\mathbf{b}$ are shared learnable parameters.
For the target series, we zero-fill the unknown future and use a mask $\mathbf{m}\in\{0,1\}^P$ to indicate observed positions. Its embedding $\mathbf{e}^f$ adds an indicator term $\mathbf{Vm}$ to the Fourier features to distinguish the history from the forecast horizon. 
The resulting token tensor has shape $(C+1)\times S\times (M+1)\times D$.

\subsubsection{3D Transformer Block}
The core of our architecture is the 3D Transformer block, an encoder-only module that jointly models correlations across the variable, temporal, and context dimensions. Given the encoded tokens, it applies a sequence of specialized attention layers over these axes, followed by a feed-forward network (FFN). Residual connections and layer normalization follow each attention layer and the FFN.

Compared to conventional 1D or 2D Transformer blocks~\cite{hoo2025tabular, ansari2024chronos} that operate on flattened or reduced-dimensional embeddings, our block preserves a structured three-dimensional layout. This factorized design provides a stronger inductive bias while retaining the flexibility of the standard Transformer.

Formally, let $\mathbf{Z} \in \mathbb{R}^{(C+1) \times S \times (M+1) \times D}$ denote the token representation (contexts × temporal patches × variables × embedding dimension). All attention layers use standard multi-head self-attention (MHA)~\cite{vaswani2017attention} along different axes of this representation. We describe these stages in detail below.

\textbf{Temporal Attention.} This module learns temporal dependencies within each context to capture how patterns evolve over time. For each fixed context $c$ and variable $m$, we extract the slice {$\mathbf{T}_{c,m}=\mathbf{Z}_{c,:,m,:} {\in} \mathbb{R}^{S\times D}$} as the representation along the temporal axis. To model these dependencies, we apply MHA integrated with Rotary Position Embeddings (RoPE). By encoding relative phase information into the query and key vectors, RoPE allows the mechanism to naturally capture relative temporal distances and periodic patterns, which are crucial for time series forecasting.

\textbf{Variable Attention:} Instead of temporal sequences, this branch operates on the variable dimension {$\mathbf{V}_{c,s}=\mathbf{Z}_{c,s,:,:} \in \mathbb{R}^{(M+1)\times D}$} for each time step $s$. To account for distinct variable semantics, we augment the representation with a learnable variable-wise embeddings before performing MHA to capture cross-variable correlations.

\textbf{Context Attention:} This branch models relationships across different instances by taking the slice $\mathbf{C}_{s,m}=\mathbf{Z}_{:,s,m,:} \in \mathbb{R}^{(C+1)\times D}$. Unlike the temporal and variable branches, context attention omits positional encodings, as the context dimension typically lacks an inherent sequential ordering and instead focuses on global information sharing.

\begin{figure*}[t]
    \centering
    \subfloat[Different context organization strategies]{
        \includegraphics[width=0.71\linewidth]{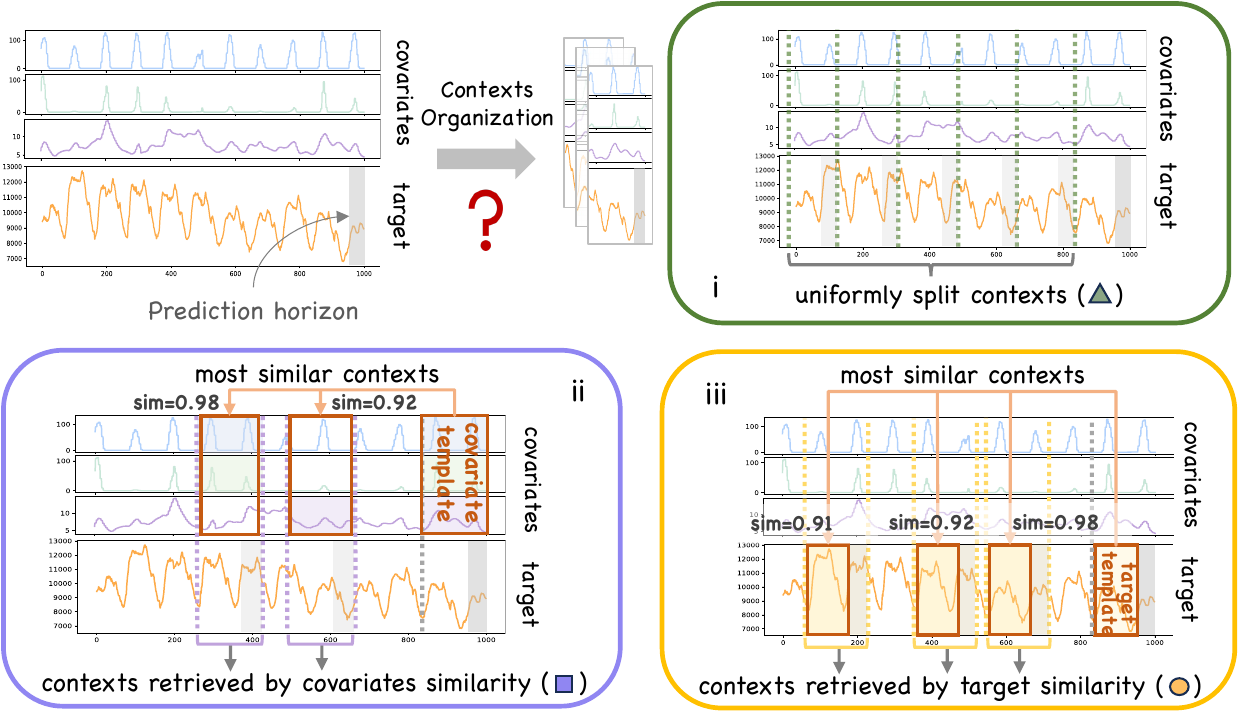}
        \label{fig:rag_illustration_1}
    }
    \hfill
    \subfloat[t-SNE visualization]{
        \includegraphics[width=0.25\linewidth]{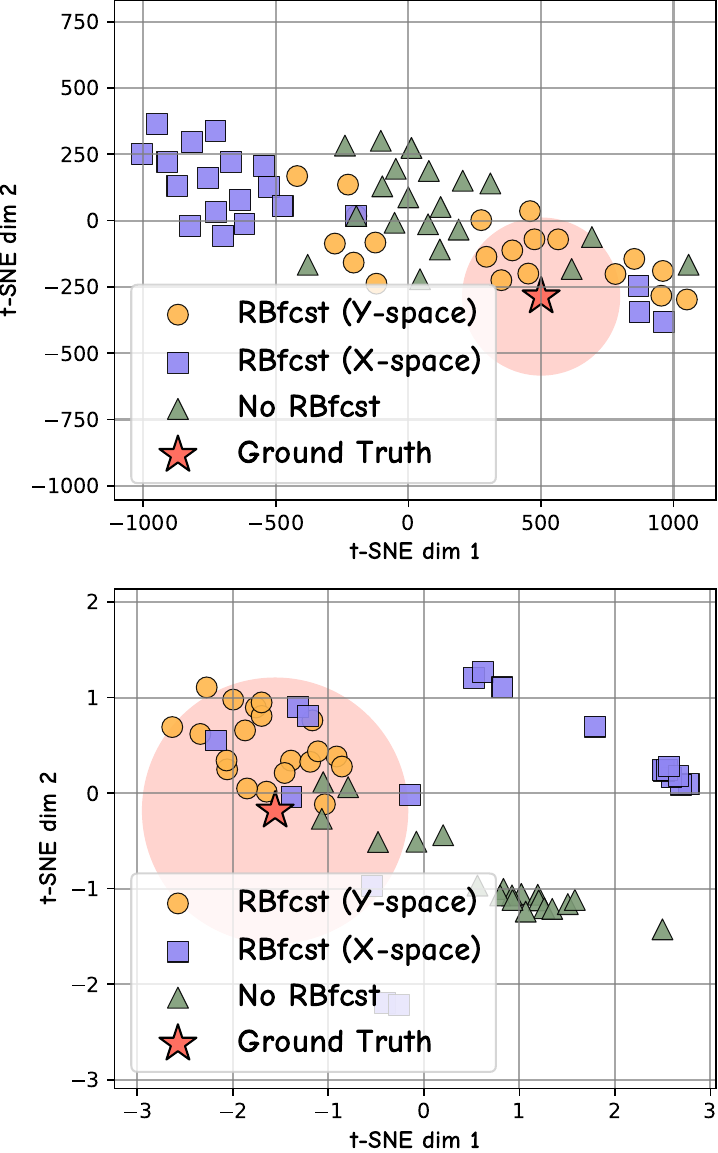}
        \label{fig:rag_illustration_2}
    }
    \caption{Context organization strategies and t-SNE visualization.
(a) Three context organization methods: (i) uniform splits (\textcolor{themeGreen}{green}); (ii) covariate-based retrieval in X-space (\textcolor{themePurple}{purple}, relies on feature engineering); (iii) target-based retrieval in Y-space (ours, \textcolor{themeYellow}{yellow}), which focuses on historical patterns and is feature-agnostic. Higher similarity scores indicate stronger contextual relevance.
(b) t-SNE plots of prediction horizons on \texttt{epf} (top) and \texttt{entsoe} (bottom) for three RBfcst variants. The ground truth (red star) lies closest to the Y-space RBfcst cluster (shaded), indicating it best captures the true pattern.}
\label{fig:rag_illustration}
\end{figure*}

\subsubsection{Prediction Head}
After a stack of 3D Transformer blocks, we obtain latent representations for future patches. A lightweight MLP head maps these features to per-horizon predictive distributions. Because inputs are $z$-normalized on the lookback (approximately zero mean and unit variance), most forecast values lie within a bounded range; we fix $[-10, 10]$ and uniformly discretize it into $K=5000$ bins. For each horizon step, the head outputs logits over these bins; applying softmax yields a probability vector $\mathbf{p}\in\mathbb{R}^K$. This provides a full probabilistic forecast with both point estimates (via the expected value over bin centers) and quantiles (via the empirical CDF).


\subsection{Training}
Our 3D Transformer operates on context–target episode sets. During training, we (i) build these contexts via retrieval-based forecasting; and (ii) stabilize in-context learning with a context-overfitting strategy. Implementation details on training data, loss functions, and inference settings are in Appendix~\ref{app:imple_details}.

\subsubsection{Retrieval-Based Forecasting}
To construct informative and scalable contexts, we adopt retrieval-based forecasting (RBfcst) to select informative contexts at scale. For a series of length $N$ and a context window $T+H$, there are $N-T-H+1$ candidate subsequences of length $T+H$, so full enumeration is memory-intensive. Instead, given the target lookback $\mathbf{y}^b \in \mathbb{R}^T$ as the query, we retrieve its $K_{\text{ctx}}$ nearest $T$-length subsequences from a sliding-window index over the historical region (to avoid leakage). Each retrieved $T$-length window, together with the subsequent $H$ points, forms one context of length $T+H$ (with covariates aligned). Distances are computed on $z$-normalized windows (e.g., Euclidean or cosine), and $K_{\text{ctx}}$ is chosen to balance relevance and memory.

To align training with this strategy and increase diversity, we first select a reference sequence of length $T+H$, sample a lookback of length $T$, and retrieve $2K_{\text{ctx}}$ nearest $T$-length subsequences using the same procedure. This leads to $2K_{\text{ctx}}+1$ candidates (including the selected reference), which is stacked into a tensor of shape $(2K_{\text{ctx}}+1)\times S\times(M+1)$. We then randomly choose $K_{\text{ctx}}$ slices to form a tensor of shape $K_{\text{ctx}}\times S\times(M+1)$ for that step. A subset of these $K_{\text{ctx}}$ slices is designated as the target (with its future masked), and the remainder serve as context. This randomized subsampling from a larger pool mimics imperfect retrieval, encouraging context diversity and robustness to retrieval noise.

Related tabular foundation models~\cite{thomas2024retrieval,xu2024mixture,gorishniy2024tabr} typically retrieve in covariate space (X-space). In contrast, we retrieve by similarity of the target history (Y-space), i.e., nearest neighbors of the lookback segment. For time series, the trajectory often summarizes the combined effects of all drivers, making Y-space retrieval more informative and feature-agnostic, without hand-crafted feature design.

To illustrate robustness, we compare three context-organization strategies in Fig.~\ref{fig:rag_illustration}: (i) uniform splitting (green), where windows are taken at equal intervals; (ii) X-space retrieval (purple), guided by covariate similarity and feature engineering; and (iii) Y-space retrieval (yellow, ours), which focuses on historical patterns most similar to the target. Higher similarity scores indicate stronger contextual relevance. t-SNE plots on \texttt{epf} (top) and \texttt{entsoe} (bottom) show that horizons retrieved in Y-space cluster near the ground truth (red star), suggesting this strategy best captures the true pattern.

\subsubsection{Context-Overfitting Strategy}
\begin{figure}[!hb]
    \vspace{-12pt}
    \centering
    \subfloat[Original design\label{fig:icl_ori}]{
        \includegraphics[width=0.45\linewidth]{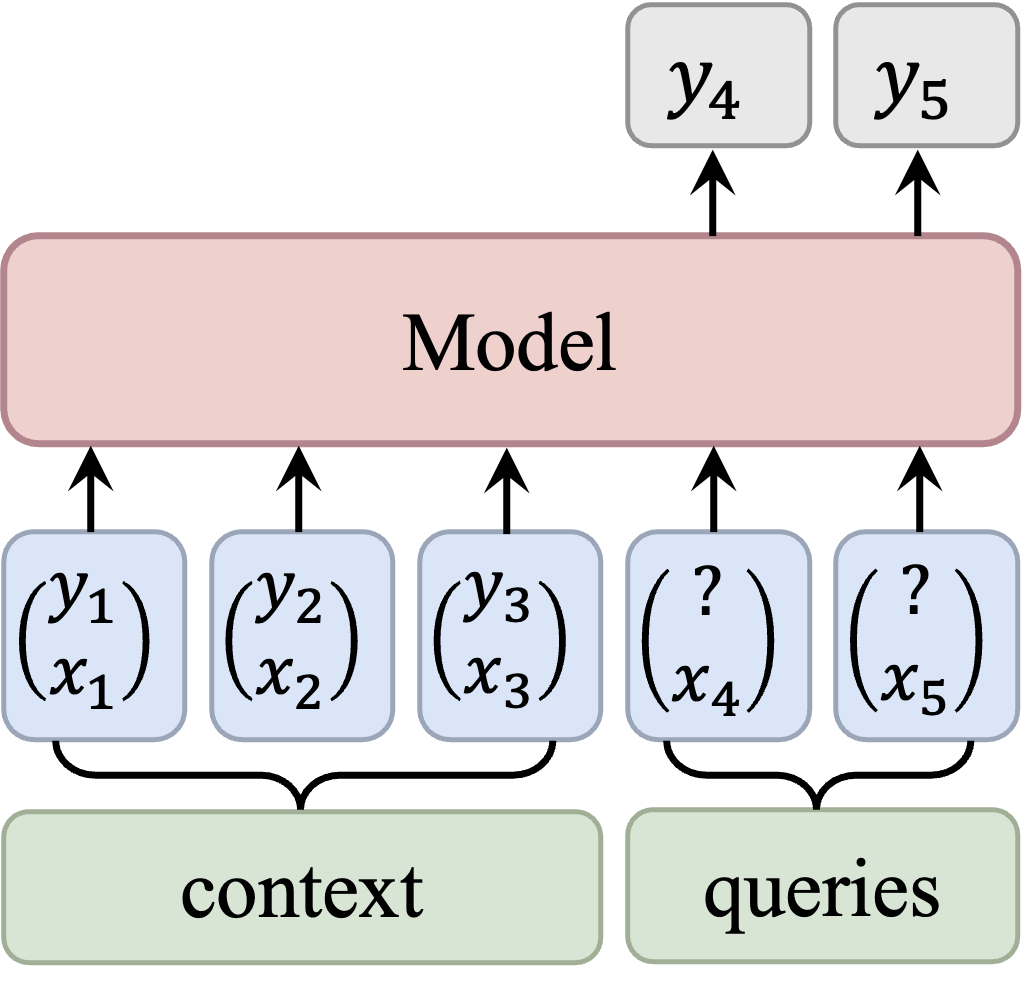}
    }\hfill
    \subfloat[Duplicate-context design\label{fig:icl_new}]{
        \includegraphics[width=0.45\linewidth]{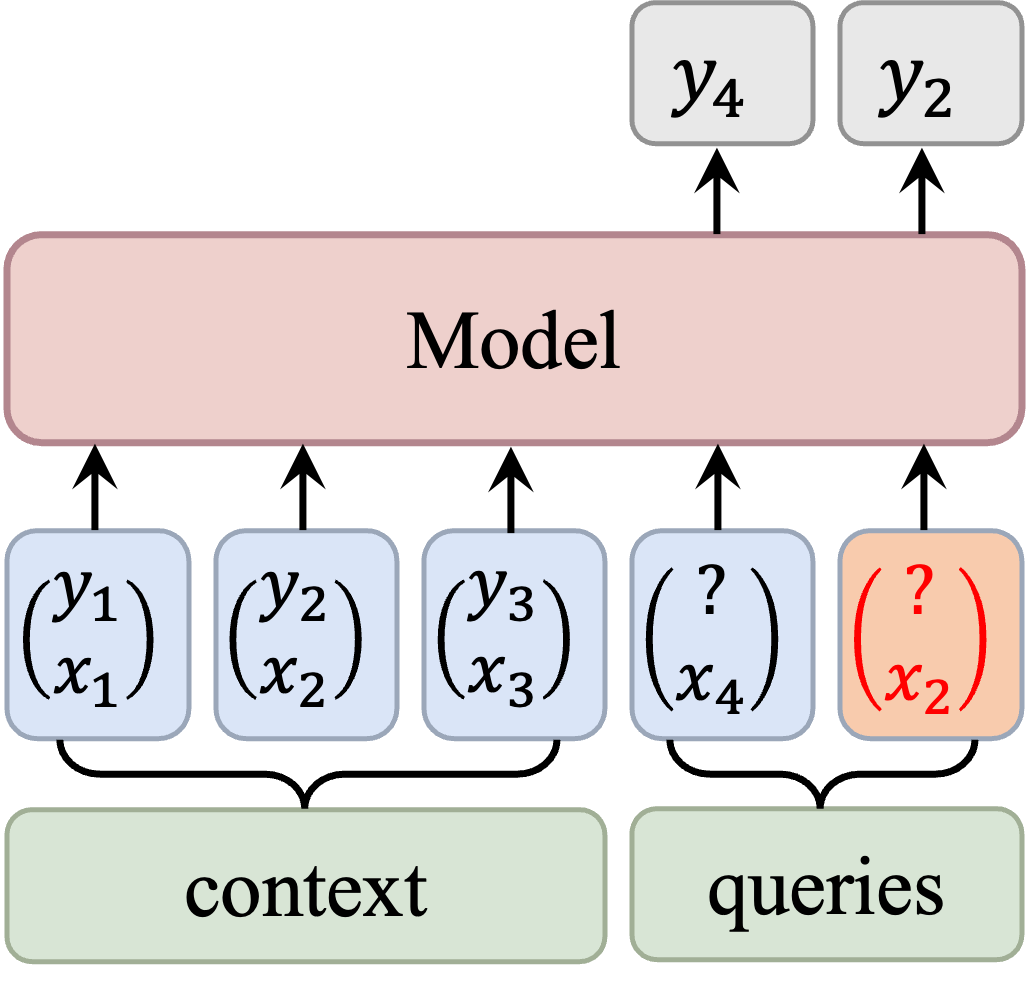}
    }
    \caption{Illustration of the context-overfitting strategy. (a) Original design, where the model forecasts query targets from retrieved context episodes. (b) Duplicate-context design: a short segment from one context slice is copied into a query, and the model is trained to retrieve the matching context and reconstruct its targets.}
    \label{fig:context-overfitting}
\end{figure}
\begin{figure}[!t]
    \centering
    \includegraphics[width=\linewidth]{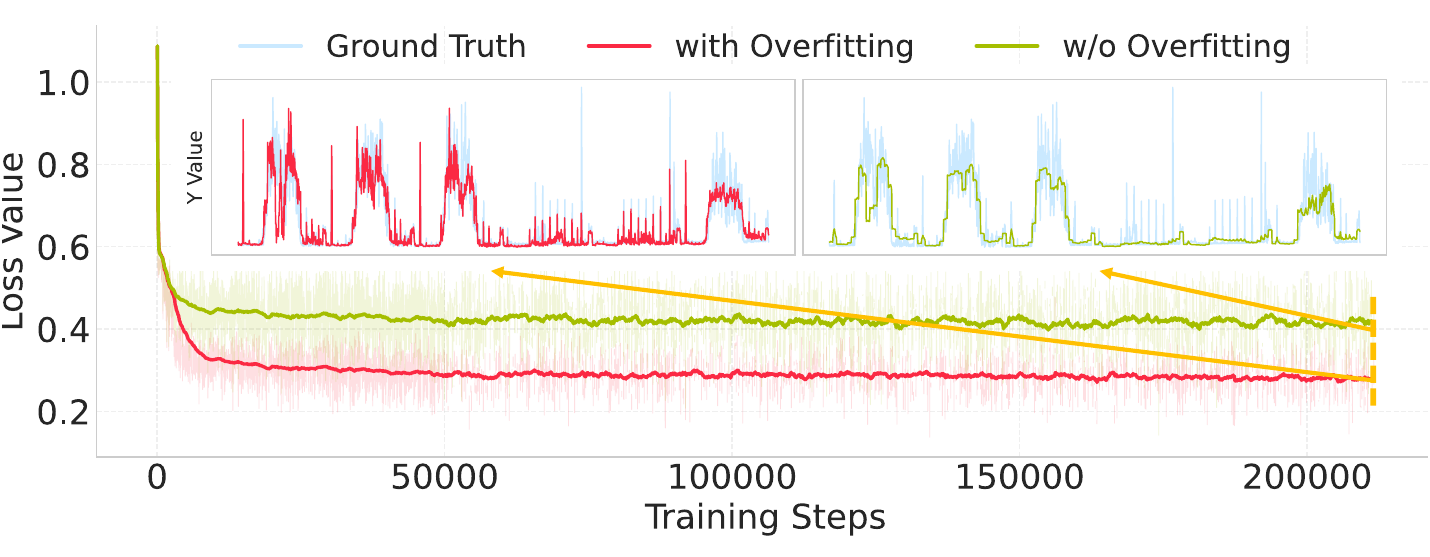}
    \caption{Effect of the context-overfitting strategy. Main: training loss curves for the baseline model (green) and our context-overfitting strategy (red). Insets: example forecasts compared with ground truth (blue). The baseline oversmooths outputs and misses high-frequency spikes (right); our strategy keeps a low loss while recovering spike patterns by matching contextual templates (left).}
    \label{fig:illustration-of-overfitting}
    \vspace{-0.4cm}
\end{figure}

The direct application of in-context learning to time series forecasting reveals a critical limitation: during early training, models tend to predict smooth, low-frequency trends while systematically suppressing periodic spike signals. This happens even when identical spike patterns are clearly present in the context window, indicating that the model misinterprets such abrupt fluctuations as noise rather than valid signal components. Although this tendency may enhance robustness against irregular outliers in certain datasets, it severely undermines performance in time series domains where periodic spikes constitute essential features (e.g., physiological monitoring or demand surges). To mitigate this oversmoothing issue where the model fails to leverage contextual spike templates for query inference, we introduce a context-overfitting strategy. Concretely, we enrich the query with a short segment copied verbatim from one context slice and add an auxiliary self-retrieval objective: the model is required to identify and align the matching context segment and reconstruct the corresponding target values (see Fig.~\ref{fig:context-overfitting}, where we include a synthetic test sample $x_2$ whose lookback contains a motif duplicated from the context, and train the model to retrieve that context). This explicitly encourages template matching: model must identify and retrieve target values from the context when encountering identical patterns in the query. 
As demonstrated by the training dynamics and qualitative results in Fig.~\ref{fig:illustration-of-overfitting}, the baseline model underutilizes contextual templates and produces oversmoothed predictions, while our strategy enables accurate reconstruction of periodic spikes without compromising trend accuracy.


\subsection{Adaptive Inference}
A key property of the 3D Transformer is its structural flexibility: it can operate either as a full sequence-native model or as a 2D tabular-style predictor.
Setting the context length $S=1$ collapses the temporal axis and restrict attention to feature and context dimensions. We find this 2D mode useful when temporal dependence is mostly captured by covariates or is weak (e.g., unreliable history under distribution shift). Thus, we ensemble the 2D and 3D modes in inference, which consistently improves SQL and reduces quantile calibration errors compared to using either mode alone.



\section{Experiments}
\subsection{Experiment Settings}
We compare Baguan-TS with recent time series foundation models:
Sundial-Base~\cite{liu2025sundial}, TabPFN-TS~\cite{hoo2025tabular}, TimesFM-2.0~\cite{dasdecoder}, TiRex~\cite{auer2025tirex}, Toto-1.0~\cite{cohen2025time}, Chronos-Bolt~\cite{ansari2024chronos}, and Moirai-2.0~\cite{liu2025moirai}. 
Among them, only TabPFN-TS natively supports covariates; models without covariate support are evaluated on target-only inputs.

We report mean absolute scaled error (MASE), weighted absolute percentage error (WAPE), scaled quantile loss (SQL), and weighted quantile loss (WQL), and the corresponding average win rates for each metric~\cite{shchur2025fev}, averaged over horizons and macro-averaged across datasets.
Further details of the datasets, experimental setup, and additional results are provided in Appendices~\ref{app:sec_ab} and~\ref{app:exp_fev}.

\subsection{Zero-Shot Forecasting}
\subsubsection{Time Series Forecasting with Covariates}
\textbf{Evaluation on public datasets.}
We first evaluate Baguan-TS on the \texttt{fev-bench} benchmark~\cite{shchur2025fev}. 
Focusing on time series tasks with covariate information, we select 30 representative tasks from the 100 available datasets that provide at least one known dynamic covariate, and refer to this subset as \texttt{fev-bench-cov}.
We evaluate both point and probabilistic forecasting performance, comparing Baguan-TS against leading time series foundation models. 
The results are reported in Fig.~\ref{fig:cov_winRateSQL_SQL_fev} and Figs.~\ref{fig:cov_winRate_fev}--\ref{fig:cov_skill_score_fev} (Appendix~\ref{app:exp_fev_cov}).
As shown in Table~\ref{tab:fev_cov_ranking}, Baguan-TS consistently outperforms both univariate baselines, including TiRex and Toto-1.0, and models specifically designed for covariate-informed forecasting, such as TabPFN-TS. This demonstrates Baguan-TS can effectively leverage both historical target series and future covariates for accurate predictions.

\begin{table}[htbp]
\centering
\caption{Average results on \texttt{fev-bench-cov}.
The best results are highlighted in \textbf{bold}, and the second-best results are \underline{underlined}.}
\label{tab:fev_cov_ranking}\setlength{\tabcolsep}{4mm}
\resizebox{\linewidth}{!}{
\begin{tabular}{lcccc}
\toprule
Model & SQL & MASE & WAPE & WQL \\
\midrule
Chronos-Bolt & 0.9190 & 1.1233 & 0.2702 & 0.2181 \\
Moirai-2.0 & 0.9202 & 1.1233 & 0.2750 & 0.2226 \\
Sundial-Base & 1.0172 & 1.1449 & 0.2813 & 0.2498 \\
TimesFM-2.5 & \underline{0.8276} & \underline{1.0122} & 0.2592 & 0.2095 \\
TabPFN-TS & 0.8404 & 1.0430 &\textbf{ 0.2015} & \textbf{0.1643} \\
TiRex & 0.8927 & 1.1081 & 0.2753 & 0.2233 \\
Toto-1.0 & 0.9600 & 1.1755 & 0.2737 & 0.2248 \\
\midrule
Ours  & \textbf{0.7997} & \textbf{0.9857} & \underline{0.2173} & \underline{0.1724} \\
\bottomrule
\end{tabular}
}
\end{table}

\textbf{Evaluation on real-world applications.}
We further evaluate Baguan-TS on 27 real-world datasets with historical and future covariates (e.g., weather,
calendar). Table~\ref{tab:cov_real} reports averages over all datasets.
Among the baselines, only TabPFN-TS and Baguan-TS use covariates; the others operate on the target series only. Baguan-TS achieves the best overall SQL, MASE, WAPE, and WQL, improving over TabPFN-TS and all non-covariate baselines, indicating more effective use of contextual signals. Detailed dataset descriptions and experimental results are provided in the Appendix~\ref{app:exp_fev_cov}.



\begin{table}[htbp]
    \centering
    \caption{Average results on real-world application datasets. The best results are highlighted in \textbf{bold}, and the second-best results are \underline{underlined}.}
    \label{tab:cov_real}\setlength{\tabcolsep}{4mm}
    \resizebox{\linewidth}{!}{
    \begin{tabular}{lcccc}
        \toprule
        Model & SQL & MASE & WAPE & WQL \\
        \midrule
        Chronos-Bolt   & 0.5101 & 0.6350 & 0.0976 & 0.0783 \\
        Moirai-2.0     & 0.7139 & 0.8875 & 0.1031 & 0.0860 \\
        Sundial-Base   & 0.7486 & 0.9042 & 0.1133 & 0.0981 \\
        TabPFN-TS      & \underline{0.4663} & \underline{0.6115} & \underline{0.0688} & \underline{0.0537} \\
        TimesFM-2.0    & 0.8418 & 0.9278 & 0.1093 & 0.0954 \\
        TiRex          & 0.5710 & 0.7094 & 0.0975 & 0.0784 \\
        Toto-1.0       & 0.8414 & 1.0566 & 0.1269 & 0.1034 \\
        \midrule
        Ours           & \textbf{0.3974} & \textbf{0.4956} & \textbf{0.0597} & \textbf{0.0490} \\
        \bottomrule
    \end{tabular}}
\end{table}


\subsubsection{Univariate Time Series Forecasting}

\begin{figure}
    \centering
    \includegraphics[width=0.95\linewidth]{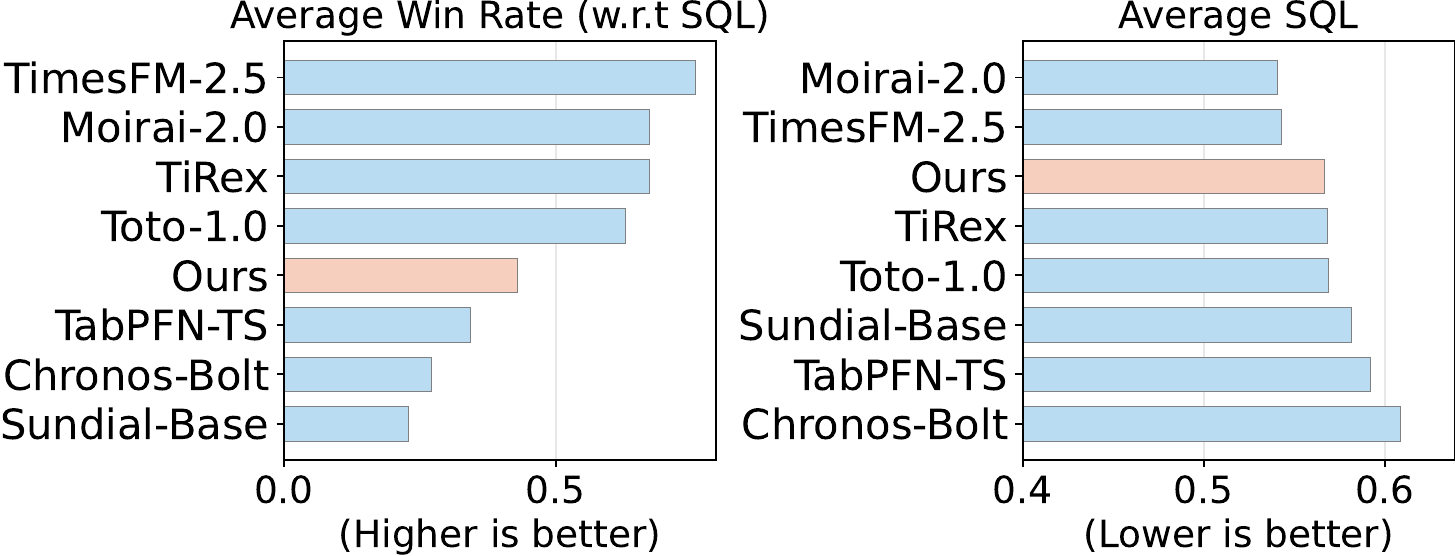}
    \caption{Evaluation on \texttt{fev-bench-uni} (10 univariate/multivariate tasks). Baguan-TS achieves competitive probabilistic forecasting performance on tasks without covariates.}
    \label{fig:uni_winRateSQL_SQL}
\end{figure}

We next consider univariate time series forecasting without covariates, where Baguan-TS adapts by incorporating null or time-only feature columns. 
Specifically, we evaluated 10 tasks from the \texttt{fev-bench-mini} dataset~\cite{shchur2025fev}, where multivariate datasets are flattened into standard univariate forecasting problems; we refer to this subset as \texttt{fev-bench-uni}.
As shown in Fig.~\ref{fig:uni_winRateSQL_SQL}, Baguan-TS achieves competitive macro-averaged SQL across tasks even without covariates. 
Notably, it outperforms TabPFN-TS, which also natively supports covariates, on more than 50\% of these datasets; it also achieves SOTA performance in terms of macro-averaged WAPE and WQL (see Fig.~\ref{fig:uni_WAPE_WQL}, Appendix~\ref{app:exp_fev_uni}).
These results indicate strong macro-level stability and robustness on aggregate-sensitive metrics, highlighting system-level accuracy. 

\subsection{Ablation Study}

\subsubsection{Effect of RBfcst}
We first ablate the proposed RBfcst module on an energy task (\texttt{entsoe\_1h}) and a retail task (\texttt{hermes}). The baseline (w/o RBfcst) uses uniformly split contexts. We compare three retrieval variants: X-space (covariates), XY-space (covariates + targets), and our Y-space RBfcst (targets only). All variants use the average of cosine and $L_2$ distances for segment similarity.


As shown in Table~\ref{tab:rbfcst_comparison}, Y-space RBfcst consistently outperforms both X-space and XY-space variants, all of which significantly surpass the w/o RBfcst baseline.
This validates the critical role of RBfcst in ICL-based time series forecasting and confirms that historical target trajectories generally provide more informative context than covariates, without requiring sophisticated feature selection or aggregation mechanisms.
The performance advantage of Y-space RBfcst is particularly large on \texttt{entsoe\_1h}, whereas the gap is smaller on \texttt{hermes}, which has shorter lookback windows and sparser covariates.

\begin{table}[!t]
\centering
\caption{Probabilistic and point forecasting evaluation for different RBfcst variants on \texttt{entsoe\_1H} and \texttt{hermes} datasets.}
\label{tab:rbfcst_comparison}\setlength{\tabcolsep}{1mm}
\resizebox{\linewidth}{!}{
\begin{tabular}{lcccccccc}
\toprule
\multirow{2}{*}{\textbf{Method}} & 
\multicolumn{4}{c}{\texttt{entsoe\_1H}} & 
\multicolumn{4}{c}{\texttt{hermes}} \\
\cmidrule(lr){2-5} \cmidrule(lr){6-9}
& SQL & MASE & WAPE & WQL & SQL & MASE & WAPE & WQL \\
\midrule
w/o RBfcst      & 0.4284 & 0.5387 & 0.0328 & 0.0262 & 0.6607 & 0.8445 & 0.0032 & 0.0025 \\
X-space    & 0.4185 & 0.5178 & 0.0313 & 0.0256 & 0.6186 & 0.7973 & 0.0030 & 0.0023 \\
XY-space   & 0.4005 & 0.4972 & 0.0297 & 0.0241 & 0.6186 & 0.7972 & 0.0030 & 0.0023 \\
\midrule
Y-space    & 0.3859 & 0.4819 & 0.0287 & 0.0231 & 0.6185 & 0.7971 & 0.0030 & 0.0023 \\
\bottomrule
\end{tabular}
}
\end{table}
\begin{figure}
    \centering
    \includegraphics[width=0.99\linewidth]{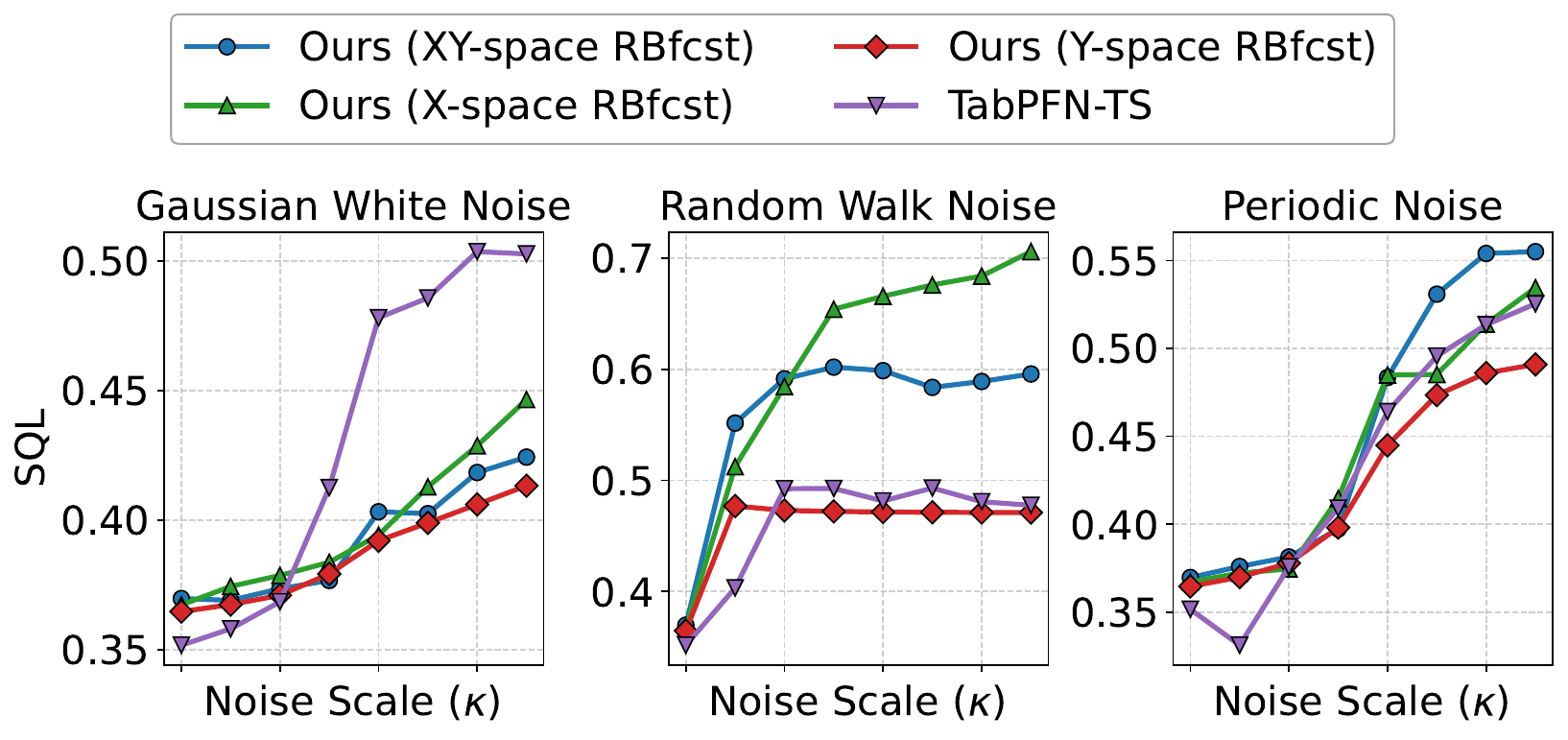}
    \caption{
    Robustness evaluation on \texttt{epf\_fr} under Gaussian white noise, random walk noise, and periodic noise, with $\kappa \in \{0,0.05,0.1,0.2,0.4,0.6,0.8,1.0\}$. 
    }
    \label{fig:injected_noise_epf_fr}
\end{figure}

To further evaluate the robustness of our local calibration module, we conduct noise-injection experiments on \texttt{epf\_fr} using three representative noise types: (i) Gaussian white noise (stationary); (ii) random walk noise (non-stationary with trend); and (iii) periodic noise (seasonal interference). The noise intensity is controlled by a scaling factor $\kappa$ relative to the series’ empirical standard deviation (see Appendix~\ref{app:exp_fev_noise} for details).
As shown in Fig.~\ref{fig:injected_noise_epf_fr}, although TabPFN-TS achieves the best SQL in the clean setting ($\kappa=0$), its performance degrades sharply under stationary Gaussian noise as $\kappa$ increases (especially when $\kappa \geq 0.1$). 
In contrast, Baguan-TS maintains stable performance across all noise types and intensities. Notably, the Y-space RBfcst variant exhibits the highest robustness, with minimal performance drop even at high noise levels. This shows that our feature-agnostic locality calibration effectively mitigates the impact of diverse external perturbations.

\subsubsection{Effect of Context-Overfitting}

We ablate the proposed context-overfitting strategy by retraining two models (a full model with context-overfitting and a baseline without it) from scratch for 220K steps on a dataset with clear daily and weekly seasonality and periodic spikes. Fig.~\ref{fig:illustration-of-overfitting} shows training losses and qualitative forecasts. With context-overfitting, the training loss is consistently lower—unsurprising, as the auxiliary self-retrieval task (recovering targets for a query containing a duplicated context segment) is easier than forecasting unseen targets. More importantly, without context-overfitting the forecasts are overly smooth and miss most spikes, while the full model captures much more high-frequency structure and accurately predicts the periodic spikes without distorting the trend.

\begin{figure}
\vspace{-0.2cm}
    \centering
    \subfloat[\tiny RMSE vs. Spike Sharpness]{
        \includegraphics[width=0.316\linewidth]{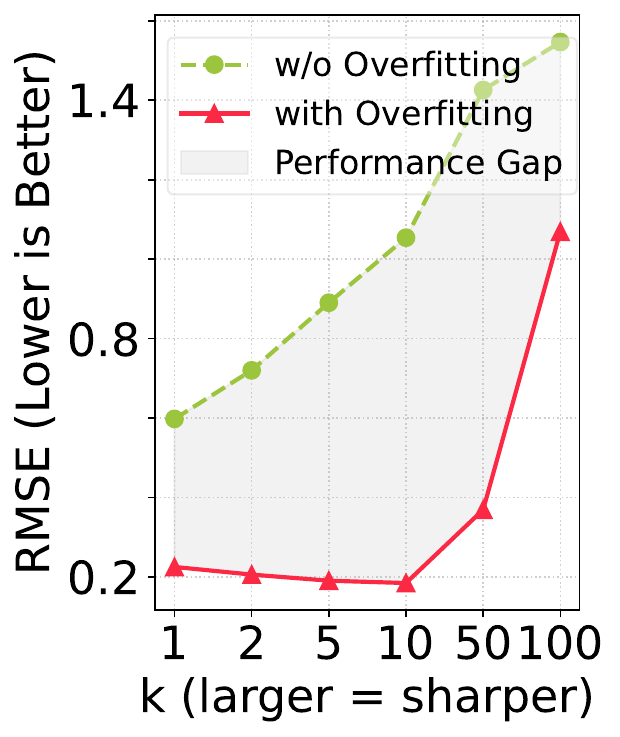} 
        \label{fig:rmse_spike_diff_k}
    }
    \hfill
    \subfloat[\tiny Attention Weight Map]{
        \includegraphics[width=0.614\linewidth]{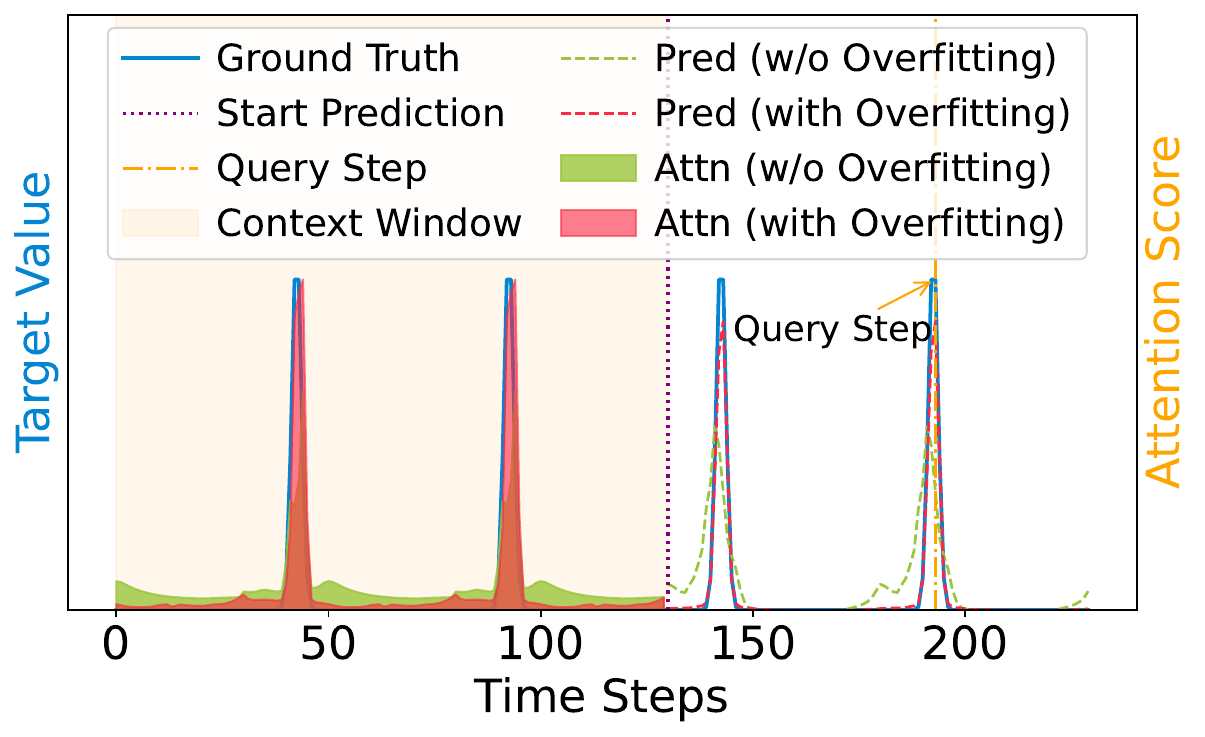} 
        \label{fig:attention_layer10_peak_partial}
    }
    \caption{Ablation of the context-overfitting strategy.
(a) The full model (red) consistently outperforms the baseline (green) on synthetic periodic spike data for all $k$, where larger $k$ corresponds to sharper spikes. (b) Attention map for a representative peak step (orange dotted line).}
\label{fig:ablation_context_overfitting}
\vspace{-0.2cm}
\end{figure}

To further assess robustness and clarify the mechanism, we compare these two models on synthetic periodic toy data. As shown in Fig.~\ref{fig:rmse_spike_diff_k}, the full model substantially reduces RMSE and more consistently recovers periodic spikes from the lookback window. To examine attention behavior, Fig.~\ref{fig:attention_layer10_peak_partial} visualizes a local attention map: with context-overfitting, the peak step attends strongly to historical peaks, whereas the baseline exhibits a pronounced dip at these locations, supporting our explanation of over-smoothing in vanilla attention.
Details of the data generation and additional visualizations are provided in Appendix~\ref{app:exp_contextoverfitting}.


\subsubsection{Different Inference Modes}

\begin{table}[!b]
\centering
\caption{Average results on \texttt{fev-bench-cov} under different inference modes of Baguan-TS model.}
\label{tab:inf_mode_ablation}\setlength{\tabcolsep}{4mm}
\resizebox{\linewidth}{!}{
\begin{tabular}{lcccc}
\toprule
Model & SQL & MASE & WAPE & WQL \\
\midrule
Ours (2D mode)  & 0.8769 & 1.0705 & 0.2333 & 0.1863 \\
Ours (3D mode)  & 0.8333 & 1.0220 & 0.2285 & 0.1830 \\
\midrule
Ours (Ensemble)  & 0.7997 & 0.9857 & 0.2173 & 0.1724 \\
\bottomrule
\end{tabular}
}
\end{table}

\begin{figure}[ht]
    \centering\vspace{-1em}
    \subfloat[Heatmap]{
        \includegraphics[width=0.4\linewidth]{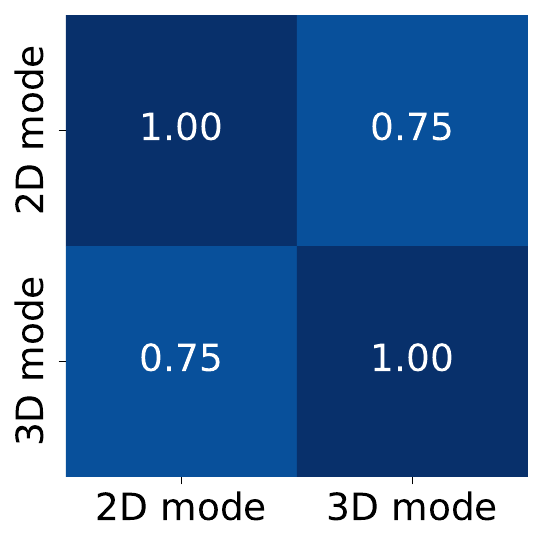}
        \label{fig:inf_mode_heatmap_all}
    }
    \hfill
    \subfloat[Residual Correlation]{
        \includegraphics[width=0.46\linewidth]{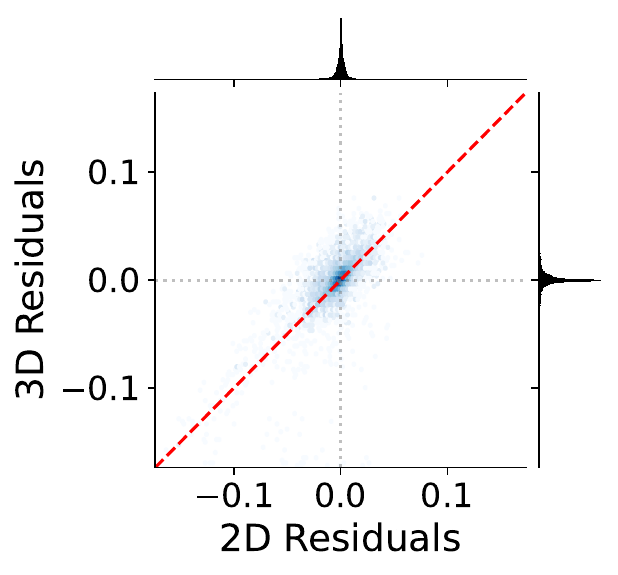}
        \label{fig:inf_mode_res_entsoe1h}
    }
    \caption{Residual correlation analysis of 2D and 3D inference modes. (a) Aggregated Pearson correlation across 30 tasks in \texttt{fev-bench-cov}. (b) Joint residual distribution for \texttt{entsoe\_1H}, where darker blue bins indicate higher data density. Points along the red dashed identity line ($y=x$) represent cases in which the errors in both modes are equal. Deviations from this diagonal, particularly in the second and fourth quadrants, highlight the complementary error structures between the two modes.}
\label{fig:inf_mode_residual}
\vspace{-0.2cm}
\end{figure}

\begin{figure}[!h]
    \centering
    \includegraphics[width=\linewidth]{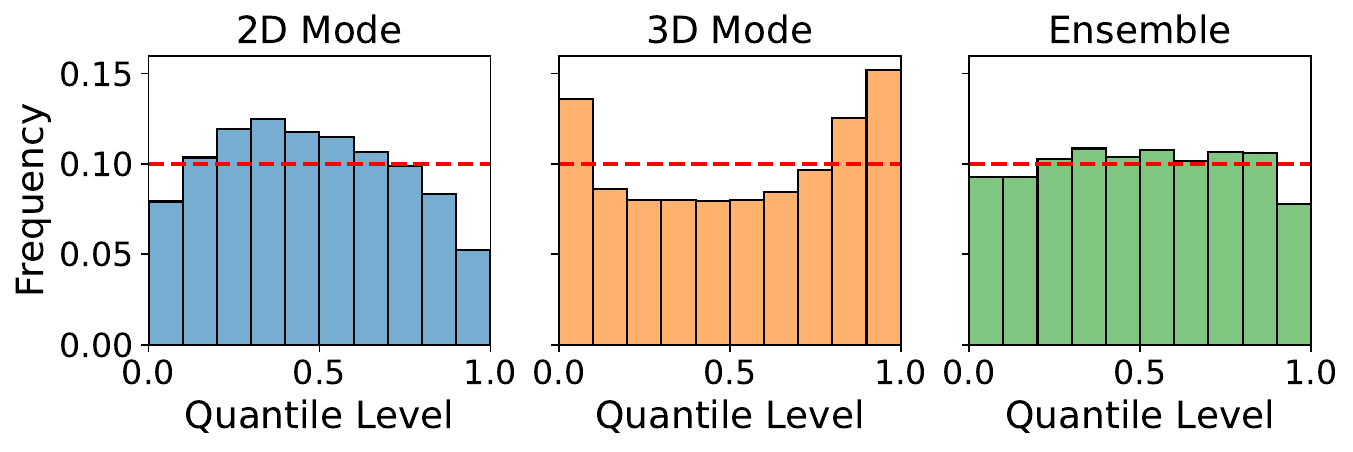}
    \caption{Calibration histograms of ground-truth quantile levels in predicted CDFs for different inference modes on \texttt{entsoe\_1H}. A well-calibrated model yields a uniform histogram (red dashed line). The 2D mode is under-confident (peaked), the 3D mode is over-confident (U-shaped), whereas the ensemble effectively cancels these biases and approaches the ideal uniform distribution.}
    \label{fig:inf_mode_calibration}
    \vspace{-0.4cm}
\end{figure}

In this section, we compare the 2D, 3D, and ensemble inference modes.
Averaged results on the \texttt{fev-bench-cov} are presented in Table~\ref{tab:inf_mode_ablation}. 
While the standalone 3D and 2D modes already deliver strong performance, their ensemble consistently improves all metrics for both point and probabilistic forecasting.
To understand these gains, we compute the Pearson correlation between 2D and 3D residuals pooled across all tasks (Fig.~\ref{fig:inf_mode_heatmap_all}), obtaining 0.75. This confirms that both modes are accurate and successfully capture the dominant temporal dynamics, while the correlation being well below 1.0 indicates diverse error structures.
To further visualize this diversity, we plot the joint residual distribution for the \texttt{entsoe\_1H} task in Fig.~\ref{fig:inf_mode_res_entsoe1h}, where points in the second and fourth quadrants highlight opposite-signed residuals, suggesting complementary errors.
We also evaluate the probabilistic calibration using calibration histograms (Fig.~\ref{fig:inf_mode_calibration})~\cite{gneiting2007probabilistic}. 
The x-axis shows ground-truth quantile levels under predicted CDFs; perfectly calibrated models yield a uniform histogram.
We observe opposite systematic biases in the base modes: the 2D mode shows a peaked histogram indicating under-confident predictions, whereas the 3D mode exhibits a U-shape distribution with over-confidence. The ensemble mode balances these errors, yielding a near-uniform histogram for better calibration.

\section{Conclusion}
We presented Baguan-TS, a unified framework that brings in-context learning to raw time series by combining a 3D Transformer over temporal, variable, and context axes with a practical 2D+3D inference scheme. A target-space retrieval-based calibration module and a context-overfitting strategy make this high-capacity architecture more stable, better calibrated, and less prone to oversmoothing. Across \texttt{fev-bench-cov}, \texttt{fev-bench-uni} and 27 in-house datasets, Baguan-TS consistently exhibits strong performance, showing that raw-sequence ICL can be both effective and robust in realistic forecasting settings.




\bibliography{main.bib}

@InProceedings{pmlr-v206-hegselmann23a,
  title = 	 {Tab{LLM}: Few-shot Classification of Tabular Data with Large Language Models},
  author =       {Hegselmann, Stefan and Buendia, Alejandro and Lang, Hunter and Agrawal, Monica and Jiang, Xiaoyi and Sontag, David},
  booktitle = 	 {Proceedings of the 26th International Conference on Artificial Intelligence and Statistics},
  pages = 	 {5549--5581},
  year = 	 {2023},
}

@article{NEURIPS2023_0731f0e6,
 author = {Dooley, Samuel and Khurana, Gurnoor Singh and Mohapatra, Chirag and Naidu, Siddartha V and White, Colin},
 journal = {Advances in Neural Information Processing Systems},
 pages = {2403--2426},
 title = {Forecast{PFN}: Synthetically-Trained Zero-Shot Forecasting},
 volume = {36},
 year = {2023}
}

@inproceedings{zhu2023xtab,
  title={Xtab: Cross-table pretraining for tabular transformers},
  author={Zhu, Bingzhao and Shi, Xingjian and Erickson, Nick and Li, Mu and Karypis, George and Shoaran, Mahsa},
  booktitle = {Proceedings of the International Conference on Machine Learning},
  year      = {2023}
}

@article{NEURIPS2024_97dc07f1,
 author = {Feuer, Benjamin and Schirrmeister, Robin Tibor and Cherepanova, Valeriia and Hegde, Chinmay and Hutter, Frank and Goldblum, Micah and Cohen, Niv and White, Colin},
 journal = {Advances in Neural Information Processing Systems},
 pages = {83430--83464},
 title = {Tune{T}ables: Context Optimization for Scalable Prior-Data Fitted Networks},
 volume = {37},
 year = {2024}
}

@article{hoo2025tabular,
  title={The tabular foundation model {T}ab{PFN} outperforms specialized time series forecasting models based on simple features},
  author={Hoo, Shi Bin and M{\"u}ller, Samuel and Salinas, David and Hutter, Frank},
  journal={arXiv preprint arXiv:2501.02945},
  year={2025}
}

@article{ansari2024chronos,
  title={Chronos: Learning the language of time series},
  author={Ansari, Abdul Fatir and Stella, Lorenzo and Turkmen, Caner and Zhang, Xiyuan and Mercado, Pedro and Shen, Huibin and Shchur, Oleksandr and Rangapuram, Syama Sundar and Arango, Sebastian Pineda and Kapoor, Shubham and others},
  journal={arXiv preprint arXiv:2403.07815},
  year={2024}
}

@inproceedings{woo2024unified,
  title={Unified training of universal time series forecasting transformers},
  author={Woo, Gerald and Liu, Chenghao and Kumar, Akshat and Xiong, Caiming and Savarese, Silvio and Sahoo, Doyen},
  booktitle={Proceedings
of the International Conference on Machine Learning},
  year={2024}
}

@article{vaswani2017attention,
  title={Attention is all you need},
  author={Vaswani, Ashish and Shazeer, Noam and Parmar, Niki and Uszkoreit, Jakob and Jones, Llion and Gomez, Aidan N. and Kaiser, {\L}ukasz and Polosukhin, Illia},
  journal={Advances in Neural Information Processing Systems},
  year      = {2017},
  volume    = {30},
}

@inproceedings{liutimer,
  title={Timer: Generative Pre-trained Transformers Are Large Time Series Models},
  author={Liu, Yong and Zhang, Haoran and Li, Chenyu and Huang, Xiangdong and Wang, Jianmin and Long, Mingsheng},
  booktitle={Proceedings of the 41st International Conference on Machine Learning},
  year={2024}
}

@inproceedings{wang2024card,
  title={{CARD}: Channel aligned robust blend transformer for time series forecasting},
  author={Wang, Xue and Zhou, Tian and Wen, Qingsong and Gao, Jinyang and Ding, Bolin and Jin, Rong},
  booktitle={Proceedings of the International Conference on Learning Representations},
  year={2024}
}

@inproceedings{chen2024multi,
  title={Pathformer: Multi-scale transformers with adaptive pathways for time series forecasting},
  author={Chen, Peng and Zhang, Yingying and Cheng, Yunyao and Shu, Yang and Wang, Yihang and Wen, Qingsong and Yang, Bin and Guo, Chenjuan},
  booktitle={Proceedings of the International Conference on Learning Representations},
  year={2024}
}

@inproceedings{patchtstnietime,
  title={A Time Series is Worth 64 Words: Long-term Forecasting with Transformers},
  author={Nie, Yuqi and Nguyen, Nam H and Sinthong, Phanwadee and Kalagnanam, Jayant},
  booktitle={Proceedings of the 11th International Conference on Learning Representations},
  year={2023}
}

@inproceedings{dasdecoder,
  title={A decoder-only foundation model for time-series forecasting},
  author={Das, Abhimanyu and Kong, Weihao and Sen, Rajat and Zhou, Yichen},
  booktitle={Proceedings of the 41st International Conference on Machine Learning},
  year={2024}
}

@article{rangapuram2018deep,
  title={Deep state space models for time series forecasting},
  author={Rangapuram, Syama Sundar and Seeger, Matthias W and Gasthaus, Jan and Stella, Lorenzo and Wang, Yuyang and Januschowski, Tim},
  journal={Advances in Neural Information Processing Systems},
  volume={31},
  year={2018}
}

@article{salinas2020deepar,
  title={{DeepAR}: Probabilistic forecasting with autoregressive recurrent networks},
  author={Salinas, David and Flunkert, Valentin and Gasthaus, Jan and Januschowski, Tim},
  journal={International Journal of Forecasting},
  volume={36},
  number={3},
  pages={1181--1191},
  year={2020}
}

@inproceedings{oreshkinn,
  title={N-BEATS: Neural basis expansion analysis for interpretable time series forecasting},
  author={Oreshkin, Boris N and Carpov, Dmitri and Chapados, Nicolas and Bengio, Yoshua},
  booktitle={Proceedings of the International Conference on Learning Representations},
  year={2020}
}

@inproceedings{liuitransformer,
  title={i{T}ransformer: Inverted Transformers Are Effective for Time Series Forecasting},
  author={Liu, Yong and Hu, Tengge and Zhang, Haoran and Wu, Haixu and Wang, Shiyu and Ma, Lintao and Long, Mingsheng},
  booktitle={Proceedings of the 12th International Conference on Learning Representations},
  year={2024}
}

@article{hu2024attractor,
  title={Attractor memory for long-term time series forecasting: A chaos perspective},
  author={Hu, Jiaxi and Hu, Yuehong and Chen, Wei and Jin, Ming and Pan, Shirui and Wen, Qingsong and Liang, Yuxuan},
  journal={Advances in Neural Information Processing Systems},
  volume={37},
  pages={20786--20818},
  year={2024}
}

@article{sen2019think,
  title={Think globally, act locally: A deep neural network approach to high-dimensional time series forecasting},
  author={Sen, Rajat and Yu, Hsiang-Fu and Dhillon, Inderjit S},
  journal={Advances in Neural Information Processing Systems},
  volume={32},
  pages={4837--4846},
  year={2019}
}

@inproceedings{wangtimemixer,
  title={{TimeMixer}: Decomposable Multiscale Mixing for Time Series Forecasting},
  author={Wang, Shiyu and Wu, Haixu and Shi, Xiaoming and Hu, Tengge and Luo, Huakun and Ma, Lintao and Zhang, James Y and Zhou, Jun},
  booktitle={Proceedings of the International Conference on Learning Representations},
  year={2024}
}

@article{jin2022multivariate,
  title={Multivariate time series forecasting with dynamic graph neural odes},
  author={Jin, Ming and Zheng, Yu and Li, Yuan-Fang and Chen, Siheng and Yang, Bin and Pan, Shirui},
  journal={IEEE Transactions on Knowledge and Data Engineering},
  volume={35},
  number={9},
  pages={9168--9180},
  year={2022},
  publisher={IEEE}
}

@article{qi2024pdetime,
  title={{PDET}ime: Rethinking Long-Term Multivariate Time Series Forecasting from the perspective of partial differential equations},
  author={Qi, Shiyi and Xu, Zenglin and Li, Yiduo and Wen, Liangjian and Wen, Qingsong and Wang, Qifan and Qi, Yuan},
  journal={arXiv preprint arXiv:2402.16913},
  year={2024}
}

@inproceedings{goswamimoment,
  title={{MOMENT}: A Family of Open Time-series Foundation Models},
  author={Goswami, Mononito and Szafer, Konrad and Choudhry, Arjun and Cai, Yifu and Li, Shuo and Dubrawski, Artur},
  booktitle={Proceedings of the 41st International Conference on Machine Learning},
  year={2024}
}

@inproceedings{zhou2022fedformer,
  title={{FEDformer}: Frequency enhanced decomposed transformer for long-term series forecasting},
  author={Zhou, Tian and Ma, Ziqing and Wen, Qingsong and Wang, Xue and Sun, Liang and Jin, Rong},
  booktitle={Proceedings of the International Conference on Machine Learning},
  year={2022}
}

@inproceedings{chen2016xgboost,
  title={{XGB}oost: A scalable tree boosting system},
  author={Chen, Tianqi and Guestrin, Carlos},
  booktitle={Proceedings of the 22nd {ACM} {SIGKDD} {I}nternational {C}onference on {K}nowledge {D}iscovery and Data Mining},
  pages={785--794},
  year={2016}
}

@article{ke2017lightgbm,
  title={Light{GBM}: A highly efficient gradient boosting decision tree},
  author={Ke, Guolin and Meng, Qi and Finley, Thomas and Wang, Taifeng and Chen, Wei and Ma, Weidong and Ye, Qiwei and Liu, Tie-Yan},
  journal={Advances in Neural Information Processing Systems},
  volume={30},
  year={2017}
}

@article{rasul2023lagllama,
    title={{Lag-Llama}: Towards Foundation Models for Time Series Forecasting}, 
    author={Kashif Rasul and Arjun Ashok and Andrew Robert Williams and Arian Khorasani and George Adamopoulos and Rishika Bhagwatkar and Marin Biloš and Hena Ghonia and Nadhir Vincent Hassen and Anderson Schneider and Sahil Garg and Alexandre Drouin and Nicolas Chapados and Yuriy Nevmyvaka and Irina Rish},
    year={2023},
    journal={arXiv preprint arXiv:2310.08278}
}

@article{wu2021autoformer,
    title={Autoformer: Decomposition transformers with auto-correlation for long-term series forecasting},
    author={Wu, Haixu and Xu, Jiehui and Wang, Jianmin and Long, Mingsheng},
    journal={Advances in Neural Information Processing Systems},
    volume={34},
    pages={22419--22430},
    year={2021}
}

@inproceedings{kim2022reversible,
    title={Reversible Instance Normalization for Accurate Time-Series Forecasting against Distribution Shift},
    author={Taesung Kim and Jinhee Kim and Yunwon Tae and Cheonbok Park and Jang-Ho Choi and Jaegul Choo},
    booktitle={Proceedings of the International Conference on Learning Representations},
    year={2022}
}

@article{shi2024time,
  title={{Time-MoE}: Billion-Scale Time Series Foundation Models with Mixture of Experts},
  author={Shi, Xiaoming and Wang, Shiyu and Nie, Yuqi and Li, Dianqi and Ye, Zhou and Wen, Qingsong and Jin, Ming},
  journal={Proceedings of the
International Conference on Learning Representations},
  year={2025}
}

@article{zhou2023one,
  title={One fits all: Power general time series analysis by pretrained {LM}},
  author={Zhou, Tian and Niu, Peisong and Sun, Liang and Jin, Rong and others},
  journal={Advances in Neural Information Processing Systems},
  volume={36},
  pages={43322--43355},
  year={2023}
}

@article{zhou2022film,
  title={Film: Frequency improved legendre memory model for long-term time series forecasting},
  author={Zhou, Tian and Ma, Ziqing and Wen, Qingsong and Sun, Liang and Yao, Tao and Yin, Wotao and Jin, Rong and others},
  journal={Advances in Neural Information Processing Systems},
  volume={35},
  pages={12677--12690},
  year={2022}
}

@article{thomas2024retrieval,
  title={Retrieval \& fine-tuning for in-context tabular models},
  author={Thomas, Valentin and Ma, Junwei and Hosseinzadeh, Rasa and Golestan, Keyvan and Yu, Guangwei and Volkovs, Maksims and Caterini, Anthony},
  journal={Advances in Neural Information Processing Systems},
  volume={37},
  pages={108439--108467},
  year={2024}
}

@article{xu2024mixture,
  title={Mixture of in-context prompters for tabular {PFN}s},
  author={Xu, Derek and Cirit, Olcay and Asadi, Reza and Sun, Yizhou and Wang, Wei},
  journal={Proceedings of the International Conference on Learning Representations},
  year={2025}
}

@inproceedings{
gorishniy2024tabr,
title={Tab{R}: Tabular Deep Learning Meets Nearest Neighbors},
author={Yury Gorishniy and Ivan Rubachev and Nikolay Kartashev and Daniil Shlenskii and Akim Kotelnikov and Artem Babenko},
booktitle={Proceedings of the 12th International Conference on Learning Representations},
year={2024},
}

@article{auer2025tirex,
  title={TiRex: Zero-Shot Forecasting Across Long and Short Horizons with Enhanced In-Context Learning},
  author={Auer, Andreas and Podest, Patrick and Klotz, Daniel and B{\"o}ck, Sebastian and Klambauer, G{\"u}nter and Hochreiter, Sepp},
  journal={arXiv preprint arXiv:2505.23719},
  year={2025}
}

@article{cohen2025time,
  title={This Time is Different: An Observability Perspective on Time Series Foundation Models},
  author={Cohen, Ben and Khwaja, Emaad and Doubli, Youssef and Lemaachi, Salahidine and Lettieri, Chris and Masson, Charles and Miccinilli, Hugo and Ram{\'e}, Elise and Ren, Qiqi and Rostamizadeh, Afshin and others},
  journal={arXiv preprint arXiv:2505.14766},
  year={2025}
}

@article{shchur2025fev,
  title={Fev-bench: A Realistic Benchmark for Time Series Forecasting},
  author={Shchur, Oleksandr and Ansari, Abdul Fatir and Turkmen, Caner and Stella, Lorenzo and Erickson, Nick and Guerron, Pablo and Bohlke-Schneider, Michael and Wang, Yuyang},
  journal={arXiv preprint arXiv:2509.26468},
  year={2025}
}

@article{chang2010arrhythmia,
  title={Arrhythmia {ECG} noise reduction by ensemble empirical mode decomposition},
  author={Chang, Kang-Ming},
  journal={Sensors},
  volume={10},
  number={6},
  pages={6063--6080},
  year={2010},
  publisher={Molecular Diversity Preservation International (MDPI)}
}

@article{kim2024extraction,
  title={Extraction of features for time series classification using noise injection},
  author={Kim, Gyu Il and Chung, Kyungyong},
  journal={Sensors},
  volume={24},
  number={19},
  pages={6402},
  year={2024},
  publisher={MDPI}
}

@article{wang2021model,
  title={A model for non-stationary time series and its applications in filtering and anomaly detection},
  author={Wang, Shixiong and Li, Chongshou and Lim, Andrew},
  journal={IEEE Transactions on Instrumentation and Measurement},
  volume={70},
  pages={1--11},
  year={2021},
  publisher={IEEE}
}

@article{gneiting2007probabilistic,
  title={Probabilistic forecasts, calibration and sharpness},
  author={Gneiting, Tilmann and Balabdaoui, Fadoua and Raftery, Adrian E},
  journal={Journal of the Royal Statistical Society Series B: Statistical Methodology},
  volume={69},
  number={2},
  pages={243--268},
  year={2007},
}

@article{liu2025sundial,
  title={Sundial: A family of highly capable time series foundation models},
  author={Liu, Yong and Qin, Guo and Shi, Zhiyuan and Chen, Zhi and Yang, Caiyin and Huang, Xiangdong and Wang, Jianmin and Long, Mingsheng},
  journal={arXiv preprint arXiv:2502.00816},
  year={2025}
}

@article{liu2025moirai,
  title={Moirai 2.0: When Less Is More for Time Series Forecasting},
  author={Liu, Chenghao and Aksu, Taha and Liu, Juncheng and Liu, Xu and Yan, Hanshu and Pham, Quang and Sahoo, Doyen and Xiong, Caiming and Savarese, Silvio and Li, Junnan},
  journal={arXiv preprint arXiv:2511.11698},
  year={2025}
}

@article{aksu2024gift,
  title={Gift-eval: A benchmark for general time series forecasting model evaluation},
  author={Aksu, Taha and Woo, Gerald and Liu, Juncheng and Liu, Xu and Liu, Chenghao and Savarese, Silvio and Xiong, Caiming and Sahoo, Doyen},
  journal={arXiv preprint arXiv:2410.10393},
  year={2024}
}

@article{rahimi2007random,
  title={Random features for large-scale kernel machines},
  author={Rahimi, Ali and Recht, Benjamin},
  journal={Advances in Neural Information Processing Systems},
  volume={20},
  year={2007}
}
\bibliographystyle{icml2026}

\clearpage
\appendix
\onecolumn
\section{Implement details} \label{app:imple_details}
\subsection{Model Details}
Baguan-TS uses a 3D Transformer-based architecture with hyperparameters summarized in
Table~\ref{tab:baguan_hparams}. The model has 22.4 million parameters.
\begin{table}[htbp]
\centering
\caption{Baguan-TS model architecture hyperparameters.}
\label{tab:baguan_hparams}\setlength{\tabcolsep}{7mm}
\resizebox{0.45\linewidth}{!}{
\begin{tabular}{ll}
\toprule
Parameter & Value \\
\midrule
Temporal patch size ($P$)                    & 8   \\
Number of 3D Transformer blocks ($L$)        & 12  \\
Number of heads                              & 6 \\
Embedding dimension ($D$)                    & 192 \\
Feed-forward dimension                       & 768 \\
Output dimension ($K$)                       & 5000 \\
\bottomrule
\end{tabular}}
\end{table}

During training, $C$, $T$, $H$, and $M$ are randomly sampled for each batch, up to fixed maximum values. We set $C_{\max}=50$ for the contexts, $T_{\max}=2048$ for the lookback length, $H_{\max}=192$ for the prediction horizon, and $M_{\max}=80$ for the number of covariates.
\subsection{Training Data}
Training data is critical to the performance of a foundation model. Baguan-TS is trained on a mixture of synthetic data and real-world benchmark data. The real-world data provide grounding in practical distributions and noise characteristics, while the synthetic data are designed to span a broad family of dynamics, covariate roles, and latent-factor structures that are hard to obtain exhaustively from any single corpus.

\textbf{Synthetic Data.} 
The diverse and task-specific semantics of covariates in different forecasting tasks make it difficult to handcraft a single, general-purpose synthetic data simulator.
Our key idea is that many covariate-aware forecasting problems can be viewed as regression with partially observed, time-correlated latent factors. Observed covariates are noisy or incomplete proxies for these latent drivers, while the target series combines autoregressive structure with nonlinear effects of both observed and latent inputs. Accordingly, we build a generic generator in which latent trajectories are drawn from a kernel dictionary, and transformed through a dynamic structural causal mechanism with a random MLP to induce nonlinear interactions. Only a subset of nodes is exposed as observed covariates and the remainder act as unobserved drivers; process and measurement noise, as well as optional regime shifts, are added to simulate different data characteristics. Within each task, the mechanism (graph, MLP weights, kernels, noise scales, exposure set) is fixed, whereas context and target instances draw independent latent realizations. 

\textbf{Real-World Data.} We include the GIFT-Eval pretraining corpus~\cite{aksu2024gift}
in our training set and augment each time step with a time index feature. 
To simulate distribution shifts and expose the model to regime changes, we concatenate multiple normalized series into longer sequences and randomly sample contiguous subsequences as training examples.

\subsection{Training Loss}
We use the continuous ranked probability score (CRPS) as the training loss. For a predictive CDF $F$ and observation $\mathrm{y}$, 
$\text{CRPS}(F,\mathrm{y}) = \int_{-\infty}^{\infty} (F(\mathrm{z})-\mathbb{I}(\mathrm{z}\ge \mathrm{y}))^2 d\mathrm{z}$. 
When the predictive distribution is represented by $K$ bins with centers $\mathrm{h}_i$, we collect the probabilities into a vector
$\mathbf{p} = (\mathrm{p}_1,\ldots,\mathrm{p}_K)^\top \in \mathbb{R}^K$ with $\sum_i \mathrm{p}_i = 1$, the corresponding discrete form of each time step is
\begin{align}
    \text{CRPS} = \sum_{i=1}^{K} \mathrm{p}_i |\mathrm{h}_i-\mathrm{y}|-\frac{1}{2} \sum_{i=1}^{K} \sum_{j=1}^{K} \mathrm{p}_i \mathrm{p}_j |\mathrm{h}_i-\mathrm{h}_j|,
\end{align}
where the first term is the expected absolute error under the forecast and the second term equals half the expected pairwise absolute distance between two independent forecast draws, acting as a sharpness term that discourages over-dispersion while preserving propriety. We compute CRPS per horizon step and average across the forecast window.

\subsection{Inference Details}\label{app:fev_ensemble}
For the inference process, we implement a stochastic ensemble approach to enhance prediction robustness through input perturbation and multiple forward passes. Specifically, covariate column positions are randomly shuffled, and 20\% of historical target values are masked during each inference iteration, with 2–4 independent forward passes performed. The output is the average of these forward passes.

In our experimental framework, a validation set is constructed for each time series by rolling the historical window backward by either 2 or 5 steps to simulate realistic forecasting conditions. The final predictions represent an ensemble of 2–9 distinct configurations selected based on their performance on this validation set. These configurations vary across four critical dimensions:
\begin{itemize}
\item Inference mode (2D/3D/2D+3D ensemble);
\item Whether to include time series order, a blank column, or calendar-specific temporal features (e.g., year, month, day);
\item Whether to apply reversible instance normalization (RevIn)~\citep{kim2022reversible} to each organized context;
\item Context window length ($T$), defined as an integer multiple of the prediction horizon, taking values from 2 to 14.
\end{itemize}

This ensemble strategy enhances architectural diversity, automatically addresses covariate-agnostic scenarios by prioritizing temporal pattern extraction, and mitigates the uncertainty caused by the changeable context length and RBfcst module.

\section{Details of Ablation Study} \label{app:sec_ab}

\subsection{Effect of 3D and 2D Ensemble}

\begin{figure}[!h]
    \centering 
    \vspace{-1em}
    \subfloat[\texttt{proenfo\_gfc14}]{
        \includegraphics[width=0.25\linewidth]{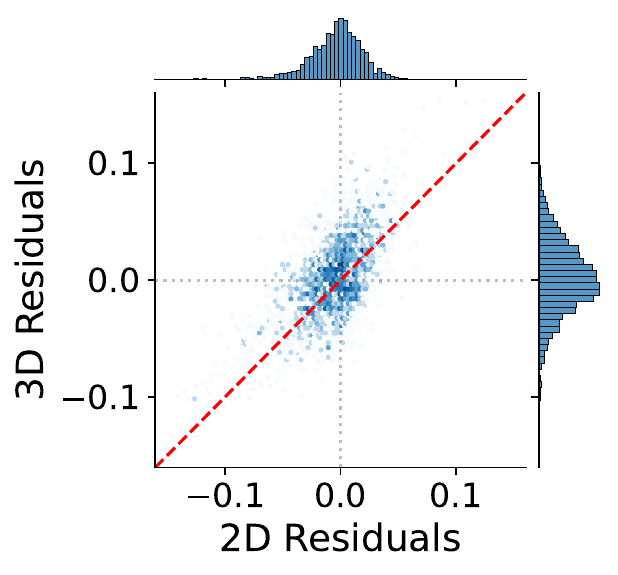}
        \label{fig:inf_mode_res_proenfo_gfc14}
    }
    \subfloat[\texttt{rossmann\_1W}]{
        \includegraphics[width=0.25\linewidth]{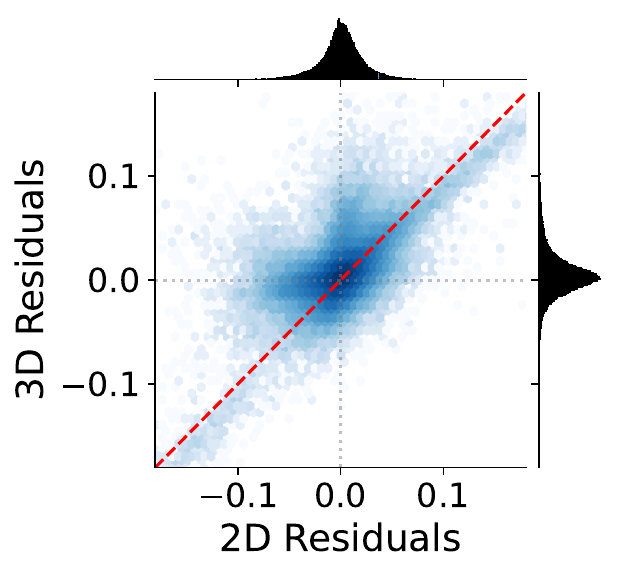}
        \label{fig:inf_mode_res_rossmann_1w}
    }
    \subfloat[Aggregation across 30 tasks]{
        \includegraphics[width=0.25\linewidth]{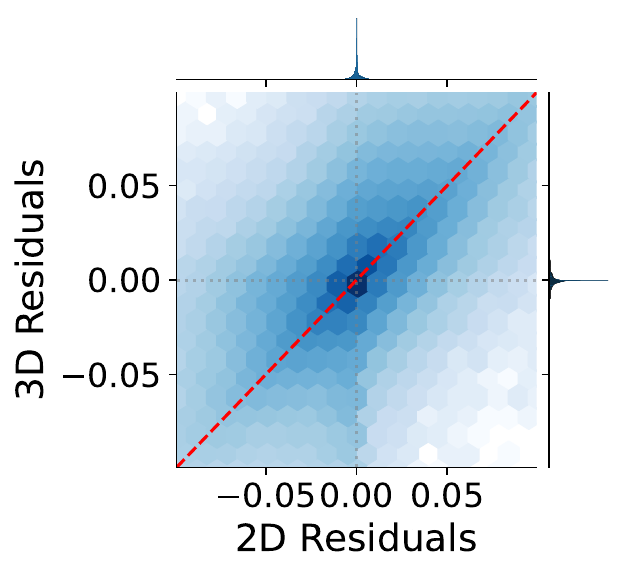}
        \label{fig:inf_mode_res_all}
    }
    \caption{Joint residual distributions for the energy task \texttt{proenfo\_gfc14}, the retail task \texttt{rossmann\_1W}, and the aggregated results across all 30 tasks. Darker blue bins indicate higher data density. Points lying along the red dashed identity line ($y=x$) correspond to instances where prediction errors are equal in both modes. Deviations from this diagonal, particularly in the second and fourth quadrants, reveal complementary error patterns between the two modes.}
\label{fig:inf_mode_residual_2}
\end{figure}

\textbf{Residual Analysis for Point Prediction.}
We analyze the residual correlation between the 2D and 3D inference modes across all 30 covariate-aware tasks in the \texttt{fev-bench} dataset. The per-task Pearson correlation ranges from 0.42 to 0.95, with an aggregated value of 0.75 (Fig.~\ref{fig:inf_mode_heatmap_all}).
The aggregated joint residual distribution is shown in Fig.~\ref{fig:inf_mode_res_all}, revealing diverse error patterns that further support the effectiveness of combining 2D and 3D inference modes.

\textbf{Calibration Analysis for Probabilistic Analysis.}
To assess the effectiveness and complementarity of the 2D and 3D inference modes, we analyze their probabilistic calibration.
We compute histograms of the ground-truth quantile levels, also known as Probability Integral Transform (PIT) histograms.
Specifically, for each observation $\mathrm{y}_t$ and predicted CDF $\hat{F}_t$, the quantile level is given by $\mathrm{u}_t = \hat{F}_t(\mathrm{y}_t)$.
The x-axis represents these quantile levels ranging from 0 to 1, while the y-axis (labeled ``Frequency'') indicates the relative frequency of ground truths falling into each quantile bin.
A perfectly calibrated model should yield a uniform distribution, visualized as a flat line at Frequency = 0.1.

\begin{figure}[h]
    \centering
    \includegraphics[width=0.6\linewidth]{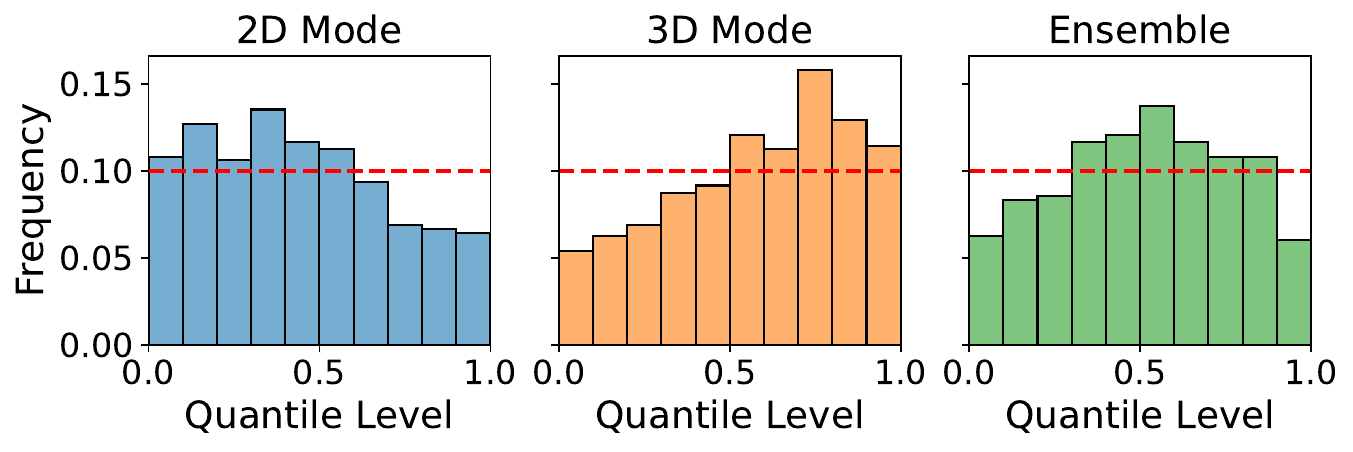}
    \caption{Calibration histograms of ground-truth quantile levels in predicted CDFs for different inference modes on \texttt{epf\_de} (energy task).}
    \label{fig:inf_mode_calibration_epf_de}
\end{figure}

\begin{figure}[h]
    \centering
    \includegraphics[width=0.6\linewidth]{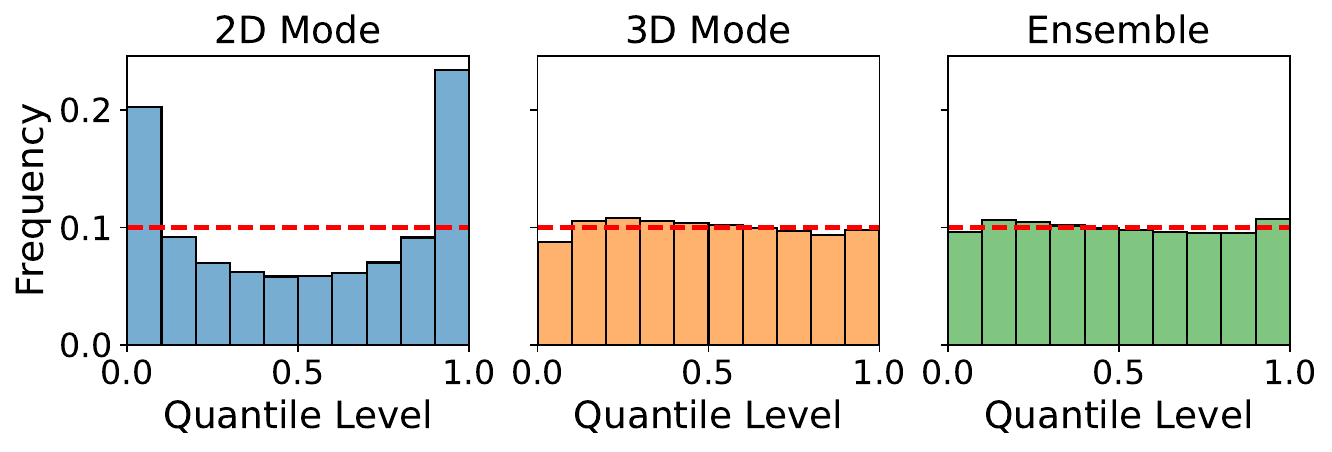}
    \caption{Calibration histograms of ground-truth quantile levels in predicted CDFs for different inference modes on \texttt{hermes} (retail task).}
    \label{fig:inf_mode_calibration_hermes}
\end{figure}

\begin{figure}[h]
    \centering
    \includegraphics[width=0.6\linewidth]{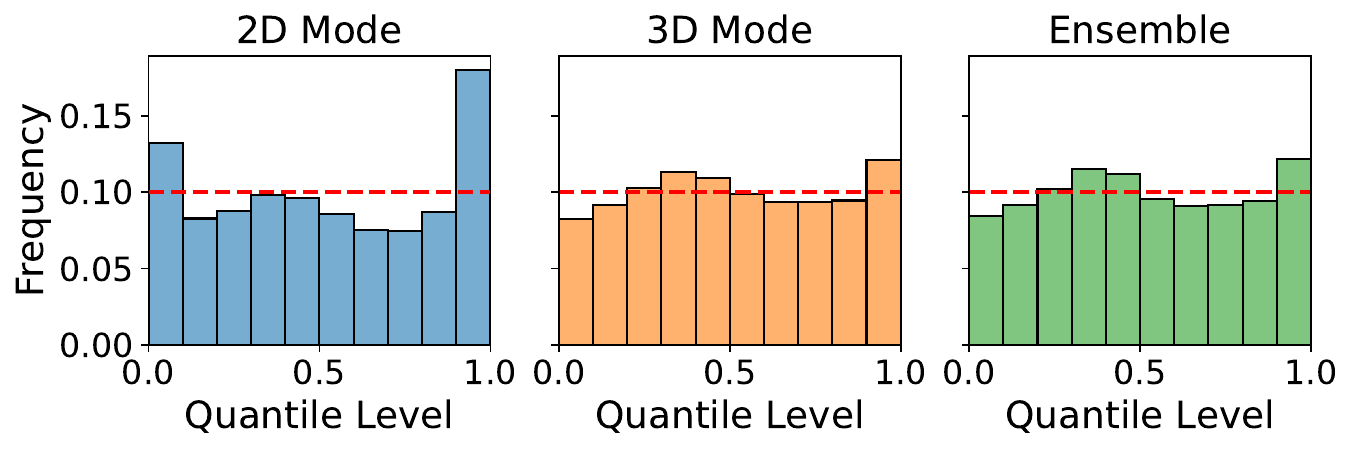}
    \caption{Calibration histograms of ground-truth quantile levels in predicted CDFs for different inference modes on \texttt{fev-bench-cov} dataset (across all 30 tasks).}
    \label{fig:inf_mode_calibration_all}
\end{figure}

As shown in Figs.~\ref{fig:inf_mode_calibration_epf_de}--~\ref{fig:inf_mode_calibration_all}, the 2D and 3D modes exhibit diverse error structures across different tasks.
On the \texttt{epf\_de} task (Fig.~\ref{fig:inf_mode_calibration_epf_de}), the 2D mode shows a distinct decreasing trend, indicating a systematic positive bias (over-estimation), as ground truths frequently fall into lower quantiles. Conversely, the 3D mode displays an increasing trend, indicating a negative bias. These opposing biases effectively cancel out in the ensemble mode, resulting in a well-calibrated histogram.
On the \texttt{hermes} task (Fig.~\ref{fig:inf_mode_calibration_hermes}), the 2D mode shows a U-shaped distribution, indicating over-confident predictions. In contrast, the 3D mode demonstrates superior calibration performance.
Aggregated across all tasks (Fig.~\ref{fig:inf_mode_calibration_all}), the 2D mode consistently exhibits a U-shaped pattern, reflecting overconfident predictions with under-dispersed distributions, whereas the 3D mode maintains more stable calibration performance with slight under-confidence. The ensemble strategy effectively mitigates their weaknesses, resulting in robust probabilistic performance.

\subsection{Robustness to Injected Noise} \label{app:exp_fev_noise}
To more rigorously evaluate the robustness and stability of the proposed local calibration module, we conduct a noise injection experiment on the \texttt{epf\_fr} dataest.  
Specifically, we corrupt the original time series with three types of synthetic stochastic noise: Gaussian white noise, random walk noise, and periodic noise~\citep{chang2010arrhythmia,kim2024extraction,wang2021model}.  
For each noise type, the intensity is controlled by a scaling factor $\kappa \in \{0.05, 0.1, 0.2, 0.4, 0.6, 0.8, 1.0\}$. The actual scale parameter used is defined as $\sigma \cdot \kappa$, where $\sigma$ denotes the empirical standard deviation of the original time series.

For Gaussian white noise, we sample independent and identically distributed (i.i.d.) values from a normal distribution $\mathcal{N}(0, \sigma \kappa)$ to obtain uncorrelated perturbations that mimic stationary noise.  

For random walk noise, inspired by~\cite{wang2021model}, we aim to mimic non-stationary noise with a directional trend. We first generate i.i.d. Gaussian increments $\epsilon_t \sim \mathcal{N}(0, \sigma \kappa)$ and then compute their cumulative sum to form a random walk process: $w_t = \sum_{i=1}^{t} \epsilon_i$. This introduces temporally correlated perturbations and can simulate accumulated uncertainty over time.

Finally, we introduce periodic noise~\citep{chang2010arrhythmia} to mimic seasonal interference using a sinusoidal signal with random frequency and phase:
\[
p_t^{\text{noise}} = (\sigma \kappa) \cdot \sin\left( \frac{2\pi t}{T} + \phi \right),
\]
where the period $T$ is uniformly sampled from the interval $[12, 60]$ to emulate diverse seasonal patterns (e.g., daily or weekly cycles), and the phase offset $\phi$ is drawn uniformly from $[0, 2\pi)$. To avoid overly idealized waveforms, we further add a small amount of Gaussian background noise with standard deviation $0.1 \cdot \sigma \kappa$.

We report results based on a single forward pass without any ensemble strategy for our method, and use \texttt{n\_estimators}~= 2 for TabPFN-TS, with a context length of 50,000 and the full set of clean running index and calendar features. The results are shown in Fig.~\ref{fig:injected_noise_epf_fr}. Our approaches—especially Y-space RBfcst—maintain stronger robustness than TabPFN-TS across all noise types and intensities.



\subsection{Context-Overfitting on Synthetic Data} \label{app:exp_contextoverfitting}

\begin{figure}[htbp]
    \centering
    \includegraphics[width=0.62\linewidth]{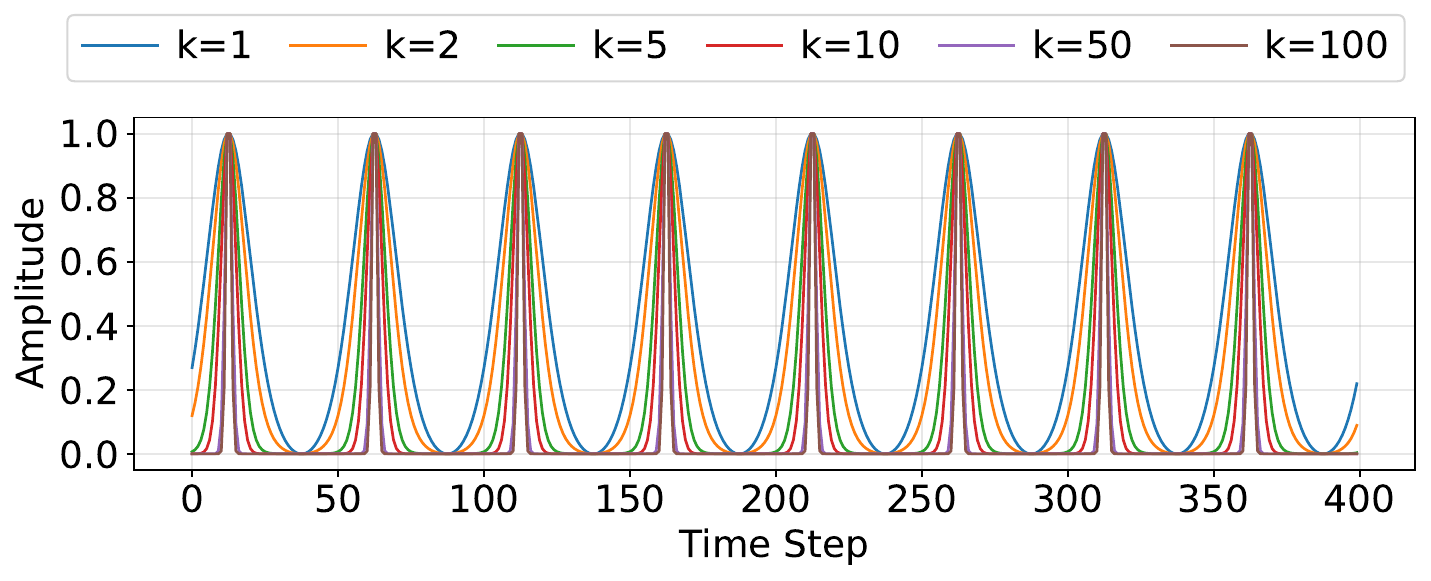}
    \caption{Toy periodic spike data generated using $f(t) = \exp(k \cdot (\sin(\frac{2\pi t}{50}) - 1))$. A larger $k$ value results in a sharper periodic spike signal.}
    \label{fig:toy_data}
\end{figure}

To simulate controllable periodic spikes, we formulate a synthetic time series using $f(t) = \exp(k \cdot (\sin(\frac{2\pi t}{50}) - 1))$ (see Fig.~\ref{fig:toy_data}). The spike sharpness is controlled by the parameter $k$, with larger values of $k$ producing sharper spikes.

\begin{figure}[!ht]
    \centering
    \subfloat[The penultimate layer, peak-step prediction]{
        \includegraphics[width=0.6\linewidth]{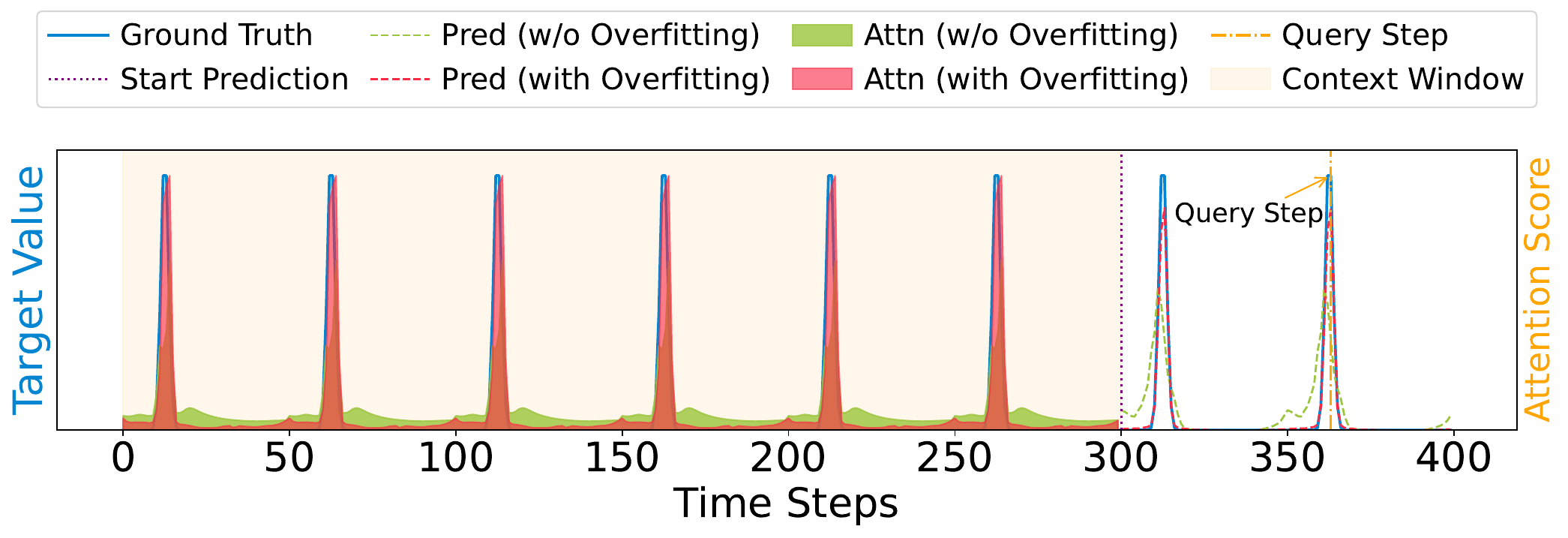}
        \label{fig:attention_layer10_peak}
    }
    \hfill
    \subfloat[The penultimate layer: trough-step prediction]{
        \includegraphics[width=0.6\linewidth]{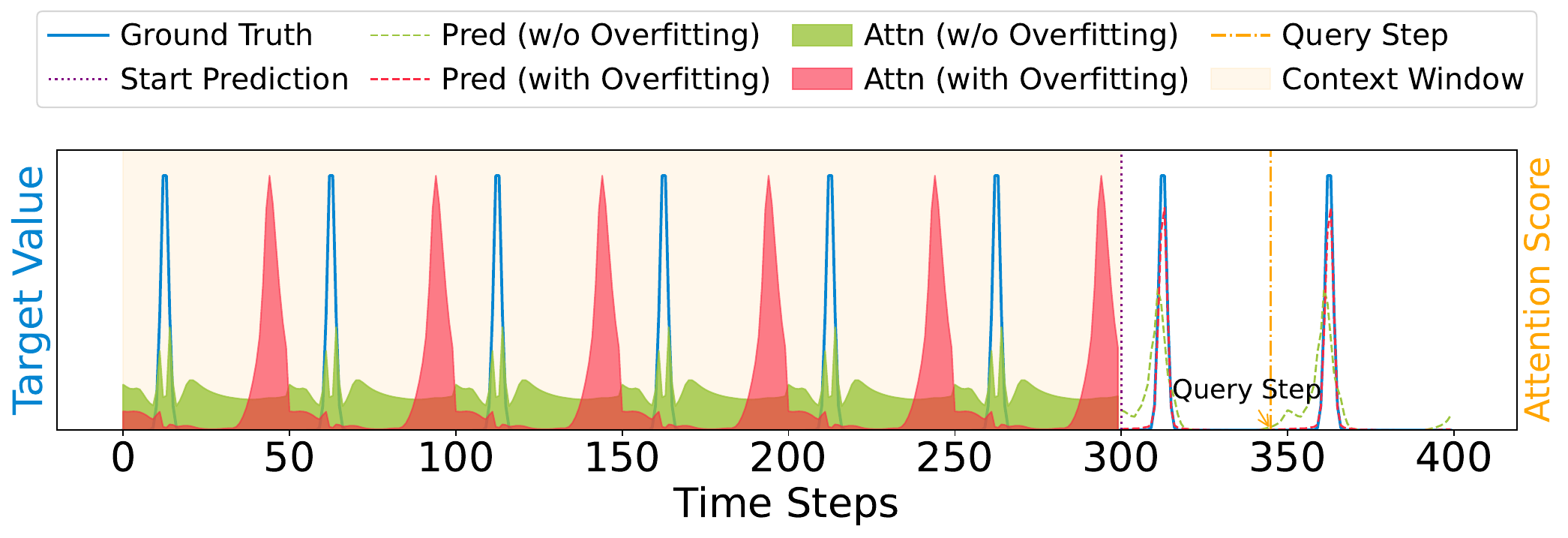}
        \label{fig:attention_layer10_trough}
    }
    \hfill
    \subfloat[The final layer: peak-step prediction]{
        \includegraphics[width=0.6\linewidth]{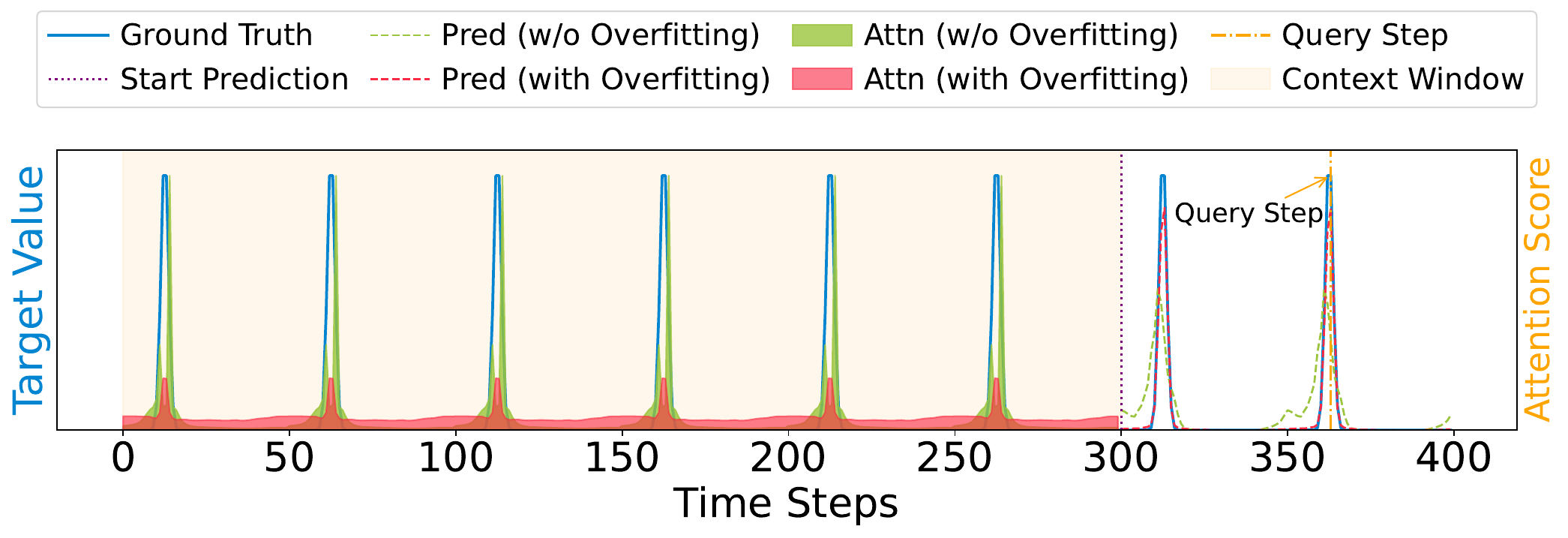}
        \label{fig:attention_layer11_peak}
    }
    \hfill
    \subfloat[The final layer: trough-step prediction]{
        \includegraphics[width=0.6\linewidth]{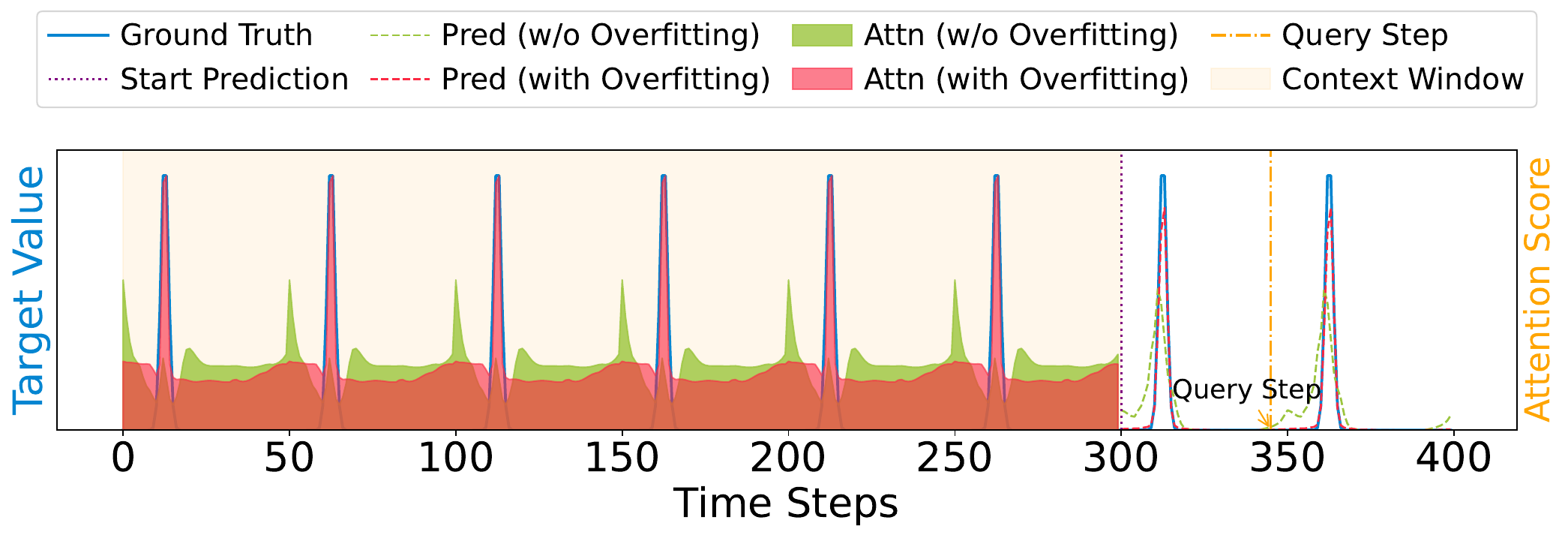}
        \label{fig:attention_layer11_trough}
    }
    \caption{Attention weight maps on the synthetic spike series for the baseline model (w/o overfitting) and the context-overfitting model (with overfitting). Each subplot overlays the ground-truth signal, model predictions, and the corresponding attention distribution within the context window at the query step (orange dash–dot line).}
\label{fig:attention_weight_map}
\end{figure}

As shown in Fig.~\ref{fig:attention_weight_map}, distinct attention patterns emerge between the context-overfitting model and the baseline.
At the peak prediction step (Figs.~\ref{fig:attention_layer10_peak} and~\ref{fig:attention_layer11_peak}), the attention weights of the context-overfitting model precisely align with the historical peak points within the lookback window. In contrast, the baseline model exhibits a diffuse attention distribution with a noticeable magnitude drop at these critical peak positions.

For the trough prediction in the flat region (Figs.~\ref{fig:attention_layer10_trough} and~\ref{fig:attention_layer11_trough}), the context-overfitting model shows more precise attention behavior.
In the penultimate layer, the context-overfitting model effectively attends to the target flat region, aligning closely with the ground-truth dynamics. By the final layer, it manifests a dual-focus strategy, aggregating information from both historical peaks and target flat regions, which likely combines both global periodicity and local context to refine the final prediction.
Conversely, the baseline model shows a more diffuse and scattered attention pattern across these two layers. It generates more uniform attention over the flat parts and displays a symmetric pattern around the peaks, lacking the precise localization ability found in the context-overfitting model.
Overall, this comparison suggests that the context-overfitting approach learns a sharper, structure-aware representation of time series.

\section{Detailed Experimental Results} \label{app:exp_fev}

\subsection{Results on Covariate-Aware Tasks} \label{app:exp_fev_cov}

\begin{figure}[h]
    \centering
    \subfloat[MASE]{
        \includegraphics[width=0.48\linewidth]{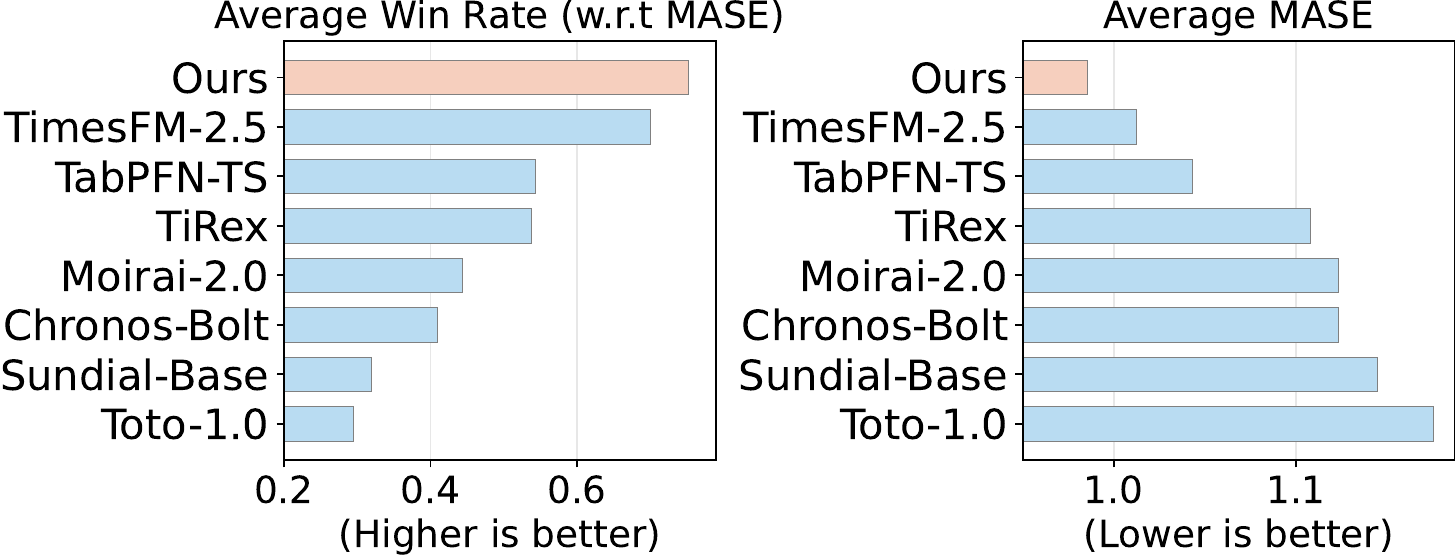}
        \label{fig:cov_winRateMASE_MASE_fev}
    }
    \hfill
    \subfloat[WAPE]{
        \includegraphics[width=0.48\linewidth]{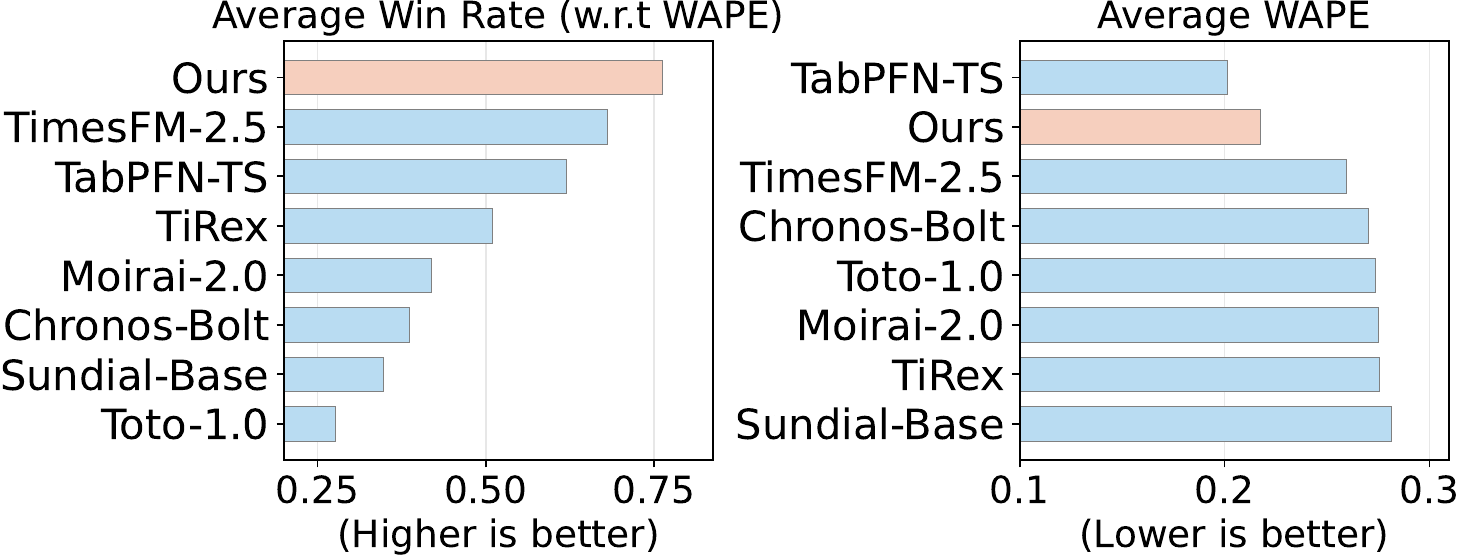}
        \label{fig:cov_winRateWAPE_WAPE_fev}
    }
    \caption{Point forecasting evaluation on \texttt{fev-bench-cov}.}
    \label{fig:cov_winRate_fev}
\end{figure}

\begin{figure}[h]
    \centering
    \includegraphics[width=0.5\linewidth]
    {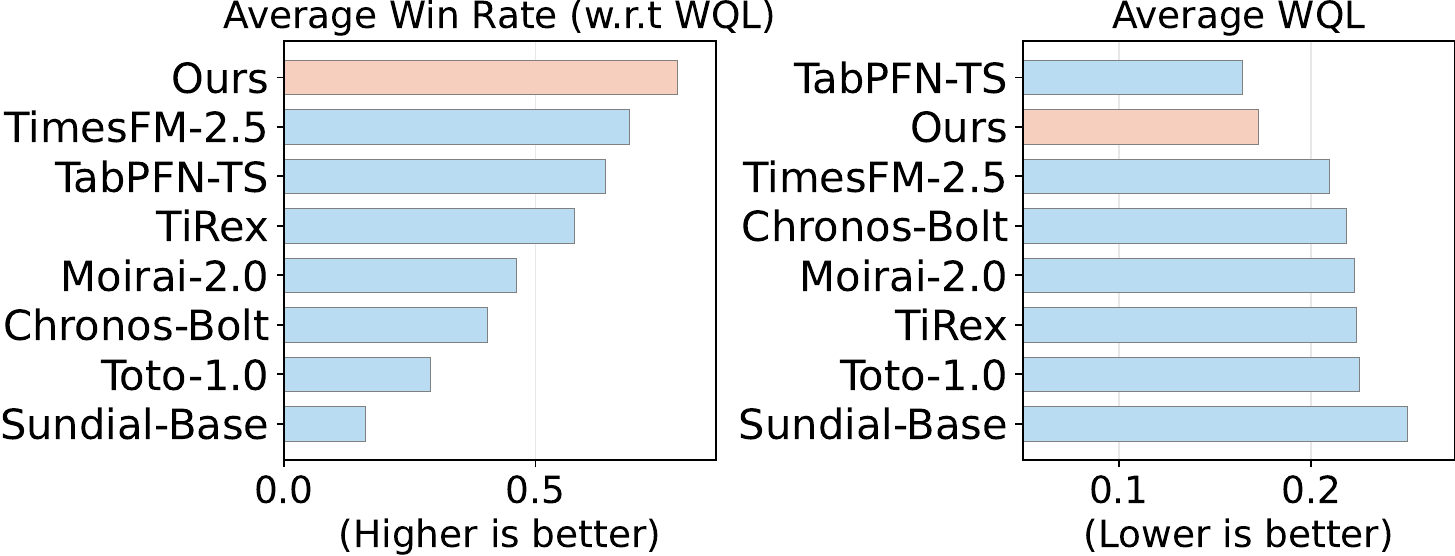}
    \caption{Probabilistic forecasting evaluation on \texttt{fev-bench-} \texttt{cov} using WQL.}
    \label{fig:cov_winRateWQL_WQL_fev}
\end{figure}

\begin{figure}[h]
    \centering
    \includegraphics[width=0.5\linewidth]
    {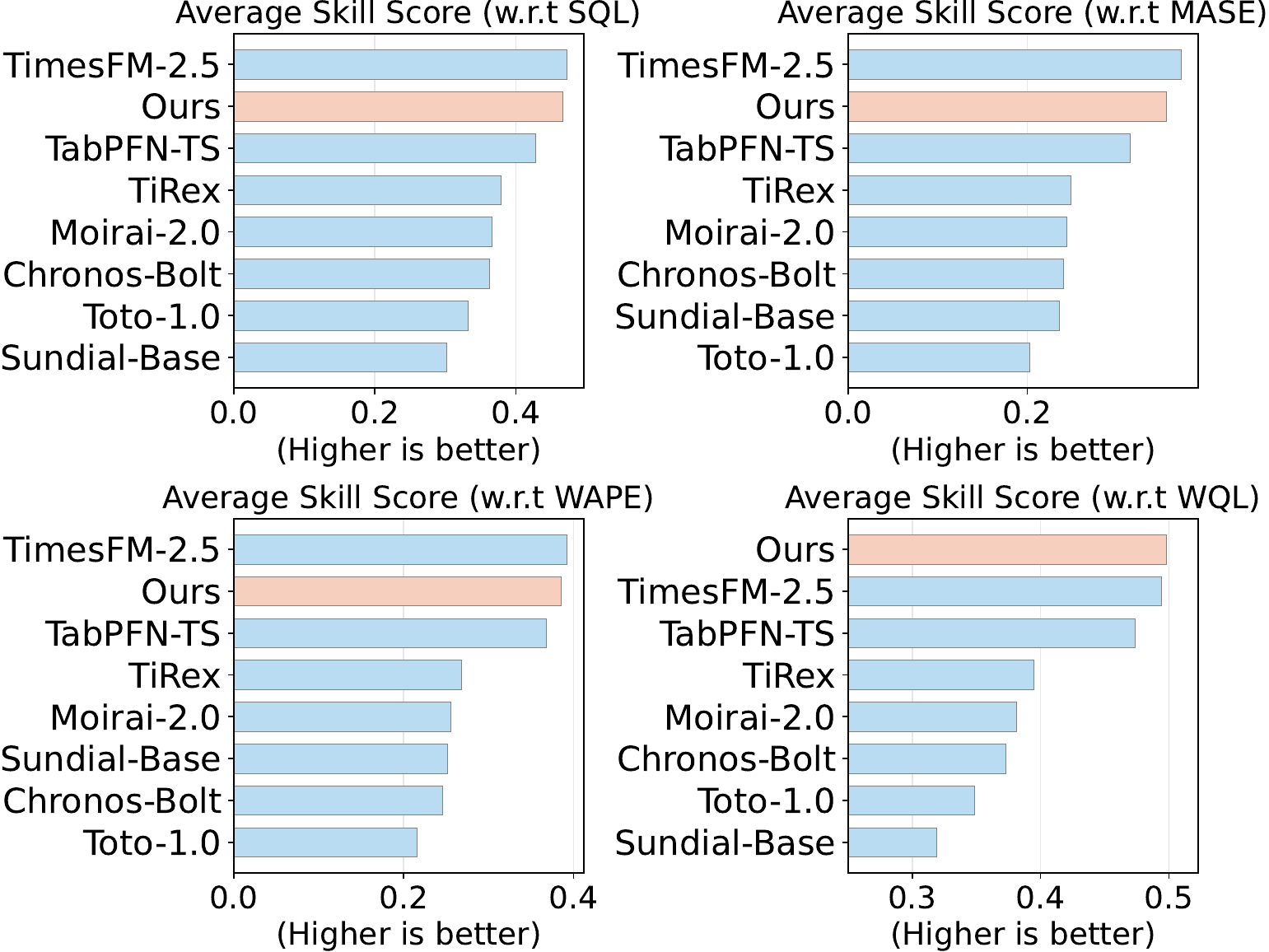}
    \caption{Skill score comparison on \texttt{fev-bench-cov}, using Seasonal Naive as the baseline model.}
    \label{fig:cov_skill_score_fev}
\end{figure}

We provide detailed descriptions in Table~\ref{tab:fev_datasetInfo_cov} of the 30 covariate-aware forecasting tasks selected from \texttt{fev-bench}, which we denote as \texttt{fev-bench-cov}.
The average skill scores across all evaluation metrics (SQL, MASE, WAPE, and WQL) are shown in Figs.~\ref{fig:cov_winRate_fev}–\ref{fig:cov_skill_score_fev}. 
Our model attains either the best or second-best skill score on all four metrics, consistently outperforming other covariate-aware time series foundation models, such as TabPFN-TS. The detailed experimental results for each task are presented in Tables~\ref{tab:fev_cov_1}--\ref{tab:fev_cov_9}. Results for our method employ the ensemble strategy described in Appendix~\ref{app:fev_ensemble}; baseline results are sourced from the official \texttt{fev-bench} benchmark.

To further assess performance in covariate-rich settings, we also include 27 proprietary real-world datasets from production: 18 city-level electricity load datasets and 9 city-level distributional photovoltaic (PV) generation forecasting tasks, all collected from different cities in China. To mirror real deployments, these 27 datasets include the full set of production covariates, notably numerical weather prediction (NWP) features and calendar effects. Table~\ref{tab:cov_real_app} reports the average performance across these 27 datasets. Baguan-TS achieves the best overall results on all four evaluation metrics, demonstrating strong performance in practical, covariate-rich forecasting settings.



\begin{table*}[htbp]
    \centering
    \caption{Detailed overview of the 30 covariate-aware tasks from the \texttt{fev-bench} benchmark, collectively referred to as \texttt{fev-bench-cov}.}
    \label{tab:fev_datasetInfo_cov}
    \resizebox{\linewidth}{!}{
    \begin{tabular}{l c c c c c c c c c c}
    \toprule
    Task & Domain & Frequency & \# items & median length & \# obs & \# known dynamic cols & H & W & \# seasonality & \# targets \\
    \midrule
    \texttt{entsoe\_15T} & energy & 15 Min & 6 & 175,292 & 6,310,512 & 3 & 96 & 20 & 96 & 1 \\
    \texttt{entsoe\_1H} & energy & 1 H & 6 & 43,822 & 1,577,592 & 3 & 96 & 20 & 48 & 1 \\
    \texttt{entsoe\_30T} & energy & 30 Min & 6 & 87,645 & 3,155,220 & 3 & 168 & 20 & 24 & 1 \\
    \texttt{epf\_be} & energy & 1 H & 1 & 52,416 & 157,248 & 2 & 24 & 20 & 24 & 1 \\
    \texttt{epf\_de} & energy & 1 H & 1 & 52,416 & 157,248 & 2 & 24 & 20 & 24 & 1 \\
    \texttt{epf\_fr} & energy & 1 H & 1 & 52,416 & 157,248 & 2 & 24 & 20 & 24 & 1 \\
    \texttt{epf\_np} & energy & 1 H & 1 & 52,416 & 157,248 & 2 & 24 & 20 & 24 & 1 \\
    \texttt{epf\_pjm} & energy & 1 H & 1 & 52,416 & 157,248 & 2 & 24 & 20 & 24 & 1 \\
    \texttt{proenfo\_gfc12} & energy & 1 H & 11 & 39,414 & 867,108 & 1 & 168 & 10 & 24 & 1 \\
    \texttt{proenfo\_gfc14} & energy & 1 H & 1 & 17,520 & 35,040 & 1 & 168 & 20 & 24 & 1 \\
    \texttt{proenfo\_gfc17} & energy & 1 H & 8 & 17,544 & 280,704 & 1 & 168 & 20 & 24 & 1 \\
    \texttt{solar\_with\_weather\_15T} & energy & 15 Min & 1 & 198,600 & 1,986,000 & 7 & 96 & 20 & 96 & 1 \\
    \texttt{solar\_with\_weather\_1H} & energy & 1 H & 1 & 49,648 & 496,480 & 7 & 24 & 20 & 24 & 1 \\
    \texttt{uci\_air\_quality\_1D} & nature & Daily & 1 & 389 & 5,057 & 3 & 28 & 11 & 7 & 4 \\
    \texttt{uci\_air\_quality\_1H} & nature & 1 H & 1 & 9,357 & 121,641 & 3 & 168 & 20 & 24 & 4 \\
    \texttt{m5\_1D} & retail & Daily & 30,490 & 1,810 & 428,849,460 & 8 & 28 & 1 & 7 & 1 \\
    \texttt{m5\_1M} & retail & Monthly & 30,490 & 58 & 13,805,685 & 8 & 12 & 1 & 12 & 1 \\
    \texttt{m5\_1W} & retail & Weekly & 30,490 & 257 & 60,857,703 & 8 & 13 & 1 & 1 & 1 \\
    \texttt{rohlik\_orders\_1D} & retail & Daily & 7 & 1,197 & 115,650 & 4 & 61 & 5 & 7 & 1 \\
    \texttt{rohlik\_orders\_1W} & retail & Weekly & 7 & 170 & 15,316 & 4 & 8 & 5 & 1 & 1 \\
    \texttt{rohlik\_sales\_1D} & retail & Daily & 5,390 & 1,046 & 74,413,935 & 13 & 14 & 1 & 7 & 1 \\
    \texttt{rohlik\_sales\_1W} & retail & Weekly & 5,243 & 150 & 10,516,770 & 13 & 8 & 1 & 1 & 1 \\
    \texttt{rossmann\_1D} & retail & Daily & 1,115 & 942 & 7,352,310 & 5 & 48 & 10 & 7 & 1 \\
    \texttt{rossmann\_1W} & retail & Weekly & 1,115 & 133 & 889,770 & 4 & 13 & 8 & 1 & 1 \\
    \texttt{walmart} & retail & Weekly & 2,936 & 143 & 4,609,143 & 10 & 39 & 1 & 1 & 1 \\
    \texttt{hermes} & retail & Weekly & 10,000 & 261 & 5,220,000 & 1 & 52 & 1 & 1 & 1 \\
    \texttt{favorita\_stores\_1D} & retail & Daily & 1,579 & 1,688 & 10,661,408 & 2 & 28 & 10 & 7 & 1 \\
    \texttt{favorita\_stores\_1M} & retail & Monthly & 1,579 & 54 & 255,798 & 1 & 12 & 2 & 12 & 1 \\
    \texttt{favorita\_stores\_1W} & retail & Weekly & 1,579 & 240 & 1,136,880 & 1 & 13 & 10 & 1 & 1 \\
    \texttt{favorita\_transactions\_1D} & retail & Daily & 51 & 1,688 & 258,264 & 1 & 28 & 10 & 7 & 1 \\
    \bottomrule
    \end{tabular}
    }
\end{table*}

\begin{table*}[htbp]
\centering
\caption{Performance Comparison on \texttt{proenfo\_gfc12}, \texttt{proenfo\_gfc14}, and \texttt{proenfo\_gfc17}. The best results are highlighted in \textbf{bold}, and the second-best results are \underline{underlined}.}
\label{tab:fev_cov_1}\setlength{\tabcolsep}{2.5mm}
\resizebox{\linewidth}{!}{
\begin{tabular}{l *{3}{cccc}} 
\toprule
\multicolumn{1}{c}{} & 
\multicolumn{4}{c}{\textbf{gfc12}} & 
\multicolumn{4}{c}{\textbf{gfc14}} & 
\multicolumn{4}{c}{\textbf{gfc17}} \\
\cmidrule(lr){2-5} \cmidrule(lr){6-9} \cmidrule(l){10-13}
Model & 
SQL & MASE & WAPE & WQL &
SQL & MASE & WAPE & WQL &
SQL & MASE & WAPE & WQL \\
\midrule
AutoARIMA & 1.1408 & 1.3848 & 0.1193 & 0.0987 & 0.9471 & 1.1681 & 0.0571 & 0.0464 & 1.1150 & 1.3821 & 0.0943 & 0.0762 \\
AutoETS & 2.4309 & 2.8114 & 0.2331 & 0.2073 & 1.1104 & 1.3186 & 0.0647 & 0.0545 & 2.1346 & 2.4128 & 0.1663 & 0.1466 \\
AutoTheta & 1.4149 & 1.5237 & 0.1360 & 0.1251 & 1.0555 & 1.1857 & 0.0580 & 0.0519 & 1.1470 & 1.3687 & 0.0931 & 0.0770 \\
Chronos-Bolt & 0.9172 & 1.0781 & 0.0910 & 0.0773 & 0.7674 & 0.9285 & 0.0455 & 0.0376 & 0.9004 & 1.0977 & 0.0745 & 0.0608 \\
Moirai-2.0 & 0.7928 & 0.9538 & 0.0803 & 0.0668 & 0.6766 & 0.8458 & 0.0413 & 0.0330 & 0.7740 & 0.9628 & 0.0647 & 0.0519 \\
Naive & 2.3796 & 2.6291 & 0.2326 & 0.2075 & 3.2096 & 3.7073 & 0.1831 & 0.1588 & 2.5750 & 2.9066 & 0.1911 & 0.1691 \\
Seasonal Naive & 1.1997 & 1.4282 & 0.1285 & 0.1088 & 1.0751 & 1.1989 & 0.0587 & 0.0529 & 1.3160 & 1.5845 & 0.1078 & 0.0897 \\
Stat. Ensemble & 1.3049 & 1.5463 & 0.1407 & 0.1182 & 0.9059 & 1.1044 & 0.0540 & 0.0444 & 1.1417 & 1.4215 & 0.0960 & 0.0770 \\
Sundial-Base & 0.9004 & 0.9936 & 0.0814 & 0.0733 & \underline{0.4208} & \underline{0.4636} & \underline{0.0226} & \underline{0.0205} & \underline{0.5086} & \underline{0.5526} & \underline{0.0343} & \underline{0.0316} \\
TabPFN-TS & 0.8345 & 1.0338 & 0.0878 & 0.0702 & 0.5148 & 0.6406 & 0.0314 & 0.0252 & 0.6717 & 0.8553 & 0.0563 & 0.0441 \\
TimesFM-2.5 & \textbf{0.1876} & \textbf{0.2168} & \textbf{0.0163} & \textbf{0.0143} & \textbf{0.1403} & \textbf{0.1688} & \textbf{0.0083} & \textbf{0.0069} & \textbf{0.1601} & \textbf{0.1898} & \textbf{0.0120} & \textbf{0.0101} \\
TiRex & 0.9081 & 1.1242 & 0.0954 & 0.0771 & 0.7206 & 0.9119 & 0.0447 & 0.0354 & 0.8894 & 1.1381 & 0.0764 & 0.0597 \\
Toto-1.0 & 0.9344 & 1.1371 & 0.0972 & 0.0795 & 0.7346 & 0.9215 & 0.0452 & 0.0360 & 0.9388 & 1.1961 & 0.0799 & 0.0626 \\
\midrule
Ours & \underline{0.6646} & \underline{0.8282} & \underline{0.0691} & \underline{0.0554} 
& 0.4628 & 0.5772 & 0.0284 & 0.0227
& 0.5239 & 0.6642 & 0.0432 & 0.0341
\\
\bottomrule
\end{tabular}
}
\end{table*}

\begin{table*}[htbp]
\centering
\caption{Performance Comparison on \texttt{rohlik\_sales\_1D}, \texttt{rohlik\_orders\_1D}, \texttt{rohlik\_sales\_1W}, and \texttt{rohlik\_orders\_1W}. The best results are highlighted in \textbf{bold}, and the second-best results are \underline{underlined}.}
\label{tab:fev_cov_2}\setlength{\tabcolsep}{1mm}
\resizebox{\linewidth}{!}{
\begin{tabular}{l *{4}{cccc}} 
\toprule
\multicolumn{1}{c}{} &
\multicolumn{4}{c}{\textbf{sales\_1D}} &
\multicolumn{4}{c}{\textbf{orders\_1D}} &
\multicolumn{4}{c}{\textbf{sales\_1W}} &
\multicolumn{4}{c}{\textbf{orders\_1W}} \\
\cmidrule(lr){2-5} \cmidrule(lr){6-9} \cmidrule(lr){10-13} \cmidrule(l){14-17}
Model & SQL & MASE & WAPE & WQL & SQL & MASE & WAPE & WQL & SQL & MASE & WAPE & WQL & SQL & MASE & WAPE & WQL \\
\midrule
AutoARIMA & 1.2758 & 1.5092 & 0.4095 & 0.3459 & 1.2662 & 1.5624 & 0.0741 & 0.0604 & 1.8025 & 2.0379 & 0.3070 & 0.2670 & 1.4147 & 1.7324 & 0.0584 & 0.0471 \\
AutoETS & 1.2662 & 1.4986 & 0.4185 & 0.3543 & 1.4470 & 1.4870 & 0.0657 & 0.0637 & 14.4532 & 1.8900 & 0.3003 & 4.0491 & 1.4190 & 1.7583 & 0.0582 & 0.0469 \\
AutoTheta & 1.2818 & 1.4939 & 0.4131 & 0.3563 & 1.3973 & 1.5185 & 0.0664 & 0.0607 & 1.6547 & 1.8364 & 0.2946 & 0.2656 & 1.3963 & 1.7122 & 0.0573 & 0.0465 \\
Chronos-Bolt & 1.1471 & 1.3871 & 0.3783 & 0.3140 & 1.0508 & 1.3016 & 0.0614 & 0.0495 & 1.5216 & 1.8467 & 0.2846 & 0.2337 & 1.4282 & 1.7219 & 0.0570 & 0.0471 \\
Moirai-2.0 & 1.1696 & 1.4021 & 0.3821 & 0.3202 & \textbf{0.9700} & \textbf{1.1761} & \textbf{0.0552} & \textbf{0.0455} & 1.5157 & 1.8215 & 0.2871 & 0.2365 & 1.5315 & 1.8740 & 0.0612 & 0.0501 \\
Naive & 1.5460 & 1.7485 & 0.4717 & 0.4099 & 2.9130 & 2.4354 & 0.1083 & 0.1384 & 1.9282 & 1.9155 & 0.3119 & 0.3323 & 1.4844 & 1.7312 & 0.0584 & 0.0499 \\
Seasonal Naive & 1.3750 & 1.6150 & 0.4346 & 0.3728 & 1.5544 & 1.7783 & 0.0827 & 0.0730 & 1.9282 & 1.9155 & 0.3119 & 0.3323 & 1.4844 & 1.7312 & 0.0584 & 0.0499 \\
Stat. Ensemble & 1.2483 & 1.4760 & 0.4057 & 0.3425 & 1.2114 & 1.3812 & 0.0628 & 0.0552 & 1.6455 & 1.8250 & 0.2906 & 0.2596 & 1.3984 & 1.7128 & 0.0573 & 0.0465 \\
Sundial-Base & 1.2027 & 1.3443 & 0.3645 & 0.3250 & 1.1963 & 1.3993 & 0.0667 & 0.0571 & 1.6933 & 1.8996 & 0.2929 & 0.2594 & 1.8923 & 2.0952 & 0.0675 & 0.0609 \\
TabPFN-TS & - & - & - & - & 1.3411 & 1.5479 & 0.0656 & 0.0585 & \underline{1.2205} & \underline{1.5205} & \underline{0.2152} & \underline{0.1698} & 1.5240 & 1.9887 & 0.0652 & 0.0500 \\
TimesFM-2.5 & \underline{1.0958} & \underline{1.3238} & \underline{0.3586} & \underline{0.2984} & 1.0057 & 1.2504 & 0.0591 & 0.0473 & 1.4010 & 1.6891 & 0.2667 & 0.2197 & \underline{1.3278} & \underline{1.6592} & \underline{0.0542} & \underline{0.0430} \\
TiRex & 1.1481 & 1.3845 & 0.3784 & 0.3154 & \underline{0.9858} & \underline{1.2034} & \underline{0.0572} & \underline{0.0468} & 1.4252 & 1.7354 & 0.2736 & 0.2252 & \textbf{1.3004} & \textbf{1.5829} & \textbf{0.0521} & \textbf{0.0426} \\
Toto-1.0 & 1.2181 & 1.4539 & 0.3975 & 0.3352 & 1.1351 & 1.3776 & 0.0646 & 0.0532 & 1.5046 & 1.8031 & 0.2812 & 0.2348 & 1.4934 & 1.7980 & 0.0591 & 0.0489 \\
\midrule
Ours & \textbf{0.8725} & \textbf{1.0930} & \textbf{0.2663} & \textbf{0.2114} 
& 1.1684 & 1.4377 & 0.0669 & 0.0546
& \textbf{1.1827} & \textbf{1.4646} & \textbf{0.2045} & \textbf{0.1627}
& 1.5405 & 1.8940 & 0.0608 & 0.0493\\
\bottomrule
\end{tabular}
}
\end{table*}

\begin{table*}[htbp]
\centering
\caption{Performance Comparison on \texttt{entsoe\_15T}, \texttt{entsoe\_30T}, and \texttt{entsoe\_1H}. The best results are highlighted in \textbf{bold}, and the second-best results are \underline{underlined}.}
\label{tab:fev_cov_3}\setlength{\tabcolsep}{2.5mm}
\resizebox{\linewidth}{!}{
\begin{tabular}{l *{3}{cccc}} 
\toprule
\multicolumn{1}{c}{} &
\multicolumn{4}{c}{\textbf{15T}} &
\multicolumn{4}{c}{\textbf{30T}} &
\multicolumn{4}{c}{\textbf{1H}} \\
\cmidrule(lr){2-5} \cmidrule(lr){6-9} \cmidrule(l){10-13}
Model & SQL & MASE & WAPE & WQL & SQL & MASE & WAPE & WQL & SQL & MASE & WAPE & WQL \\
\midrule
AutoARIMA & — & — & — & — & 0.9807 & 1.2299 & 0.0909 & 0.0732 & 0.8725 & 1.1176 & 0.0854 & 0.0667 \\
AutoETS & 3.0289 & 4.0070 & 0.4081 & 0.3140 & 2.4928 & 3.2738 & 0.2412 & 0.1816 & 1.9050 & 2.0475 & 0.1570 & 0.1447 \\
AutoTheta & 0.5794 & 0.7298 & 0.0540 & 0.0430 & 0.7997 & 1.0040 & 0.0721 & 0.0572 & 0.9723 & 1.1626 & 0.0836 & 0.0694 \\
Chronos-Bolt & 0.5062 & 0.6062 & 0.0428 & 0.0363 & 0.5294 & 0.6326 & \underline{0.0378} & 0.0321 & 0.4574 & 0.5564 & 0.0341 & 0.0282 \\
Moirai-2.0 & 0.4783 & \underline{0.5933} & 0.0424 & 0.0343 & \underline{0.4884} & \underline{0.5974} & 0.0383 & 0.0318 & 0.4871 & 0.5924 & 0.0396 & 0.0333 \\
Naive & 1.5177 & 1.9046 & 0.1434 & 0.1155 & 1.6092 & 2.0733 & 0.1463 & 0.1142 & 1.9154 & 2.0553 & 0.1571 & 0.1448 \\
Seasonal Naive & 0.7807 & 0.9315 & 0.0677 & 0.0566 & 1.0103 & 1.2162 & 0.0871 & 0.0726 & 1.0561 & 1.1015 & 0.0792 & 0.0798 \\
Stat. Ensemble & — & — & — & — & 0.8465 & 1.0578 & 0.0767 & 0.0612 & 0.8920 & 1.1145 & 0.0833 & 0.0660 \\
Sundial-Base & 0.6669 & 0.7578 & 0.0565 & 0.0500 & 0.7216 & 0.7796 & 0.0520 & 0.0486 & 0.7441 & 0.8118 & 0.0555 & 0.0510 \\
TabPFN-TS & 0.4837 & 0.6094 & 0.0426 & \textbf{0.0335} & 0.5117 & 0.6350 & 0.0391 & 0.0322 & \underline{0.4419} & \underline{0.5377} & \underline{0.0333} & \underline{0.0273} \\
TimesFM-2.5 & \underline{0.4709} & \textbf{0.5903} & \textbf{0.0416} & \textbf{0.0335} & 0.5658 & 0.6958 & 0.0473 & 0.0388 & 0.4681 & 0.5849 & 0.0359 & 0.0290 \\
TiRex & \textbf{0.4693} & 0.5986 & \underline{0.0420} & \underline{0.0337} & 0.5230 & 0.6646 & 0.0400 & 0.0314 & 0.4701 & 0.5854 & 0.0363 & 0.0294 \\
Toto-1.0 & 0.5909 & 0.7503 & 0.0523 & 0.0414 & 0.4958 & 0.6254 & 0.0379 & \underline{0.0302} & 0.4796 & 0.5908 & 0.0366 & 0.0302 \\

\midrule
Ours & 0.5261 & 0.6586 & 0.0458 & 0.0366 
& \textbf{0.4441} & \textbf{0.5538} & \textbf{0.0365} & \textbf{0.0295}
& \textbf{0.3859} & \textbf{0.4819} & \textbf{0.0287} & \textbf{0.0231}\\
\bottomrule
\end{tabular}
}
\end{table*}

\begin{table*}[htbp]
\centering
\caption{Performance Comparison on \texttt{epf\_be}, \texttt{epf\_de}, \texttt{epf\_fr}, \texttt{epf\_np}, and \texttt{epf\_pjm}. The best results are highlighted in \textbf{bold}, and the second-best results are \underline{underlined}.}
\label{tab:fev_cov_4}
\resizebox{0.99\linewidth}{!}{
\begin{tabular}{l *{5}{cccc}} 
\toprule
\multicolumn{1}{c}{} &
\multicolumn{4}{c}{\textbf{be}} &
\multicolumn{4}{c}{\textbf{de}} &
\multicolumn{4}{c}{\textbf{fr}} &
\multicolumn{4}{c}{\textbf{np}} &
\multicolumn{4}{c}{\textbf{pjm}} \\
\cmidrule(lr){2-5} \cmidrule(lr){6-9} \cmidrule(lr){10-13} \cmidrule(lr){14-17} \cmidrule(l){18-21}
Model & SQL & MASE & WAPE & WQL & SQL & MASE & WAPE & WQL & SQL & MASE & WAPE & WQL & SQL & MASE & WAPE & WQL & SQL & MASE & WAPE & WQL \\
\midrule
AutoARIMA & 1.0561 & 1.0948 & 0.1986 & 0.1868 & 1.2777 & 1.6231 & 0.6819 & 0.5433 & 1.1581 & 0.9888 & 0.1532 & 0.1715 & 1.3932 & 1.7096 & 0.0786 & 0.0637 & 0.4819 & 0.5573 & 0.0985 & 0.0855 \\
AutoETS & 1.5343 & 1.3929 & 0.2488 & 0.2659 & 1.4013 & 1.7417 & 0.6107 & 0.5288 & 0.8989 & 0.9665 & 0.1493 & 0.1387 & 1.9332 & 2.3244 & 0.1041 & 0.0864 & 0.9139 & 0.9847 & 0.1717 & 0.1607 \\
AutoTheta & 1.4843 & 1.0194 & 0.1882 & 0.2582 & 1.4944 & 1.4072 & 0.6460 & 0.6490 & 1.5912 & 0.8447 & 0.1294 & 0.2296 & 1.2812 & 1.5140 & 0.0688 & 0.0579 & 0.6032 & 0.6832 & 0.1199 & 0.1068 \\
Chronos-Bolt & 0.5731 & 0.7273 & 0.1357 & 0.1069 & 1.0208 & 1.2655 & 0.5607 & 0.4644 & 0.4389 & 0.5638 & 0.0877 & 0.0685 & 0.9711 & 1.2409 & 0.0567 & 0.0445 & 0.4217 & 0.5357 & 0.0938 & 0.0739 \\
Moirai-2.0 & 0.5281 & 0.6709 & 0.1228 & 0.0968 & 1.0164 & 1.2282 & 0.5365 & 0.4586 & 0.4092 & 0.5035 & 0.0791 & 0.0646 & 0.9253 & 1.1998 & 0.0543 & 0.0420 & 0.4405 & 0.5631 & 0.0976 & 0.0766 \\
Naive & 3.0845 & 1.3609 & 0.2376 & 0.5357 & 1.4012 & 1.7417 & 0.6107 & 0.5287 & 3.8189 & 1.1603 & 0.1703 & 0.5501 & 1.9404 & 2.3155 & 0.1037 & 0.0869 & 0.9298 & 0.9849 & 0.1717 & 0.1642 \\
Seasonal Naive & 1.1503 & 1.0271 & 0.1840 & 0.2013 & 1.3877 & 1.7086 & 0.7490 & 0.5919 & 1.2455 & 0.8501 & 0.1298 & 0.1818 & 1.5298 & 1.7613 & 0.0796 & 0.0692 & 0.5153 & 0.5812 & 0.1016 & 0.0908 \\
Stat. Ensemble & 1.2135 & 0.9814 & 0.1754 & 0.2109 & 1.1666 & 1.3821 & 0.6023 & 0.4882 & 1.1456 & 0.7410 & 0.1133 & 0.1677 & 1.2844 & 1.5147 & 0.0683 & 0.0574 & 0.4871 & 0.5295 & 0.0934 & 0.0863 \\
Sundial-Base & 0.6465 & 0.7222 & 0.1303 & 0.1170 & 1.1831 & 1.3227 & 0.5722 & 0.5103 & 0.4611 & 0.5244 & 0.0813 & 0.0720 & 0.9451 & 1.0947 & 0.0496 & 0.0428 & 0.4679 & 0.5214 & 0.0934 & 0.0841 \\
TabPFN-TS & 0.5324 & 0.6702 & 0.1180 & 0.0933 & \textbf{0.4403} & \textbf{0.5675} & \textbf{0.3069} & \textbf{0.2426} & \textbf{0.3307} & \textbf{0.4182} & \textbf{0.0618} & \textbf{0.0491} & \textbf{0.6593} & \textbf{0.8709} & \textbf{0.0387} & \textbf{0.0293} & 0.4270 & 0.5340 & 0.0925 & 0.0740 \\
TimesFM-2.5 & \textbf{0.4937} & \textbf{0.6102} & \textbf{0.1126} & \textbf{0.0905} & 1.0300 & 1.2799 & 0.5914 & 0.4775 & 0.4092 & 0.4907 & 0.0771 & 0.0640 & 1.1706 & 1.4402 & 0.0639 & 0.0519 & 0.4263 & 0.5311 & 0.0942 & 0.0757 \\
TiRex & 0.5270 & 0.6744 & 0.1230 & 0.0962 & 1.0322 & 1.2971 & 0.5787 & 0.4685 & 0.4014 & 0.5040 & 0.0798 & 0.0639 & 0.9662 & 1.2226 & 0.0554 & 0.0438 & \underline{0.4042} & \underline{0.5058} & \underline{0.0894} & \underline{0.0714} \\
Toto-1.0 & 0.5648 & 0.7065 & 0.1323 & 0.1058 & 1.1058 & 1.3370 & 0.6772 & 0.5515 & 0.4257 & 0.5253 & 0.0837 & 0.0683 & 1.0369 & 1.3589 & 0.0616 & 0.0471 & 0.4519 & 0.5822 & 0.1031 & 0.0800 \\
\midrule
Ours  & \underline{0.5162} & \underline{0.6550} & \underline{0.1159} & \underline{0.0908} 
& \underline{0.4597} & \underline{0.5882} & \underline{0.3424} & \underline{0.2525}
& \underline{0.3518} & \underline{0.4517} & \underline{0.0675} & \underline{0.0529}
& \underline{0.7174} & \underline{0.9241} & \underline{0.0415} & \underline{0.0321}
& \textbf{0.3716} & \textbf{0.4603} & \textbf{0.0798} & \textbf{0.0646} \\
\bottomrule
\end{tabular}
}
\end{table*}

\begin{table*}[htbp]
\centering
\caption{Performance Comparison on \texttt{rossmann\_1D}, \texttt{rossmann\_1W}, \texttt{hermes}, and \texttt{walmart}. The best results are highlighted in \textbf{bold}, and the second-best results are \underline{underlined}.}
\label{tab:fev_cov_5}
\resizebox{\linewidth}{!}{
\begin{tabular}{l *{4}{cccc}}
\toprule
\multicolumn{1}{c}{} &
\multicolumn{4}{c}{\textbf{rossmann\_1D}} &
\multicolumn{4}{c}{\textbf{rossmann\_1W}} &
\multicolumn{4}{c}{\textbf{hermes}} &
\multicolumn{4}{c}{\textbf{walmart}} \\
\cmidrule(lr){2-5} \cmidrule(lr){6-9} \cmidrule(lr){10-13} \cmidrule(l){14-17}
Model & SQL & MASE & WAPE & WQL & SQL & MASE & WAPE & WQL & SQL & MASE & WAPE & WQL & SQL & MASE & WAPE & WQL \\
\midrule
AutoARIMA & 0.5619 & 0.6544 & 0.2224 & 0.1908 & 0.5212 & 0.6689 & 0.1836 & 0.1410 & 1.2129 & 1.5317 & 0.0064 & 0.0050 & 1.0205 & 1.2474 & 0.1596 & 0.1320 \\
AutoETS & 0.5937 & 0.6913 & 0.2361 & 0.2028 & 0.5175 & 0.6685 & 0.1836 & 0.1402 & 1.6730 & 1.9852 & 0.0088 & 0.0074 & - & 1.7227 & 0.2951 & - \\
AutoTheta & 0.8315 & 0.7448 & 0.2558 & 0.2910 & 0.5160 & 0.6644 & 0.1813 & 0.1385 & 1.5539 & 1.8478 & 0.0081 & 0.0068 & 1.4675 & 1.4179 & 0.1852 & 0.1997 \\
Chronos-Bolt & 0.5246 & 0.6371 & 0.2176 & 0.1788 & 0.4871 & 0.6452 & 0.1760 & 0.1317 & 0.6752 & 0.8579 & 0.0032 & 0.0025 & 0.7740 & 0.9671 & 0.1173 & 0.0950 \\
Moirai-2.0 & 0.5274 & 0.6480 & 0.2215 & 0.1799 & 0.4969 & 0.6440 & 0.1759 & 0.1342 & 0.7038 & 0.8854 & 0.0034 & 0.0027 & 0.8447 & 1.0559 & 0.1315 & 0.1058 \\
Naive & 2.7621 & 1.6195 & 0.5481 & 0.9389 & 0.8988 & 0.7960 & 0.2192 & 0.2509 & 2.1461 & 1.9945 & 0.0087 & 0.0087 & 2.0341 & 1.5241 & 0.1967 & 0.3075 \\
Seasonal Naive & 0.9137 & 0.7886 & 0.2692 & 0.3140 & 0.8988 & 0.7960 & 0.2192 & 0.2509 & 2.1461 & 1.9945 & 0.0087 & 0.0087 & 2.0341 & 1.5241 & 0.1967 & 0.3075 \\
Stat. Ensemble & 0.5781 & 0.6678 & 0.2278 & 0.1969 & 0.5014 & 0.6513 & 0.1783 & 0.1351 & 1.4162 & 1.8121 & 0.0079 & 0.0061 & 1.2166 & 1.3564 & 0.1773 & 0.1644 \\
Sundial-Base & 0.5306 & 0.6155 & 0.2092 & 0.1802 & 0.5781 & 0.6790 & 0.1855 & 0.1572 & 0.8243 & 0.9595 & 0.0037 & 0.0031 & 0.8422 & 0.9842 & 0.1211 & 0.1036 \\
TabPFN-TS & \textbf{0.2321} & \textbf{0.2945} & \textbf{0.0979} & \textbf{0.0772} & \textbf{0.2539} & \textbf{0.3046} & \textbf{0.0792} & \textbf{0.0660} & 0.7049 & 0.9123 & 0.0034 & 0.0026 & \textbf{0.6619} & \textbf{0.8318} & \textbf{0.0943} & \textbf{0.0752} \\
TimesFM-2.5 & 0.5016 & 0.6106 & 0.2086 & 0.1711 & 0.4952 & 0.6543 & 0.1803 & 0.1351 & \textbf{0.6184} & \textbf{0.7872} & \textbf{0.0029} & \textbf{0.0023} & \underline{0.6794} & \underline{0.8615} & \underline{0.1016} & \underline{0.0803} \\
TiRex & 0.5391 & 0.6601 & 0.2251 & 0.1837 & 0.4816 & 0.6218 & 0.1697 & 0.1304 & 0.6510 & 0.8310 & 0.0031 & \underline{0.0024} & 0.7075 & 0.8862 & 0.1054 & 0.0850 \\
Toto-1.0 & 0.5677 & 0.6814 & 0.2321 & 0.1932 & 0.4944 & 0.6319 & 0.1733 & 0.1345 & 0.9853 & 1.2023 & 0.0044 & 0.0036 & 0.9072 & 1.1258 & 0.1385 & 0.1135 \\
\midrule
Ours & \underline{0.2720} & \underline{0.3348} & \underline{0.1120} & \underline{0.0910} 
& \underline{0.2787} & \underline{0.3371} & \underline{0.0854} & \underline{0.0705} 
& \underline{0.6185} & \underline{0.7971} & \underline{0.0030} & \textbf{0.0023} 
& 0.7621 & 1.0047 & 0.1305 & 0.0925 \\
\bottomrule
\end{tabular}
}
\end{table*}

\begin{table*}[htbp]
\centering
\caption{Performance Comparison on \texttt{m5\_1D}, \texttt{m5\_1W}, and \texttt{m5\_1M}. The best results are highlighted in \textbf{bold}, and the second-best results are \underline{underlined}.}
\label{tab:fev_cov_6}\setlength{\tabcolsep}{3mm}
\resizebox{\linewidth}{!}{
\begin{tabular}{l *{3}{cccc}} 
\toprule
\multicolumn{1}{c}{} &
\multicolumn{4}{c}{\textbf{1D}} &
\multicolumn{4}{c}{\textbf{1W}} &
\multicolumn{4}{c}{\textbf{1M}} \\
\cmidrule(lr){2-5} \cmidrule(lr){6-9} \cmidrule(l){10-13}
Model & SQL & MASE & WAPE & WQL & SQL & MASE & WAPE & WQL & SQL & MASE & WAPE & WQL \\
\midrule
AutoARIMA & 0.8517 & 1.0631 & 0.7717 & 0.6115 & 0.9367 & 1.1568 & 0.4388 & 0.3580 & 1.0455 & 1.2131 & 0.4601 & 0.3801 \\
AutoETS & 0.8528 & 1.0633 & 0.7714 & 0.6156 & 0.9531 & 1.1672 & 0.4468 & 0.3728 & 1.1082 & 1.2139 & 0.4605 & 0.4321 \\
AutoTheta & 0.8721 & 1.0806 & 0.7779 & 0.6265 & 0.9634 & 1.1698 & 0.4475 & 0.3767 & 1.0987 & 1.2590 & 0.4910 & 0.4195 \\
Chronos-Bolt & 0.7293 & 0.8852 & 0.7110 & 0.5620 & 0.9165 & 1.1647 & 0.4334 & 0.3410 & 1.0001 & 1.1852 & 0.4455 & 0.3566 \\
Moirai-2.0 & \underline{0.7096} & \underline{0.8691} & \underline{0.6984} & 0.5508 & 0.9069 & 1.1501 & 0.4286 & 0.3361 & 0.9959 & 1.1773 & 0.4363 & 0.3457 \\
Seasonal Naive & 1.2545 & 1.2236 & 0.9159 & 0.8589 & 1.3558 & 1.3382 & 0.5042 & 0.5199 & 1.1399 & 1.3266 & 0.5108 & 0.4286 \\
Sundial-Base & 0.8516 & 0.9678 & 0.7357 & 0.6370 & 0.9748 & 1.1331 & 0.4258 & 0.3623 & 1.0815 & 1.1996 & 0.4509 & 0.3886 \\
TabPFN-TS & - & - & - & - & 0.9282 & 1.1605 & 0.4358 & 0.3437 & 1.0017 & 1.1871 & 0.4365 & 0.3467 \\
TimesFM-2.5 & 0.7189 & 0.8721 & \textbf{0.6968} & \underline{0.5507} & \textbf{0.8890} & \underline{1.1222} & \textbf{0.4201} & \textbf{0.3312} & \underline{0.9798} & \textbf{1.1576} & \underline{0.4249} & \textbf{0.3370} \\
TiRex & 0.7144 & 0.8753 & 0.7025 & 0.5523 & 0.9026 & 1.1477 & 0.4281 & 0.3359 & \textbf{0.9740} & 1.1629 & 0.4306 & \underline{0.3447} \\
Toto-1.0 & \textbf{0.7076} & \textbf{0.8683} & 0.6986 & \textbf{0.5495} & 0.9050 & 1.1448 & 0.4278 & 0.3369 & 1.0440 & 1.2422 & 0.4597 & 0.3648 \\
\midrule
Ours & 0.7235 & 1.0231 & 0.7461 & 0.5524 & \underline{0.9011} & \textbf{1.1221} & \underline{0.4214} & \underline{0.3319} & 1.0080 & \underline{1.1581} & \textbf{0.4240} & 0.3462 \\
\bottomrule
\end{tabular}
}
\end{table*}

\begin{table*}[htbp]
\centering
\caption{Performance Comparison on \texttt{favorita\_stores} (1D/1W/1M) and \texttt{favorita\_transactions\_1D}. The best results are highlighted in \textbf{bold}, and the second-best results are \underline{underlined}.}
\label{tab:fev_cov_7}
\resizebox{\linewidth}{!}{
\begin{tabular}{l *{4}{cccc}} 
\toprule
\multicolumn{1}{c}{} &
\multicolumn{4}{c}{\textbf{stores\_1D}} &
\multicolumn{4}{c}{\textbf{stores\_1W}} &
\multicolumn{4}{c}{\textbf{stores\_1M}} &
\multicolumn{4}{c}{\textbf{transactions\_1D}} \\
\cmidrule(lr){2-5} \cmidrule(lr){6-9} \cmidrule(lr){10-13} \cmidrule(l){14-17}
Model & SQL & MASE & WAPE & WQL & SQL & MASE & WAPE & WQL & SQL & MASE & WAPE & WQL & SQL & MASE & WAPE & WQL \\
\midrule
AutoARIMA & 1.2216 & 1.4188 & 0.1934 & 0.1647 & 2.2969 & 2.5801 & 0.1501 & 0.1331 & 2.0382 & 2.2075 & 0.1385 & 0.1314 & 1.5622 & 1.7409 & 0.1205 & 0.1055 \\
AutoETS & 1.2379 & 1.3799 & 0.1872 & 0.1672 & 2.3568 & 2.5165 & 0.1484 & 0.1430 & 1.9424 & 2.1221 & 0.1568 & 0.1329 & 1.1813 & 1.2765 & 0.1010 & 0.0945 \\
AutoTheta & 1.2732 & 1.3759 & 0.1835 & 0.1779 & 2.3160 & 2.4768 & 0.1444 & 0.1408 & 1.9416 & 2.1335 & 0.1535 & 0.1245 & 1.2463 & 1.2511 & 0.0961 & 0.0991 \\
Chronos-Bolt & 1.0322 & 1.2689 & 0.1743 & 0.1419 & 2.1011 & 2.4749 & 0.1580 & 0.1268 & 2.0865 & 2.4178 & 0.2637 & 0.1852 & \underline{0.9750} & \underline{1.1515} & 0.0873 & 0.0736 \\
Moirai-2.0 & 0.9798 & 1.2047 & 0.1568 & 0.1279 & 2.1965 & 2.5773 & 0.1555 & 0.1243 & 2.0913 & 2.3236 & 0.2355 & 0.1726 & 1.1211 & 1.3295 & 0.0837 & 0.0705 \\
Naive & 2.6357 & 1.8830 & 0.3056 & 0.4079 & 2.4938 & 2.5171 & 0.1528 & 0.1608 & 2.0578 & \textbf{1.9974} & \underline{0.1233} & 0.1701 & 2.7304 & 1.8618 & 0.1561 & 0.2372 \\
Seasonal Naive & 1.6902 & 1.7590 & 0.2520 & 0.2376 & 2.4938 & 2.5171 & 0.1528 & 0.1608 & 2.0967 & 2.2823 & 0.1900 & 0.1715 & 1.7338 & 1.8200 & 0.1461 & 0.1403 \\
Stat. Ensemble & 1.1971 & 1.3400 & 0.1789 & 0.1606 & 2.2196 & 2.4340 & 0.1436 & 0.1336 & 1.9426 & \underline{2.1181} & 0.1467 & 0.1265 & 1.1848 & 1.2658 & 0.1005 & 0.0949 \\
Sundial-Base & 1.0613 & 1.2200 & 0.1562 & 0.1354 & 2.3082 & 2.5612 & 0.1497 & 0.1312 & 2.2543 & 2.4060 & 0.2561 & 0.2222 & 1.1982 & 1.3247 & 0.0799 & 0.0711 \\
TabPFN-TS & 0.9698 & 1.1943 & \underline{0.1486} & \underline{0.1204} & 2.1227 & 2.5268 & \underline{0.1260} & \underline{0.1011} & \underline{1.9336} & 2.1758 & \textbf{0.1159} & \textbf{0.0943} & 1.2252 & 1.6727 & \underline{0.0773} & \underline{0.0603} \\
TimesFM-2.5 & \textbf{0.9494} & \textbf{1.1676} & \textbf{0.1452} & \textbf{0.1184} & \textbf{1.9684} & \textbf{2.2901} & 0.1308 & 0.1095 & 1.9983 & 2.2338 & 0.2103 & 0.1509 & \textbf{0.8736} & \textbf{1.0842} & \textbf{0.0655} & \textbf{0.0539} \\
TiRex & 0.9682 & \underline{1.1934} & 0.1518 & 0.1243 & 2.0462 & 2.4235 & 0.1346 & 0.1134 & \textbf{1.8559} & 2.2147 & 0.1894 & 0.1422 & 1.0314 & 1.3493 & 0.0820 & 0.0677 \\
Toto-1.0 & 1.0364 & 1.2804 & 0.1784 & 0.1439 & 2.1277 & 2.5080 & 0.1511 & 0.1223 & 2.0094 & 2.2865 & 0.1978 & 0.1392 & 1.1139 & 1.4307 & 0.0947 & 0.0770 \\
\midrule
Ours & \underline{0.9596} & 1.2168 & 0.1550 & 0.1247 & \underline{1.9960} & \underline{2.3642} & \textbf{0.1259} & \textbf{0.0982} & 1.9776 & 2.1875 & 0.1571 & \underline{0.1215} & 1.0562 & 1.2099 & 0.0801 & 0.0676 \\
\bottomrule
\end{tabular}
}
\end{table*}

\begin{table*}[htbp]
\centering
\caption{Performance Comparison on \texttt{solar\_with\_weather\_15T} and \texttt{solar\_with\_weather\_1H}. The best results are highlighted in \textbf{bold}, and the second-best results are \underline{underlined}.}
\label{tab:fev_cov_8}\setlength{\tabcolsep}{7mm}
\resizebox{\linewidth}{!}{
\begin{tabular}{l *{2}{cccc}} 
\toprule
\multicolumn{1}{c}{} &
\multicolumn{4}{c}{\textbf{15T}} &
\multicolumn{4}{c}{\textbf{1H}} \\
\cmidrule(lr){2-5} \cmidrule(l){6-9}
Model & SQL & MASE & WAPE & WQL & SQL & MASE & WAPE & WQL \\
\midrule
AutoARIMA & — & — & — & — & 1.1314 & 1.1133 & 1.2902 & 1.2703 \\
AutoETS & 2.5289 & 2.4162 & 1.4215 & 1.9572 & 2.1818 & 2.2682 & 1.1465 & 1.3928 \\
AutoTheta & 3.5851 & 1.2729 & 1.5023 & 3.2873 & 3.3576 & 1.5354 & 1.5474 & 2.6715 \\
Chronos-Bolt & 0.8094 & 0.9765 & 1.2911 & 1.1301 & 0.8157 & 1.0720 & 1.3260 & 1.0190 \\
Moirai-2.0 & 0.8387 & 1.0764 & 1.3761 & 1.1205 & 0.9071 & 1.1358 & 1.4828 & 1.2285 \\
Naive & 2.2780 & 1.9895 & \textbf{1.0180} & 1.6964 & 2.1807 & 2.2682 & 1.1464 & 1.3916 \\
Seasonal Naive & 1.1938 & 1.0501 & 1.3115 & 1.4074 & 1.2120 & 1.0619 & 1.2339 & 1.3171 \\
Stat. Ensemble & — & — & — & — & 1.4584 & 1.3148 & 1.2682 & 1.1488 \\
Sundial-Base & 0.9626 & 1.1182 & 1.5059 & 1.3508 & 1.1815 & 1.3032 & 1.4481 & 1.3598 \\
TabPFN-TS & \underline{0.7471} & 0.9445 & 1.1662 & \underline{0.9712} & \underline{0.7006} & \textbf{0.8629} & \textbf{0.9096} & \textbf{0.7960} \\
TimesFM-2.5 & 0.9063 & 1.1153 & 1.4700 & 1.2405 & 0.8154 & 1.0513 & 1.2011 & 0.9696 \\
TiRex & 0.8457 & 1.0487 & 1.4236 & 1.2068 & 0.9000 & 1.1583 & 1.5261 & 1.2499 \\
Toto-1.0 & 0.7839 & \underline{0.9403} & 1.1480 & 0.9998 & 0.8760 & 1.0585 & 1.4068 & 1.2191 \\
\midrule
Ours & \textbf{0.7402} & \textbf{0.8951} & \underline{1.0609} & \textbf{0.8912} & \textbf{0.6751} & \underline{0.8774} & \underline{1.0128} & \underline{0.8174} \\
\bottomrule
\end{tabular}
}
\end{table*}

\begin{table*}[htbp]
\centering
\caption{Performance Comparison on \texttt{uci\_air\_quality\_1H} and \texttt{uci\_air\_quality\_1D}. The best results are highlighted in \textbf{bold}, and the second-best results are \underline{underlined}.}
\label{tab:fev_cov_9}\setlength{\tabcolsep}{6mm}
\resizebox{\linewidth}{!}{
\begin{tabular}{l *{2}{cccc}} 
\toprule
\multicolumn{1}{c}{} &
\multicolumn{4}{c}{\textbf{1H}} &
\multicolumn{4}{c}{\textbf{1D}} \\
\cmidrule(lr){2-5} \cmidrule(l){6-9}
Model & SQL & MASE & WAPE & WQL & SQL & MASE & WAPE & WQL \\
\midrule
AutoARIMA & 1.1924 & 1.3698 & 0.4518 & 0.3924 & 1.2403 & 1.5585 & 0.3229 & 0.2530 \\
AutoETS & 4.10 $\times 10^{5}$ & 10.6979 & 4.0642 & 2.05 $\times 10^{5}$ & 1.1813 & \textbf{1.3629} & \textbf{0.2791} & 0.2425 \\
AutoTheta & 1.9560 & 1.3458 & 0.4168 & 0.6724 & 1.2334 & 1.4428 & 0.2964 & 0.2579 \\
Chronos-Bolt & 0.8990 & 1.1162 & 0.3758 & 0.3020 & \textbf{1.0920} & 1.3887 & \underline{0.2843} & \textbf{0.2229} \\
Moirai-2.0 & 0.9454 & 1.1965 & 0.3904 & 0.3063 & 1.1380 & 1.4417 & 0.2916 & \underline{0.2299} \\
Naive & 2.4250 & 1.6816 & 0.5251 & 0.8361 & 1.8739 & 2.1091 & 0.4315 & 0.3990 \\
Seasonal Naive & 1.3840 & 1.4425 & 0.4563 & 0.4609 & 1.4013 & 1.7337 & 0.3624 & 0.2922 \\
Stat. Ensemble & 1.5607 & 1.2992 & 0.4116 & 0.5269 & 1.1225 & 1.3837 & 0.2860 & 0.2304 \\
Sundial-Base & 1.0009 & 1.1924 & 0.4007 & 0.3369 & 1.2145 & 1.3991 & 0.2893 & 0.2497 \\
TabPFN-TS & 0.9312 & 1.1751 & 0.3858 & 0.3067 & 1.1863 & 1.5308 & 0.3106 & 0.2398 \\
TimesFM-2.5 & 0.8769 & 1.1226 & 0.3715 & 0.2904 & 1.2051 & 1.5147 & 0.3069 & 0.2425 \\
TiRex & \textbf{0.8650} & \textbf{1.1062} & 0.3698 & 0.2890 & 1.1280 & 1.4312 & 0.2942 & 0.2322 \\
Toto-1.0 & \underline{0.8700} & \underline{1.1112} & \textbf{0.3658} & \underline{0.2863} & 1.2602 & 1.5881 & 0.3233 & 0.2553 \\
\midrule
Ours & 0.8732 & 1.1224 & \underline{0.3685} & \textbf{0.2794} & \underline{1.1122} & \underline{1.3667} & 0.2881 & 0.2317 \\
\bottomrule
\end{tabular}
}
\end{table*}


\begin{table*}[htbp]
\centering
\caption{Average results on load, photovoltaic, and all real application datasets. The best results are highlighted in \textbf{bold}, and the second-best results are \underline{underlined}.}\label{tab:cov_real_app}\setlength{\tabcolsep}{2.5mm}
\resizebox{\textwidth}{!}{%
\begin{tabular}{l *{3}{cccc}} 
\toprule
\multicolumn{1}{c}{} &
\multicolumn{4}{c}{\textbf{Load datasets (18)}} &
\multicolumn{4}{c}{\textbf{Photovoltaic datasets (9)}} &
\multicolumn{4}{c}{\textbf{All real-world application datasets (27)}} \\
\cmidrule(lr){2-5} \cmidrule(lr){6-9} \cmidrule(l){10-13}
Model & SQL & MASE & WAPE & WQL & SQL & MASE & WAPE & WQL & SQL & MASE & WAPE & WQL \\
\midrule
AutoARIMA      & 2.5730 & 0.8526 & 0.0395 & 0.1289 & 3.5350 & 0.7858 & 0.2366 & 1.0570 & 2.8936 & 0.8303 & 0.1052 & 0.4383 \\
AutoETS        & 7.4514 & 6.4652 & 0.2931 & 0.3218 & 22.4679 & 5.7401 & 1.7096 & 6.7013 & 12.4569 & 6.2235 & 0.7653 & 2.4483 \\
AutoTheta      & 0.8977 & 1.2644 & 0.0632 & 0.0385 & 38.8877 & 6.6861 & 1.9921 & 11.5991 & 13.5611 & 3.0717 & 0.7062 & 3.8920 \\
Chronos-Bolt      & 0.4415 & \underline{0.5491} & \underline{0.0253} & \underline{0.0203} & 0.6471 & 0.8066 & 0.2424 & 0.1945 & 0.5101 & 0.6350 & 0.0976 & 0.0783 \\
Moirai-2.0     & 0.7621 & 0.9701 & 0.0460 & 0.0360 & 0.6175 & 0.7222 & 0.2175 & 0.1859 & 0.7139 & 0.8875 & 0.1031 & 0.0860 \\
Naive          & 2.4675 & 3.0599 & 0.1479 & 0.1182 & 4.9932 & 5.7400 & 1.7095 & 1.4883 & 3.3094 & 3.9533 & 0.6684 & 0.5749 \\
Seasonal Naive & 1.0799 & 1.2065 & 0.0596 & 0.0499 & 6.2536 & 0.7382 & 0.2221 & 1.8697 & 2.8045 & 1.0504 & 0.1138 & 0.6565 \\
Stat. Ensemble & 1.6205 & 2.2320 & 0.1051 & 0.0760 & 3.2080 & 3.2719 & 0.9756 & 0.9570 & 2.1497 & 2.5786 & 0.3952 & 0.3696 \\
Sundial-Base   & 0.7536 & 0.9411 & 0.0450 & 0.0360 & 0.7387 & 0.8304 & 0.2500 & 0.2225 & 0.7486 & 0.9042 & 0.1133 & 0.0981 \\
TabPFN-TS      & \underline{0.5120} & 0.6817 & 0.0325 & 0.0242 & \underline{0.3750} & \underline{0.4710} & \underline{0.1415} & \underline{0.1127} & \underline{0.4663} & \underline{0.6115} & \underline{0.0688} & \underline{0.0537} \\
TimesFM-2.0    & 0.9410 & 1.0132 & 0.0506 & 0.0469 & 0.6434 & 0.7572 & 0.2266 & 0.1925 & 0.8418 & 0.9278 & 0.1093 & 0.0954 \\
TiRex          & 0.5544 & 0.6896 & 0.0336 & 0.0270 & 0.6042 & 0.7490 & 0.2253 & 0.1813 & 0.5710 & 0.7094 & 0.0975 & 0.0784 \\
Toto-1.0       & 0.8888 & 1.1359 & 0.0557 & 0.0430 & 0.7465 & 0.8981 & 0.2694 & 0.2241 & 0.8414 & 1.0566 & 0.1269 & 0.1034 \\
\midrule
\textbf{Ours}  & \textbf{0.4089} & \textbf{0.5182} & \textbf{0.0219} & \textbf{0.0174} &
                  \textbf{0.3743} & \textbf{0.4504} & \textbf{0.1352} & \textbf{0.1121} &
                  \textbf{0.3974} & \textbf{0.4956} & \textbf{0.0597} & \textbf{0.0490} \\
\bottomrule
\end{tabular}%
}
\end{table*}

\subsection{Results on Univariate Tasks}\label{app:exp_fev_uni}
\begin{table*}[htbp]
    \centering
    \caption{Summary of the 10 representative univariate forecasting tasks selected from \texttt{fev-bench-mini}, collectively referred to as \texttt{fev-bench-uni}. Datasets with multiple targets (\# targets $>1$) are decomposed into independent univariate series for evaluation.}
    \label{tab:fev_datasetInfo_uni}
    \resizebox{\linewidth}{!}{
    \begin{tabular}{l c c c c c c c c c c}
    \toprule
    Task & Domain & Frequency & \# items & median length & \# obs & \# known dynamic cols & H & W & \# seasonality & \# targets \\
    \midrule
    \texttt{ETT\_15T} & energy & 15 Min & 2 & 69,680 & 975,520 & 0 & 96 & 20 & 96 & 7 \\
    \texttt{ETT\_1H} & energy & 1 H & 2 & 17,420 & 243,880 & 0 & 168 & 20 & 24 & 7 \\
    \texttt{bizitobs\_l2c\_5T} & cloud & 5 Min & 1 & 31,968 & 223,776 & 0 & 288 & 20 & 288 & 7 \\
    \texttt{boomlet\_619} & cloud & 1 Min & 1 & 16,384 & 851,968 & 0 & 60 & 20 & 1440 & 52 \\
    \texttt{boomlet\_1282} & cloud & 1 Min & 1 & 16,384 & 573,440 & 0 & 60 & 20 & 1440 & 35 \\
    \texttt{boomlet\_1676} & cloud & 30 Min & 1 & 10,463 & 1,046,300 & 0 & 96 & 20 & 48 & 100 \\
    \texttt{hospital\_admissions\_1D} & healthcare & Daily & 8 & 1,731 & 13,846 & 0 & 28 & 20 & 7 & 1 \\
    \texttt{hospital\_admissions\_1W} & healthcare & Weekly & 8 & 246 & 1,968 & 0 & 13 & 16 & 1 & 1 \\
    \texttt{jena\_weather\_1H} & nature & 1 H & 1 & 8,784 & 184,464 & 0 & 24 & 20 & 24 & 21 \\
    \texttt{M\_DENSE\_1D} & mobility & Daily & 30 & 730 & 21,900 & 0 & 28 & 10 & 7 & 1 \\
    \bottomrule
\end{tabular}
    }
\end{table*}

\begin{figure}[htbp]
    \centering
    \includegraphics[width=0.6\linewidth]{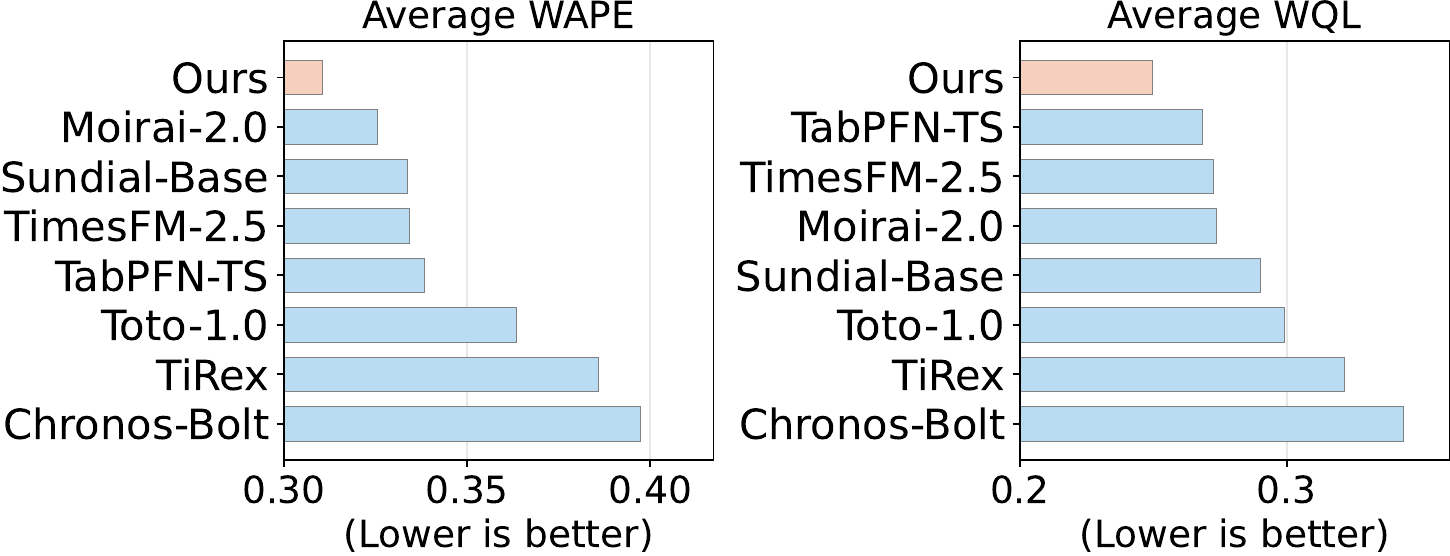}
    \caption{Probabilistic forecasting evaluation on \texttt{fev-bench-uni} using WAPE and WQL.}
    \label{fig:uni_WAPE_WQL}
\end{figure}

We provide detailed information in Table~\ref{tab:fev_datasetInfo_uni} for the 10 univariate and multivariate forecasting tasks selected from \texttt{fev-bench-mini}, which we denote as \texttt{fev-bench-uni}.
The corresponding ranking results, evaluated by macro-averaged SQL, MASE, WAPE, and WQL across these 10 datasets, are presented in Table~\ref{tab:fev_uni_ranking}.
While adapting our 3D model to univariate time series forecasting, we formulate the input with a dummy covariate (e.g., a column of zeros) alongside intrinsic temporal features such as year, month, day, and hour to preserve compatibility with the model’s covariate-aware architecture.
Despite the absence of meaningful external covariates, Baguan-TS exhibits highly competitive performance on univariate forecasting tasks. Notably, it achieves best results in both WAPE and WQL (see Table~\ref{tab:fev_uni_ranking} and Fig.~\ref{fig:uni_WAPE_WQL}). 
This strong performance underscores the model’s robustness in capturing distributional characteristics and its reliability in high-stakes or extreme-value scenarios, where WAPE emphasizes large errors proportionally to actual magnitudes, and WQL explicitly optimizes quantile-based risk awareness.
Detailed experimental results are presented in Tables~\ref{tab:fev_uni_1}--\ref{tab:fev_uni_4}.

\begin{table}[H]
\centering
\caption{Average results on \texttt{fev-bench-uni}. The best results are highlighted in \textbf{bold}.}
\label{tab:fev_uni_ranking}\setlength{\tabcolsep}{4mm}
\resizebox{0.5\linewidth}{!}{
\begin{tabular}{lcccc}
\toprule
Model & SQL & MASE  & WAPE  & WQL \\
\midrule
Chronos-Bolt & 0.6083 & 0.7369 & 0.3974 & 0.3438 \\
Moirai-2.0 & \textbf{0.5405} & \textbf{0.6728} & 0.3254 & 0.2735 \\
Sundial-Base & 0.5814 & 0.6833 & 0.3338 & 0.2902 \\
TabPFN-TS & 0.5920 & 0.7364 & 0.3384 & 0.2685 \\
TimesFM-2.5 & 0.5428 & 0.6831 & 0.3341 & 0.2726 \\
TiRex & 0.5684 & 0.7104 & 0.3858 & 0.3215 \\
Toto-1.0 & 0.5689 & 0.7147 & 0.3635 & 0.2992 \\
\midrule
AutoARIMA & 0.6913 & 0.8301 & 0.5843 & 0.5548 \\
Stat. Ensemble & 0.7575 & 0.8371 & 0.5132 & 0.5844 \\
AutoETS & 0.9083 & 0.9519 & 0.4877 & 0.6018 \\
AutoTheta & 0.9409 & 0.8532 & 0.4972 & 0.6327 \\
Seasonal Naive & 1.0381 & 1.0373 & 0.6649 & 0.7433 \\
Naive & 1.3600 & 1.0902 & 0.5254 & 0.7872 \\
\midrule
Ours & 0.5664& 0.7126 & \textbf{0.3105} & \textbf{0.2497} \\
\bottomrule
\end{tabular}
}
\end{table}

\begin{table}[htbp]
\centering
\caption{Performance Comparison on \texttt{ETT\_15T} and \texttt{ETT\_1H}. The best results are highlighted in \textbf{bold}, and the second-best results are \underline{underlined}.}
\label{tab:fev_uni_1}\setlength{\tabcolsep}{1mm}
\resizebox{0.5\linewidth}{!}{
\begin{tabular}{l *{2}{cccc}} 
\toprule
\multicolumn{1}{c}{} &
\multicolumn{4}{c}{\textbf{15T}} &
\multicolumn{4}{c}{\textbf{1H}} \\
\cmidrule(lr){2-5} \cmidrule(l){6-9}
Model & SQL & MASE & WAPE & WQL & SQL & MASE & WAPE & WQL \\
\midrule
AutoARIMA & — & — & — & — & 1.0524 & 1.2616 & 0.2730 & 0.2275 \\
AutoETS & 1.2626 & 1.4293 & 0.2090 & 0.1890 & 1.7645 & 1.6021 & 0.3240 & 0.3554 \\
AutoTheta & 1.0985 & 0.8022 & 0.1328 & 0.1702 & 2.0613 & 1.2847 & 0.2708 & 0.3879 \\
Chronos-Bolt & 0.5737 & \textbf{0.7033} & \textbf{0.1146} & 0.0933 & 0.9436 & 1.1267 & 0.2465 & 0.2059 \\
Moirai-2.0 & \textbf{0.5634} & \underline{0.7119} & 0.1160 & \textbf{0.0915} & 0.8993 & 1.1240 & 0.2470 & 0.1952 \\
Naive & 1.3269 & 1.3671 & 0.2028 & 0.2051 & 2.3137 & 1.7184 & 0.3389 & 0.4903 \\
Seasonal Naive & 0.7625 & 0.9169 & 0.1473 & 0.1224 & 1.2069 & 1.3227 & 0.2864 & 0.2597 \\
Stat. Ensemble & — & — & — & — & 1.2717 & 1.2519 & 0.2621 & 0.2620 \\
Sundial-Base & 0.5971 & 0.7139 & \underline{0.1157} & 0.0970 & 0.9634 & 1.1439 & 0.2548 & 0.2147 \\
TabPFN-TS & 0.6024 & 0.7625 & 0.1222 & 0.0966 & 0.9332 & 1.1774 & 0.2553 & 0.2034 \\
TimesFM-2.5 & 0.5772 & 0.7295 & 0.1167 & \underline{0.0925} & 0.8823 & 1.1239 & \underline{0.2452} & \underline{0.1916} \\
TiRex & \underline{0.5683} & 0.7188 & 0.1158 & \textbf{0.0915} & \underline{0.8736} & \underline{1.1178} & 0.2477 & 0.1924 \\
Toto-1.0 & 0.5930 & 0.7578 & 0.1216 & 0.0952 & \textbf{0.8727} & \textbf{1.1129} & \textbf{0.2432} & \textbf{0.1900} \\
\midrule
Ours & 0.6121 & 0.7810 & 0.1253 & 0.0980 & 0.9036 & 1.1549 & 0.2566 & 0.2022 \\
\bottomrule
\end{tabular}
}
\end{table}
\begin{table}[htbp]
\centering
\caption{Performance Comparison on \texttt{hospital\_admissions\_1D} and \texttt{hospital\_admissions\_1W}. The best results are highlighted in \textbf{bold}, and the second-best results are \underline{underlined}.}
\label{tab:fev_uni_3}\setlength{\tabcolsep}{1mm}
\resizebox{0.5\linewidth}{!}{
\begin{tabular}{l *{2}{cccc}} 
\toprule
\multicolumn{1}{c}{} &
\multicolumn{4}{c}{\textbf{1D}} &
\multicolumn{4}{c}{\textbf{1W}} \\
\cmidrule(lr){2-5} \cmidrule(l){6-9}
Model & SQL & MASE & WAPE & WQL & SQL & MASE & WAPE & WQL \\
\midrule
AutoARIMA & 0.5556 & 0.7209 & 0.5348 & 0.4122 & 0.5793 & \underline{0.7541} & \textbf{0.2123} & 0.1631 \\
AutoETS & 0.5558 & 0.7211 & 0.5350 & 0.4123 & \textbf{0.5783} & \underline{0.7541} & \textbf{0.2123} & \textbf{0.1628} \\
AutoTheta & 0.5748 & 0.7429 & 0.5510 & 0.4264 & 0.5977 & 0.7779 & 0.2191 & 0.1683 \\
Chronos-Bolt & 0.5562 & 0.7195 & 0.5337 & 0.4125 & 0.5868 & 0.7623 & 0.2146 & 0.1653 \\
Moirai-2.0 & 0.5556 & 0.7188 & 0.5332 & 0.4121 & 0.5862 & 0.7643 & 0.2153 & 0.1652 \\
Naive & 1.3251 & 0.9747 & 0.7243 & 0.9837 & 1.0492 & 1.0436 & 0.2934 & 0.2969 \\
Seasonal Naive & 0.8572 & 1.0268 & 0.7622 & 0.6361 & 1.0492 & 1.0436 & 0.2934 & 0.2969 \\
Stat. Ensemble & 0.5570 & 0.7214 & 0.5352 & 0.4131 & \underline{0.5789} & 0.7552 & 0.2126 & \underline{0.1630} \\
Sundial-Base & 0.6103 & 0.7225 & 0.5359 & 0.4526 & 0.6411 & 0.7599 & 0.2139 & 0.1802 \\
TabPFN-TS & 0.5623 & 0.7245 & 0.5375 & 0.4171 & 0.5814 & \textbf{0.7534} & \textbf{0.2123} & 0.1638 \\
TimesFM-2.5 & 0.5561 & 0.7193 & 0.5336 & 0.4125 & 0.5795 & 0.7545 & \underline{0.2124} & 0.1632 \\
TiRex & \underline{0.5551} & 0.7180 & 0.5326 & \underline{0.4118} & 0.5851 & 0.7610 & 0.2142 & 0.1648 \\
Toto-1.0 & 0.5555 & \underline{0.7173} & \underline{0.5321} & 0.4120 & 0.5976 & 0.7781 & 0.2188 & 0.1681 \\
\midrule
Ours & \textbf{0.5546} & \textbf{0.7167} & \textbf{0.5317} & \textbf{0.4114} & 0.6488 & 0.8405 & 0.2368 & 0.1827 \\
\bottomrule
\end{tabular}
}
\end{table}

\begin{table}[htbp]
\centering
\caption{Performance Comparison on \texttt{boomlet\_619}, \texttt{boomlet\_1676}, and \texttt{boomlet\_1282}. The best results are highlighted in \textbf{bold}, and the second-best results are \underline{underlined}.}
\label{tab:fev_uni_2}\setlength{\tabcolsep}{3mm}
\resizebox{\linewidth}{!}{
\begin{tabular}{l *{3}{cccc}} 
\toprule
\multicolumn{1}{c}{} &
\multicolumn{4}{c}{\textbf{619}} &
\multicolumn{4}{c}{\textbf{1676}} &
\multicolumn{4}{c}{\textbf{1282}} \\
\cmidrule(lr){2-5} \cmidrule(lr){6-9} \cmidrule(l){10-13}
Model & SQL & MASE & WAPE & WQL & SQL & MASE & WAPE & WQL & SQL & MASE & WAPE & WQL \\
\midrule
AutoARIMA & 0.5545 & 0.7170 & 0.5880 & 0.4527 & — & — & — & — & 0.5565 & 0.5925 & 0.4050 & 0.3750 \\
AutoETS & 0.8944 & 1.0938 & 0.9529 & 0.7636 & 0.7564 & 0.7610 & 0.3540 & 0.3561 & 0.9136 & 0.6890 & 0.4705 & 0.6261 \\
AutoTheta & 0.8352 & 1.0200 & 0.8821 & 0.6993 & 0.7835 & 0.7772 & 0.3606 & 0.3690 & 0.9717 & 0.6910 & 0.4718 & 0.6673 \\
Chronos-Bolt & 0.4709 & 0.5941 & 0.4683 & 0.3764 & 0.6077 & 0.7192 & 0.3289 & 0.2758 & 0.4618 & 0.5647 & 0.3918 & 0.3174 \\
Moirai-2.0 & \underline{0.3294} & 0.4306 & 0.3062 & 0.2356 & 0.5727 & 0.7128 & 0.3274 & 0.2580 & 0.4269 & 0.5230 & 0.3639 & 0.2944 \\
Naive & 1.2752 & 1.1236 & 0.9459 & 0.9679 & 1.3196 & 0.8199 & 0.3738 & 0.5376 & 1.5394 & 0.8299 & 0.5674 & 1.0473 \\
Seasonal Naive & 1.2752 & 1.1236 & 0.9459 & 0.9679 & 0.8504 & 0.9529 & 0.4490 & 0.3899 & 1.5394 & 0.8299 & 0.5674 & 1.0473 \\
Stat. Ensemble & 0.7772 & 1.0047 & 0.8714 & 0.6586 & — & — & — & — & 0.7391 & 0.6398 & 0.4330 & 0.4977 \\
Sundial-Base & 0.3705 & 0.4356 & 0.3110 & 0.2649 & 0.6146 & 0.7138 & 0.3250 & 0.2744 & 0.4522 & 0.5169 & 0.3604 & 0.3126 \\
TabPFN-TS & 0.3305 & \underline{0.4305} & \underline{0.3036} & \underline{0.2349} & 0.8311 & 0.9587 & 0.3439 & 0.2853 & 0.4253 & 0.5067 & 0.3518 & 0.2921 \\
TimesFM-2.5 & 0.3398 & 0.4364 & 0.3117 & 0.2458 & \underline{0.5626} & \underline{0.6982} & \underline{0.3195} & \underline{0.2520} & \textbf{0.4034} & \textbf{0.4938} & \textbf{0.3418} & \textbf{0.2770} \\
TiRex & 0.3411 & 0.4418 & 0.3174 & 0.2471 & 0.5712 & 0.7093 & 0.3220 & 0.2548 & 0.4089 & \underline{0.4966} & 0.3447 & 0.2814 \\
Toto-1.0 & \textbf{0.3099} & \textbf{0.4032} & \textbf{0.2789} & \textbf{0.2163} & \textbf{0.5544} & \textbf{0.6907} & \textbf{0.3174} & \textbf{0.2495} & \underline{0.4069} & 0.4983 & \underline{0.3444} & \underline{0.2786} \\
\midrule
Ours & 0.3399 & 0.4391 & 0.3128 & 0.2445 & 0.6195 & 0.7691 & 0.3482 & 0.2760 & 0.4169 & 0.5077 & 0.3529 & 0.2877 \\
\bottomrule
\end{tabular}
}
\end{table}

\begin{table}[htbp]
\centering
\caption{Performance Comparison on \texttt{bizitobs\_l2c\_5T}, \texttt{jena\_weather\_1H}, and \texttt{M\_DENSE\_1D}. The best results are highlighted in \textbf{bold}, and the second-best results are \underline{underlined}.}
\label{tab:fev_uni_4}\setlength{\tabcolsep}{3mm}
\resizebox{\linewidth}{!}{
\begin{tabular}{l *{3}{cccc}} 
\toprule
\multicolumn{1}{c}{} &
\multicolumn{4}{c}{\textbf{bizitobs\_l2c\_5T}} &
\multicolumn{4}{c}{\textbf{jena\_weather\_1H}} &
\multicolumn{4}{c}{\textbf{M\_DENSE\_1D}} \\
\cmidrule(lr){2-5} \cmidrule(lr){6-9} \cmidrule(l){10-13}
Model & SQL & MASE & WAPE & WQL & SQL & MASE & WAPE & WQL & SQL & MASE & WAPE & WQL \\
\midrule
AutoARIMA & 0.8247 & 0.9163 & 1.5980 & 1.8772 & 0.4370 & 0.5092 & 0.9375 & 0.8257 & 0.9702 & 1.1690 & 0.1261 & 0.1051 \\
AutoETS & 0.7311 & 0.8137 & 1.3367 & 1.6114 & 0.5529 & 0.5678 & 0.3522 & 1.4085 & 1.0735 & 1.0869 & 0.1301 & 0.1324 \\
AutoTheta & 0.7496 & 0.8740 & 1.6561 & 1.6069 & 0.5933 & 0.4846 & 0.2955 & 1.6859 & 1.1435 & 1.0772 & 0.1322 & 0.1459 \\
Chronos-Bolt & 0.7570 & 0.8002 & 1.3254 & 1.2750 & 0.3668 & 0.4543 & 0.2519 & 0.2341 & 0.7590 & 0.9249 & 0.0985 & 0.0817 \\
Moirai-2.0 & \textbf{0.3675} & \textbf{0.4035} & 0.7997 & 0.7682 & 0.3652 & 0.4505 & \underline{0.2487} & 0.2342 & 0.7390 & \underline{0.8884} & 0.0962 & {0.0807} \\
Naive & 0.6779 & 0.7578 & 1.2819 & 1.4508 & 0.5789 & 0.5375 & 0.3286 & 1.6602 & 2.1937 & 1.7292 & 0.1970 & 0.2323 \\
Seasonal Naive & 0.9226 & 1.0667 & 2.2838 & 2.2878 & 0.6550 & 0.7399 & 0.7712 & 1.2878 & 1.2627 & 1.3500 & 0.1420 & 0.1368 \\
Stat. Ensemble & 0.7197 & 0.8069 & 1.3437 & 1.5129 & 0.4516 & 0.4664 & 0.3279 & 1.0549 & 0.9648 & 1.0509 & 0.1193 & 0.1131 \\
Sundial-Base & \underline{0.4004} & \underline{0.4808} & \underline{0.6893} & \underline{0.5858} & 0.3803 & 0.4459 & 0.4352 & 0.4351 & 0.7845 & 0.9000 & 0.0968 & 0.0844 \\
TabPFN-TS & 0.4855 & 0.6193 & 0.8823 & 0.6797 & 0.4128 & 0.5111 & 0.2825 & 0.2343 & 0.7560 & 0.9195 & \textbf{0.0927} & \underline{0.0776} \\
TimesFM-2.5 & 0.4611 & 0.5787 & 0.9057 & 0.7905 & \underline{0.3589} & \underline{0.4392} & 0.2604 & 0.2233 & \textbf{0.7075} & \textbf{0.8576} & \underline{0.0935} & \textbf{0.0774} \\
TiRex & 0.6789 & 0.7959 & 1.4136 & 1.2725 & \textbf{0.3557} & \textbf{0.4379} & 0.2507 & \underline{0.2168} & 0.7460 & 0.9066 & 0.0993 & 0.0822 \\
Toto-1.0 & 0.5954 & 0.7136 & 1.2221 & 1.0847 & 0.3616 & 0.4476 & \textbf{0.2468} & \textbf{0.2069} & 0.8418 & 1.0279 & 0.1093 & 0.0907 \\
\midrule
Ours & 0.4360 & 0.5379 & \textbf{0.5545} & \textbf{0.4372} & 0.3949 & 0.4827 & 0.2883 & 0.2758 & \underline{0.7374} & 0.8962 & 0.0976 & 0.0812 \\
\bottomrule
\end{tabular}
}
\end{table}

\end{document}